# Deep Learning Innovations for Energy Efficiency: Advances in Non-Intrusive Load Monitoring and EV Charging Optimization for a Sustainable Grid

**Stavros Sykiotis**

School of Rural, Surveying and Geoinformatics Engineering
National Technical University of Athens

This dissertation is submitted for the degree of
*Doctor of Philosophy*

# Examination Committee

1. Anastasios Doulamis, Professor · School of Rural, Surveying and Geoinformatics Engineering · National Technical University of Athens, Supervisor
2. Charalampos Ioannidis, Professor · School of Rural, Surveying and Geoinformatics Engineering · National Technical University of Athens
3. Theodora Varvarigou, Professor · School of Electrical and Computer Engineering · National Technical University of Athens
4. Vasileios Veskoukis, Professor · School of Rural, Surveying and Geoinfromatics Engineering · National Technical University of Athens
5. Chryssi Potsiou, Professor · School of Rural, Surveying and Geoinfromatics Engineering · National Technical University of Athens
6. Dimitrios Kosmopoulos, Professor · School of Computer Engineering and Informatics · University of Patras
7. Eftychios Protopapadakis, As. Professor · School of Information Sciences · University of Macedonia

To my father
In loving memory.

# Declaration

I hereby declare that except where specific reference is made to the work of others, the contents of this dissertation are original and have not been submitted in whole or in part for consideration for any other degree or qualification in this, or any other university.

Stavros Sykiotis
March 2025

# Abstract

The global energy landscape is undergoing a profound transformation, often referred to as the energy transition, driven by the urgent need to mitigate climate change, reduce greenhouse gas emissions, and ensure sustainable energy supplies. However, the undoubted complexity of new investments in renewables, as well as the phase out of high $CO_2$-emission energy sources, hampers the pace of the energy transition and raises doubts as to whether new renewable energy sources are capable of solely meeting the climate target goals. This highlights the need to investigate alternative pathways to accelerate the energy transition, by identifying human activity domains with higher/excessive energy demands. Two notable examples where there is room for improvement, in the sense of reducing energy consumption and consequently $CO_2$ emissions, are residential energy consumption and road transport. This dissertation investigates the development of novel Deep Learning techniques to create tools which solve limitations in these two key energy domains. Reduction of residential energy consumption can be achieved by empowering end-users with the user of Non-Intrusive Load Monitoring, whereas optimization of EV charging with Deep Reinforcement Learning can tackle road transport decarbonization.

The dissertation is divided in three parts. Part I (Chapters 1, 2) presents the introduction and motivation of the problem at hand. The motivation for the importance of the tackled topic is translated into research questions, which will be answered in the subsequent chapters. Moreover, a mathematical formulation of the two examined energy efficiency domains is introduced, serving as a foundational block for the better and deeper understanding of the conducted work. Part II (Chapters 3, 4, 5) focuses on Non-Intrusive Load Monitoring applications, investigating (1) the development of a novel deep learning NILM architecture with focus on data efficiency and adaptability to diverse household contexts, (2) optimizing NILM models for deployment on edge devices with minimal performance degradation, and (3) creating methods to monitor and adapt models in the post-deployment phase, allowing efficient fine-tuning and retraining to maintain high performance. Finally, Part III (Chapters 6, 7, 8) relates to the optimization of EV charging schedules. The designed methodologies focus not only on charging cost reduction, but also on the environmental benefits, by exploring: (1) real-time EV charging schedules that prioritize renewable energy, (2) strategies to minimize grid stress by maximizing local consumption of distributed energy sources, and (3) optimizing multiple EV charging schedules simultaneously to enhance shared resource utilization. The dissertation is concluded with a summary of the key findings as well as outlines potential future research directions (Chapter 9).

# Acknowledgements

The past three and a half years of my doctoral studies have been an intense, challenging, and immensely rewarding journey. As I reach the end of this unforgettable chapter, I would like to express my heartfelt gratitude to those who have supported and guided me throughout this time.

First and foremost, I extend my deepest appreciation to my supervisor, Prof. Anastasios Doulamis, for his invaluable guidance, patience, and unwavering support during my research. He has not only imparted essential methodologies for conducting rigorous research but also fostered my independence, giving me the freedom to explore and refine my ideas. It has been an honor and a privilege to work and study under his mentorship, which has profoundly shaped my academic and professional growth. I am also sincerely grateful to Prof. Nikolaos Doulamis for his crucial guidance and encouragement over these past years. Together, they have built a collaborative and dynamic team, which I am fortunate to be a part of.

I would also like to thank the esteemed members of my advisory committee, Prof. Charalampos Ioannidis and Prof. Theodora Varvarigou, for their insightful feedback and ongoing support throughout my PhD journey. Additionally, I am grateful to the distinguished members of the examination committee - Prof. Chryssi Potsiou, Prof. Vasileios Veskoukis, Prof. Dimitrios Kosmopoulos and As. Prof. Eftychios Protopapadakis - for their participation and for their constructive and valuable comments.

A special thank you goes to my fellow PhD candidates and colleagues in the NTUA Photogrammetry Lab. The moments we shared and the friendships we formed have been invaluable, and I look forward to these bonds continuing well into the future. I am especially thankful to Ioannis Rallis and Maria Kaselimi for their guidance and advice during my first steps in research.

My gratitude extends as well to the H2020 MSCA-ITN GECKO project for providing the financial support that made this dissertation possible, and to my fellow GECKO PhD candidates for broadening my research horizons. Special thanks to Christoforos Menos and Sotirios Athanasoulias for our enriching research collaborations.

Finally, I am deeply thankful to my mother, Chryssa, my sister, Aggeliki, and my friends for their unwavering support and encouragement throughout this journey.

# Contents

# List of Figures

# List of Tables

# List of Abbreviations and Nomenclature

## Abbreviations

| | |
|---|---|
| AI | Artificial Intelligence |
| BCE | Binary Cross-Entropy |
| BEV | Battery Electric Vehicle |
| BSS | Blind Source Separation |
| CEMS | Community Energy Management System |
| CNN | Convolutional Neural Networks |
| DDQN | Double Deep Q-Learning |
| DER | Distributed Energy Resources |
| DQN | Deep Q-Learning |
| DR | Demand Response |
| DRL | Deep Reinforcement Learning |
| EV | Electric Vehicle |
| FLOPs | Floating Point Operations |
| GAN | Generative Adversarial Networks |
| GRU | Gated Recurrent Units |
| HEMS | Home Energy Management System |
| ICE | Internal Combustion Engine |
| LSTM | Long-Short Term Memory |
| MADRL | Multi-Agent Deep Reinforcement Learning |
| MDP | Markov Decision Process |
| MHA | Multi-head Attention |
| MRE | Mean Relative Error |
| MSE | Mean Squared Error |
| NILM | Non-Intrusive Load Monitoring |
| NLP | Natural Language Processing |
| PAOP | Performance-Aware Optimized Pruning |
| PAOPQ | Performance-Aware Optimized Pruning and Quantization |
| PAR | Peak-to-Average Ratio |
| PFFN | Position-wise feed-forward network |
| PV | Photovoltaic |
| RL | Reinforcement Learning |
| RNN | Recurrent Neural Networks |
| RV | Random Variable |
| SCI | Self-consumption Index |
| seq2point | Sequence-to-point |
| seq2seq | Sequence-to-sequence |
| seq2subseq | Sequence-to-subsequence |
| SoC | State-of-charge |
| TEC | Total Electricity Cost |
| ToU | Time-of-use |

# Part I

# Introduction and fundamentals

# Chapter 1

# Introduction

The global energy landscape is undergoing a profound transformation, often referred to as the energy transition, driven by the urgent need to mitigate climate change, reduce greenhouse gas emissions, and ensure sustainable energy supplies. The transformation of the energy sector is driven by the need to mitigate global warming, which, according to the Intergovernmental Panel on Climate Change (IPCC), is expected to lead to a 1.5-2°C temperature increase by 2100 [1, 2]. Recent policy-level initiatives propel infrastructure investments in new renewable energy sources, such as photovoltaic panels, wind turbines or energy storage mechanisms, towards transitioning to a clean, renewables-dominated energy mix. Notable examples of such initiatives, underlying the importance of a fast-paced transition towards climate neutrality, are the recently established European New Green Deal [3] and the European Green Law [4]. However, the undoubted complexity of new investments, as well as the phase out of high $CO_2$-emission energy sources, hampers the pace of the energy transition and raises doubts as to whether new renewable energy sources are solely capable of meeting the aforementioned goals.

A deeper dive in the search for ways to reduce $CO_2$ emissions requires the analysis of the overall societal energy consumption profile. Such an analysis assists in identifying human activity domains which contribute significantly to a higher energy demand. Two notable examples with room for improvement, in the sense of reducing energy consumption and consequently $CO_2$ emissions, are residential energy consumption and road transport. According to Eurostat, residential energy consumption accounts for 27.2% of the total energy consumption in Europe, whereas road transport is responsible for 22.2% of greenhouse gas emissions [5]. However, since both domains are closely interconnected with human activity and behavioral patterns, finding efficient ways to reduce energy consumption is a non-trivial task. At the same time, the recent rise of Artificial Intelligence (AI) presents new opportunities to develop solutions which contribute to the reduction of energy consumption. It can be argued that AI holds the key to unlock the untapped potential of energy efficiency, given its ability to process, analyze and interpret huge amounts of data in an automated way. This property of AI is very beneficial in the energy domain, where loads of data are continuously being generated. Therefore, AI can play a key role in the development of novel solutions, such as empowering end-users to better understand the intrinsic properties and patterns of their energy consumption or automating tasks in a more energy-efficient manner. This dissertation investigates the development of novel Deep Learning techniques to create tools which solve limitations in two key energy domains; the reduction of residential energy consumption through end-user empowerment via Non-Intrusive Load Monitoring and the optimization of electric vehicle (EV charging) with Deep Reinforcement Learning. The rest of this Section is devoted to establishing the motivation, and providing statistic evidence for the importance of each domain.

Residential energy consumption refers to the energy consumed by households in day-to-day activities. The analysis of energy consumption patterns in residential households is a multifaceted and complex procedure, since elements of practice theory, as well as behavioural and socioeconomic data are required to understand the observed energy patterns in depth [6]. One coherent alternative would be to provide end-users, i.e. building occupants, with insights into their consumption patterns. Relevant literature suggests that providing detailed energy consumption information to customers can lead to a total consumption decrease of up to 15% [7–9]. However, measuring the consumption of the individual appliances inside a household would require the installation of appliance-level smart meters, which bears significant installation costs, making its wide-scale adoption unrealistic. This barrier can be mitigated in a cost-effective way through the utilization of Non-Intrusive Load Monitoring, or energy disaggregation, which is the process of decomposing the aggregate household consumption onto its additive subcomponents, i.e. the consumption signals of the individual appliances. Non-intrusive Load Monitoring can, given aggregate consumption smart meter readings, provide real-time information about each appliance's consumption, thus enabling end-user empowerment. This way, end-users can recognize activities which lead to increased energy consumption and change their behavior, as well as identify whether an appliance consumes more than expected, due to required maintenance or old age, enabling its replacement with a more efficient one. Even though the potential 15% decrease in consumption may seem low as an absolute number, it can have a significant impact in meeting climate change goals due to the wide-scale nature of residential consumption.

Transport is the end-user sector with the highest global growth of $CO_2$ emissions, with an average annual growth rate of 1.7% over the last 35 years [10]. This growth showcases the necessity to move towards a low-emisssion vehicle fleet and reduce the transport sector $CO_2$ emissions if the aforementioned climate change goals are to be met. To that end, the electrification of the transport sector can be regarded as a key strategic decision, which would lead to short-term global emission mitigation and long-term net-zero transport emission achievement. However, the underlying emission reduction is based on the assumption of the decarbonization of the energy mix through investments in renewable energy generation in power networks. Even tough these investments are ongoing, the goal of a full phase-out of fossil-fuels in the energy mix has not yet matured [11]. With that in mind, it can be concluded that the replacement of internal combustion engine (ICE) vehicles by EV will provide few environmental benefits until 2030 [12]. At the same time, the increasing penetration of EV in the power grid may put additional stress on power networks, especially in high energy demand periods for EV charging, putting the capacity of power networks to test. These issues highlight the need of alternative solutions for EV charging, utilizing distributed energy resources, such as domestic solar photovoltaic (PV) panels, to ensure that a) the EV are fueled with clean, renewable energy and b) that the EV are not a burden to the power network. EV charging optimization can be considered a promising technology which, depending on certain criteria, adjusts the charging schedule of an EV in real time to achieve specific goals.

## 1.1 Research Questions

The rise of Artificial Intelligence and Deep Learning has significantly improved NILM model capabilities. Over the recent years, a lot of research effort has been invested in creating deep learning architectures for NILM. Approaches relying on convolutional neural networks (CNN), Recurrent Neural Networks (RNN) including Long-Short Term Memory Networks (LSTM) and Gated Recurrent Units (GRU), autoencoders, and even Generative Adversarial Networks (GAN) have been proposed in the literature. These advancements have shown promising results in disaggregating appliance usage from aggregate energy data, presenting new opportunities for enhancing energy efficiency and reducing household energy consumption.

However, despite these advancements, many challenges remain unsolved. Most works focus on the development of prototype solutions in a controlled environment, usually neglecting real-world situation-specific characteristics, such as varying appliance usage patterns across households, frequency of usage, or changes in user behavior over time. The majority of research also operates under the assumption of abundant, well-annotated data for training models, which is rarely the case in practical applications. This raises concerns about the scalability and generalizability of these models beyond controlled environments.

Consequently, topics such as data efficiency, model deployment on edge devices, post-deployment performance monitoring, and efficient model fine-tuning/retraining have rarely been investigated. Addressing these areas is critical for bringing NILM solutions closer to real-world applications where resources are limited, and user needs are dynamic. Furthermore, the deployment of NILM models on edge devices, which can process data locally, is becoming increasingly important as the demand for energy-efficient, privacy preserving monitoring systems grows.

This dissertation draws particular interest from such NILM problems, which led to the formulation of the following research questions:

- How can we develop a novel deep learning NILM architecture, which will be adaptable, and its performance will not heavily rely on data pre-processing and balancing?
- How can the deployment of deep learning NILM models to edge devices meet the optimal equilibrium between performance loss and model compression/optimization?
- How can we monitor a deployed NILM model, track its performance and fine-tune/retrain it efficiently to improve its capabilities in low-performing samples?

At the same time, EV charging optimization is a research topic that has drawn significant interest over the past 15 years. Older approaches utilized model-based methodologies, often relying on mathematical optimization or rule-based systems. While these methods proved to be effective in controlled scenarios, they often lacked adaptability to dynamic environments. More recent approaches leverage Reinforcement Learning and Deep Reinforcement Learning due to their better generalization capabilities and ability to learn complex strategies from data. These techniques offer promising solutions for real-time decision-making, where the system can continuously adapt based on changing conditions, such as fluctuating energy prices or varying demand.

However, most works focus primarily on cost minimization, often overlooking broader implications, particularly the potential contribution of EV charging optimization to environmental sustainability.

The integration of renewable energy sources, such as solar or wind, into the charging optimization process remains underexplored, despite the growing need to reduce the carbon footprint of transportation. Furthermore, the impact of EV charging on the stability of power grids, especially during peak demand periods, is an emerging concern. Without significant upgrades to power networks, optimizing EV charging to reduce grid stress will become a critical area of research in the near future. Additionally, optimization is typically conducted on an individual level, focusing on maximizing the efficiency of a single EV's charging process. This neglects the potential of community-oriented optimization, where multiple EVs share resources like solar panels or battery storage to maximize efficiency and sustainability. Such approaches could lead to better local energy consumption, reducing dependency on the grid and promoting a more resilient, decentralized energy system.

To that end, this dissertation employs Deep Reinforcement Learning techniques to optimize EV charging and answer the following research questions:

- How can an EV charging schedule be shaped in real time to take into account external influencing parameters and, at the same time, prioritize the utilization of clean, renewable energy?
- How can an EV charging schedule be utilized to promote local consumption of distributed energy sources to reduce grid stress?
- How can multiple EV charging schedules be optimized concurently to maximize the utilization of shared resources and achieve a better optimization result?

## 1.2 Contribution and Originality

Building on the motivation and challenges highlighted previously, this dissertation addresses key gaps in the fields of Non-Intrusive Load Monitoring and EV charging optimization. The research questions focus on developing scalable and adaptable solutions for energy monitoring and management, particularly in real-world, resource-constrained environments.

In NILM, existing approaches often fall short when dealing with real-world data, which is often unbalanced and requires extensive pre-processing. These limitations hinder the deployment of models that can generalize well across various households and appliances. Similarly, in EV charging optimization, current methods predominantly focus on individual vehicles and cost reduction, neglecting the broader potential for environmental impact and the benefits of community-based energy sharing.

In line with theaforementioned research questions, this dissertation has made significant contributions in both research areas, which are briefly presented below:

- Development of a novel adaptable deep learning architecture for Non-Intrusive Load Monitoring, which overcomes the limitations of existing sequential models, and operates with minimal data pre-processing.The developed architecture is based on state-of-the-art Deep Learning building blocks, such as Transformers, and utilizes attention mechanisms to accurately estimate the power consumption of the individual appliance.

- Design and development of an edge-deployment framework for NILM models, which will take model performance into account during model optimization/compression, to achieve minimal performance loss and limit the required deployment computational resources.
- Formulation and implementation of a performance-tracking and continual learning paradigm for NILM, which is responsible for post-deployment model monitoring. Utilizing a novel contextual importance sampling mechanism, the approach is able to efficiently assess which input samples are most beneficial for model fine-tuning, thus minimizing the re-training computational effort and overhead.
- Design and implementation of an EV charging optimization method, based on Deep Reinforcement Learning, which prioritizes the utilization of solar energy for EV charging.
- Development of an EV charging optimization scheme with Deep Reinforcement Learning, which promotes self-consumption of distributed energy sources to minimize both the charging cost and the network stress by proposing alternative load scheduling.
- Extension of the previous approach to multi-agent optimization, where a community of people supposedly commonly invest in a solar panel, meant to be used as a communal resource. Taking the community-oriented approach leads to better optimization performance, both on the utilization of the shared resources, as well as to the individual optimization criteria.

## 1.3 Relevance to Surveying and Geoinformatics Engineering

Surveying and Geoinformatics Engineering plays a crucial role in the development of digital maps, spatial databases, and geospatial analysis tools. Such technologies play a crucial role in the provision of information related to urban planning, disaster management, as well as resource allocation. They can contribute to digital twins, environmental monitoring, and the broader development of smart cities. A key aspect of these advancements is their ability to support policy shaping and decision-making processes. Within this context, the integration of AI techniques for Non-Intrusive Load Monitoring (NILM) and smart Electric Vehicle (EV) charging aligns well with the objectives of the discipline.

NILM enables real-time energy consumption monitoring in buildings with minimal investment costs. This capability has significant implications for geospatial analysis and smart city development. By integrating NILM data into digital maps, urban planners can gain valuable insights into energy consumption behavior. Such data can help identify high-energy-consuming buildings, particularly those with excessive heating or cooling demands, indicating a need for renovation or improved insulation. Additionally, by mapping consumption patterns across different neighborhoods, authorities can identify areas that require energy efficiency interventions, such as improved building codes or incentives for energy-saving technologies. Furthermore, the collected energy data can serve as a foundation for policy-making, allowing city authorities to design more effective energy efficiency programs, optimize power distribution networks, and reduce overall energy waste.

From an environmental monitoring perspective, NILM plays a role in reducing carbon footprints by helping to optimize electricity usage. Understanding how different types of buildings consume energy can provide critical information for sustainability assessments. In disaster management scenarios, real-time NILM data can help detect power outages or excessive consumption spikes that may indicate infrastructure failures. This contributes to more efficient emergency response and infrastructure resilience planning.

Similarly, smart EV charging contributes to geospatial analysis and urban modeling by optimizing the placement of charging stations and integrating renewable energy sources. Through satellite imagery and spatial data analysis, geoinformatics tools can be used to determine the most effective locations for solar panels, ensuring that renewable energy generation aligns with local demand. AI-powered geospatial models can also analyze traffic patterns, land use, and population density to predict where EV charging demand will be highest. This allows for the strategic placement of charging infrastructure, ensuring accessibility and efficiency in energy distribution while reducing urban congestion related to charging station availability.

Another important aspect of smart EV charging is its role in grid management and energy load balancing. By integrating geospatial data with AI-driven forecasting, urban planners and utility providers can predict peak charging times and distribute energy accordingly. This prevents grid overloads and allows for a smoother transition to widespread EV adoption. Moreover, mapping energy demand for EV charging stations against existing power infrastructure can help identify locations that require grid upgrades or additional renewable energy investments.

Overall, the integration of AI-driven NILM and smart EV charging systems into Surveying and Geoinformatics Engineering enhances urban sustainability and energy management. By leveraging geospatial data and AI, these technologies can significantly contribute to more efficient urban planning, resource allocation, and policy development, reinforcing the essential role of geoinformatics in modern smart city initiatives. The combination of NILM and smart EV charging enables a more dynamic understanding of energy consumption patterns, facilitates renewable energy integration, and supports the creation of resilient urban infrastructure. In doing so, these technologies bridge the gap between geospatial engineering and real-world energy efficiency solutions, highlighting the growing synergy between AI, geoinformatics, and sustainable urban development.

## 1.4 Structure of the Dissertation

The dissertation is divided in three parts and nine chapters in total. The first part (Chapters 1, 2) aims to present the motivation behind this dissertation, presenting the underlying research objectives, as well as a short description of its main contributions. In addition, a detailed problem formulation is provided. The next two parts consist of three chapters each, and encompass problems and their solution related to the aforementioned research objectives. Part 2 (Chapters 3, 4, 5) deals with Non-Intrusive Load Monitoring, whereas Part 3 (Chapters 6, 7, 8) is devoted to EV charging optimization. The dissertation is concluded with a chapter which summarizes the key findings, as well as future research directions. Below, a short description of each chapter is provided.

(i) **Chapter 2** presents a short mathematical problem formulation for the two energy application scenarios tackled, to lay the necessary mathematical foundations for the better comprehension of the next chapters.

(ii) **Chapter 3** describes a Transformer-based sequence-to-sequence architecture for Non-Intrusive Load Monitoring, which exceeds the limitation of recurrent models, an approach widely used in sequential modeling. The described approach, called ELECTRIcity, utilizes Transformer layers to accurately estimate the power signal of domestic appliances by relying entirely on attention mechanisms to extract global dependencies between the aggregate and the domestic appliance signals. Another additive value of the proposed model is that ELECTRIcity works with minimal dataset pre-processing and without requiring data balancing. Furthermore, ELECTRIcity introduces an efficient training routine compared to other traditional transformer-based architectures. According to this routine, ELECTRIcity splits model training into unsupervised pre-training and downstream task fine-tuning, which yields performance increases in both predictive accuracy and training time decrease. Experimental results indicate ELECTRIcityfls superiority compared to several state-of-the-art methods.

(iii) **Chapter 4** presents a NILM model optimization framework for edge deployment. Nowadays, the NILM research focus is shifted towards practical NILM applications, such as edge deployment, to accelerate the transition towards a greener energy future. NILM applications at the edge eliminate privacy concerns and data transmission-related problems. However, edge resource restrictions pose additional challenges to NILM. NILM approaches are usually not designed to run on edge devices with limited computational capacity, and therefore model optimization is required for better resource management. Recent works have started investigating NILM model optimization, but they utilize compression approaches arbitrarily without considering the trade-off between model performance and computational cost. The proposed edge optimization engine optimizes a NILM model for edge deployment depending on the edge devicefls limitations and includes a novel performance-aware algorithm to reduce the modelfls computational complexity. We validate our methodology on three edge application scenarios for four domestic appliances and four model architectures. Experimental results demonstrate that the proposed optimization approach can lead up to a 36.3% average reduction of model computational complexity and a 75% reduction of storage requirements.

(iv) **Chapter 5** introduces ContiNILM, a continual learning scheme for Non-Intrusive Load Monitoring. Continual adaptability is an important aspect of practical NILM applications, as they usually require frequent post-deployment maintenance to deal with non-stationary appliancesfl data distributions. However, in most approaches in the literature, the trained deep learning model weights remain static, potentially neglecting valuable information that can be used for further model training. The methodology presented alleviates the aforementioned limitation by building robust models that track environmental/seasonal alterations with direct impact on several appliancesfl operation. In our approach, model weights do not remain static, but utilize additional training data to further improve the disaggregation performance. A novel mechanism is proposed that determines whether new incoming samples would be beneficial for model training and alleviates the risk of ”forgetting” previously learned knowledge. Experimental results demonstrate the efficiency of the proposed approach.

(v) **Chapter 6** aims to address the challenge of domestic EV charging while prioritizing clean, solar energy consumption. Power sector decarbonization plays a vital role in the upcoming energy transition towards a more sustainable future. Decentralized energy resources, such as Electric Vehicles (EV) and solar photovoltaic systems (PV), are continuously integrated in residential power systems, increasing the risk of bottlenecks in power distribution networks. Time-of-Use tariffs are treated as a price-based Demand Response (DR) mechanism that can incentivize end-users to optimally shift EV charging load in hours of high solar PV generation with the use of Deep Reinforcement Learning (DRL). Historical measurements from the Pecan Street dataset are analyzed to shape a flexibility potential reward to describe end-user charging preferences. Experimental results show that the proposed DQN EV optimal charging policy is able to reduce electricity bills by an average 11.5% by achieving an average solar power utilization of 88.4%.

(vi) **Chapter 7** introduces a residential smart EV charging framework that prioritizes solar photovoltaic (PV) power self-consumption to accelerate the transition to a carbon neutral passenger vehicle fleet. Decarbonization of the transport sector is a major challenge in the transition towards net-zero emissions. Even though the penetration of electric vehicles (EV) in the passenger vehicle fleet is increasing, the energy mix is not yet dominated by renewables. This leads to the utilization of fossil-based power generation for EV charging, especially during peak hours. Our approach employs N-Step Deep Reinforcement Learning to charge the EV with clean energy from a PV, without neglecting other major factors that influence end usersfl behavior, such as electricity cost or EV charging tendencies. Historical smart-meter data from the Pecan Street dataset on total consumption, EV demand and solar generation have been utilized as input features to train the Deep RL method so that it can decide whether to charge or not the EV on a real-time basis, without the need for foresight of future observations. Experimental results on six residential houses validate that, compared to uncontrolled EV charging, the proposed method can increase the average self-consumption of solar energy for EV charging by 19.66%, as well as reduce network stress by 7% and electricity bill by 10.3%

(vii) **Chapter 8** proposes a community-driven smart EV charging optimization scheme with Multi-Agent Deep Reinforcement Learning. EV charging optimization with utilization of decentralized renewable energy sources can be seen as a promising tool towards domestic EV fleet decarbonization. However, optimization is usually conducted on an individual level, and community-driven approaches with shared resources are heavily underexplored. Compared to existing single-agent approaches, the employment of a multi-agent one allows for concurrent EV charging optimization for each household within a community. The optimization system can be deployed in a centralized, shared, energy management system, while utilizing a community-owned solar photovoltaic (PV) panel. Our approach results in reduced cost barriers for domestic EV owners that desire to optimize their charging profile, as it eliminates the investment on an individual PV panel and energy management system. Experimental results on the Pecan Street dataset validate the effectiveness of our approach compared to individual household optimization, resulting in cost savings up to 17.65%, increase in PV power utilization of up to 133.10%, as well as network stress reduction of up to 18.75%.

(viii) **Chapter 9** summarizes the key findings of this dissertation as well as outlines potential future research directions.

# Chapter 2

# Problem Formulation

In this Chapter, we provide a structured mathematical formulation of two key energy challenges addressed in this dissertation: Non-Intrusive Load Monitoring (NILM) and Electric Vehicle (EV) Charging Optimization. These problems are both rooted in the need for efficient energy management but require distinct approaches due to the unique complexities of each domain.

The first Section addresses the problem of NILM, an alternative to appliance load monitoring that circumvents the cost-prohibitive nature of individual appliance metering by inferring appliance-level power consumption from aggregate household measurements. Here, NILM is formulated as a Blind Source Separation (BSS) problem, requiring the decomposition of a mixed household power signal into the contributions of individual appliances. This decomposition is especially challenging due to the unknown number of active appliances and the non-invertible nature of the mixing matrix, resulting in a highly underdetermined problem.

The second Section formulates the EV Charging Optimization problem, which has become essential due to the rising electricity demand from the increased adoption of EVs. This problem involves optimizing charging schedules to minimize grid stress and reduce costs by leveraging data-driven decisions in real time. EV Charging Optimization is modeled as a Markov Decision Process (MDP), where sequential decisions are made based on observed variables over a defined time horizon. Recent advancements in Deep Reinforcement Learning (DRL) offer promising solutions, enabling adaptive and intelligent scheduling strategies that can respond to fluctuations in real-time data.

In presenting these formulations, this Chapter lays the mathematical groundwork necessary for the subsequent exploration and solution of each problem. By defining the structure and constraints of NILM and EV Charging Optimization, we create a foundation for the advanced methodologies applied in the Chapters that follow.

## 2.1 Non-Intrusive Load Monitoring

Appliance Load Monitoring is defined as the continuous observation of an appliance's power consumption. Such an approach can generate invaluable data to a variety of stakeholders, such as home owners/consumers, energy auditors, public policy makers etc. The most accurate way to achieve that would be to install smart meters for each appliance, which would record and transmit the electrical loads to a location with computational and storage capacity. However, this approach entails significant financial costs, thus making its wide-scale application to residential buildings improbable. A viable alternative would be Non-Intrusive Load Monitoring (NILM) [13], which can be defined as the inference of appliance level power consmption from the aggregate household smart meter readings. A visual representation of NILM can be seen in Figure 2.1.

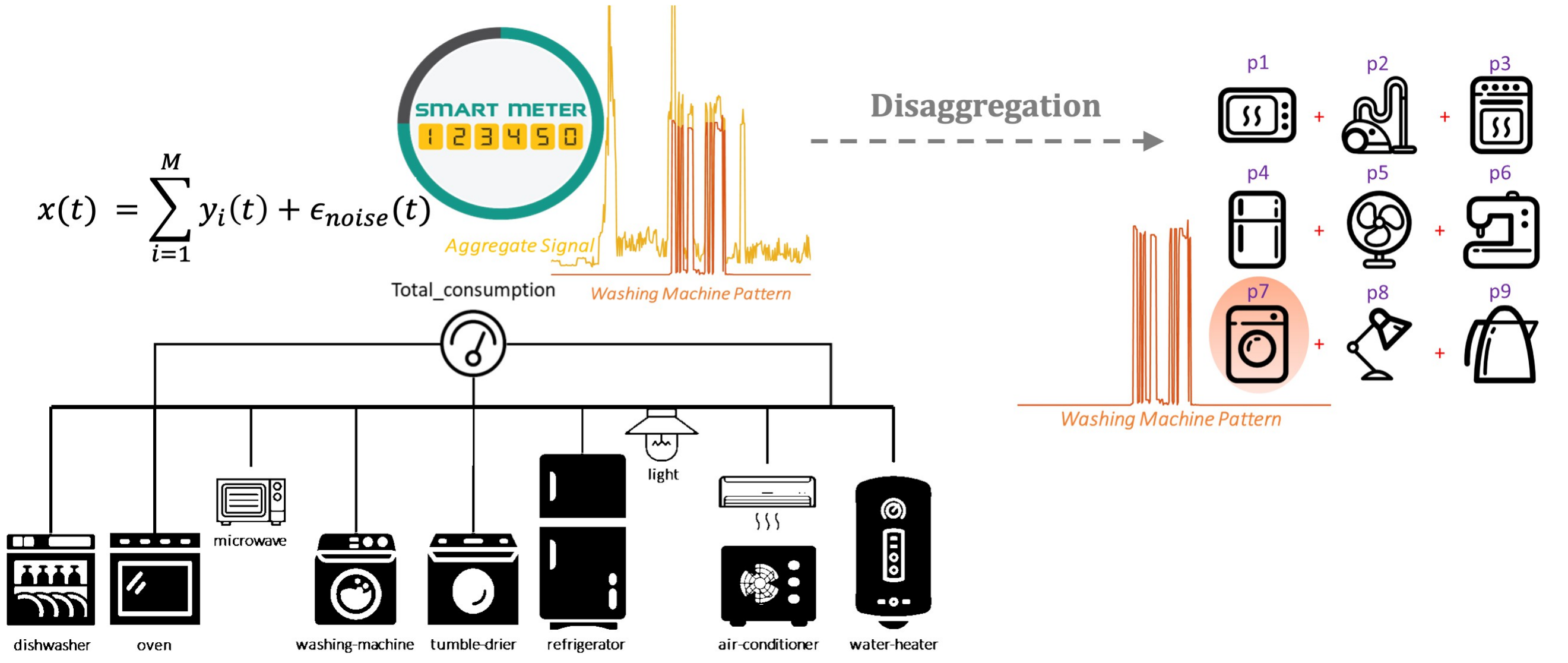

Figure 2.1 Visual Representation of Non-Intrusive Load Monitoring

.

Mathematically, NILM is a special formulation of the Blind Source Separation (BSS) problem, a family of signal processing problems where the goal is to extract a source signal from a mixed, noisy signal, without auxiliary information about the source signal or the mixing process. To describe BSS, let $x \rightarrow R^{1\uparrow n}$ denote a time series window with $n$ samples. Let us also assume that the mixed signal $x$ consists of at most $M$ source signals. In that case, BSS aims to decompose a mixed signal $x$ into is additive subcomponents $y_1, y_2, \ldots, y_M$, and can be written as:

$$x = WY \qquad W \rightarrow R^{1\uparrow M}, Y \rightarrow R^{M\uparrow n}$$
$$\begin{bmatrix} x_1 \\ x_2 \\ \vdots \\ x_n \end{bmatrix} = W \begin{bmatrix} y_{1,1} & y_{2,1} & \cdots & y_{m,1} & \cdots & y_{M,1} \\ y_{1,2} & y_{2,2} & \cdots & y_{m,2} & \cdots & y_{M,2} \\ \vdots & \vdots & & \vdots & & \vdots \\ y_{1,n} & y_{2,n} & \cdots & y_{m,n} & \cdots & y_{M,n} \end{bmatrix} \tag{2.1}$$

where $W$ denotes a mixing matrix. Solving Equation 2.1 is not trivial, especially in the case of NILM. First, the number of additive sub-components, which in the case of NILM denote the household's active appliances, is not known, thus the dimensionality of the problem cannot be determined. In addition, the mixing matrix $W$ may not be invertible, and even if it is, there may be infinitely many possible solutions that would perfectly reconstruct the mixed signal (i.e. the aggregate household consumption), making the problem non-deterministic. Therefore, a potential avenue for NILM would either need auxiliary information on the mixing matrix, or exploring data-driven approaches to approximate the solution.

To advance Equation 2.1 to correctly formulate the NILM problem, let $M$ be the number of household appliances and $i$ be the index referring to the $i$-th appliance ($i = 1, \ldots, M$) [14]. The aggregate power consumption $x$ at a given time $t$ is the sum of the power consumption of the individual appliances $M$, denoted by $y_i \downarrow i = 1, \ldots, M$. Thus, in a NILM framework [13], the total power consumption $x$ at a given time $t$ is:

$$x(t) = \sum_{i=1}^{M} y_i(t) + s_{noise}(t) \tag{2.2}$$

where $s_{noise}$ describes a noise term. Our goal is to solve the inverse problem and estimate the appliance consumption patterns $y_i$, given the aggregate power signal $x$. It is evident that NILM is formulated as a blind-source separation problem that is highly undetermined, since there are infinite combinations of $y_i$ that reconstruct $x$.

## 2.2 EV charging optimization

The increasing penetration of electric vehicles in the transport sector has created a rising electricity demand to support the growing number of EV on the road. Since this rise puts pressure on the power grid, the optimization of the EV charging schedule has received increasing research interest. EV charging optimization can be defined as the use of intelligence and connectivity to mange when and how an EV, plugged into a smart charger, will receive power for charging based on specific requirements. EV charging optimization can be seen as a promising, if not imperative way to minimize charging cost, as well as reduce stress on the power grid and streamline grid management. More and more EV chargers are being installed, both in residential households as well as public spaces such as charging lots, workplaces, retail centers/shopping malls etc.

EV charging optimization is based on time-series data and can be viewed as a sequential decision making problem in a specified time interval. At each timestep of the interval, the goal is to decide whether to charge the EV or not, given a set of observation variables. Therefore, EV charging optimization can be modeled as a Markov Decision Process (MDP), a popular modeling approach in signal processing to represent stochastic control problems.

MDP involve decision making according to specific observations, where each decision may or may not be influenced by previous ones. To proceed with a mathematical formulation of the problem, a mathematical definition of the related entities is required. Therefore, an MDP can secribed by the sets ($S$, $A$, $R$, $P$), where each set corresponds to a different core mechanic of the procedure. $S$ denotes the State space, which is the set of all possible observable conditions that may be encountered during the optimization, whereas $A$ encapsulates all the possible actions of the variable under optimization. $P$ signifies the transition probabilities between states given a certain action, and $R$ describes the potential optimizaton reward/penalty in each transition.

To calculate the optimal policy, i.e. the optimal action sequence, two variables are required. The value variable $V(s)$ contains the sum of all possible transition rewards from the current state to the next, taking into consideration the value of all potential future states, whereas the policy variable $u(s)$ captures the optimization trajectory in terms of actions. Both variables are described in Equations 2.3 and 2.4 respectively.

$$V(s) = \sum_{s'} P_{\pi(s)}(s, s')[R_{\pi(s)}(s, s') + \gamma V(s')] \quad (2.3)$$

$$\pi(s) = \operatorname*{argmax}_{a \in A} \{P_a(s, s')[R_a(s, s') + \gamma V(s')]\} \quad (2.4)$$

Earlier works on EV charging optimization used MDP as the modeling based and tried to solve the optimization with model-based approaches, such as Linear Programming or simulation-based methods, to come up with the optimal charging pattern. However, two major limitations arise. Full knowledge of the optimization horizon is needed (i.e. the complete state space $S$ must be known), meaning that either they require forecasting models for the input variables, potentially introducing error accumulation in the optimization process, or they must be applied in day-ahead optimization scenarios. In both cases, real-time adaptation to fluctuations in the input variables cannot be taken into consideration during the optimization, given the increasing penetration of connected smart devices in our lives, wastes a significant performance increase potential. To that end, Deep Reinforcement Learning (DRL) has been gaining in popularity, due to its real-time data-driven nature. Even though the optimization algorithm differs, MDP still serve as the mathematical cornerstone for the optimization process. In the subsequent Chapters of this dissertation focusing on EV charging optimization, DRL algorithms are implemented. The state space consists of the input variables observed by the model to take an action, whereas the action space is defined as a binary set (charge/remain idle). Finally, the criteria to drive the optimization are embedded into reward functions, which reward or penalize the model. Through the iterative observation of state-action-reward pairs, the model learns to converge to the desired behavior.

# Part II

# Non-Intrusive Load Monitoring

# Chapter 3

# ELECTRIcity: An Efficient Transformer for Non-Intrusive Load Monitoring

## 3.1 Introduction

Non-Intrusive Load Monitoring (NILM), or energy disaggregation, is an efficient and cost-effective framework to reduce energy consumption [15]. Energy (or Electricity) disaggregation algorithms aim to infer the consumption pattern of domestic appliances by only analyzing the aggregated household consumption signal. This process can be viewed as the decomposition of the aggregate power signal of a household into its additive sub-components, i.e. power signals of each domestic appliance. NILM presents several significant challenges that need to be overcome. The power signal exhibits severe non-linearity, since the temporal periodicity of the individual appliance activation depends on contextual characteristics [15–17], i.e., geographic and socioeconomic parameters or even residents' habits. This leads to diverse energy consumption patterns in households. Therefore, it is challenging to implement models with good generalization ability that achieve high performance when tested on unseen houses. Other notable challenges include long-range temporal dependencies in appliance activations, as well as dataset imbalance. Many appliances may not be turned on every day, and operate for a small period of time, resulting in their activation function being dominated by zeros. Various NILM approaches have been proposed in the literature. Some of the most successful exploit deep learning structures, such as recurrent [18–20] or convolutional neural networks (CNN) [21–23], to extract information about individual appliance consumption. Even though these techniques have good performance in energy disaggregation tasks, there are some limitations and challenges.

- **Challenge 1:** These algorithms are easy to be trapped in the assumption that adjacent events in a sequence are dependent, while, as long as time passes, the interactions between remote events are faded.
- **Challenge 2:** Long Short-term Memory (LSTM) [15] models have a memory mechanism, that decides the worth-remembering information from the useless one, every time a new state is entered in the sequence. Thus, local dependencies are more powerful than global ones, and old -or infrequent- events are faded in case they do not appear regularly. Data balancing is a necessary prerequisite in these approaches, to maintain important information.

- **Challenge 3:** Temporal Convolutional Neural Network (CNN) architectures capture long-range temporal dependencies in time series, with the necessary adaptations that include residual connections and dilated convolutions, but require significant model depth to catch long-range dependencies.

In this Chapter, we introduce ELECTRicity [24], a Transformer-based framework for solving the NILM problem. Transformers do not sequentially process data. Instead, they process the entire sequence of data, understand the significance of each part of the input sequence and assign importance weights accordingly, using attention mechanisms, to learn global dependencies in the sequence. Even though Transformer architectures seem suitable for NILM challenges, their applicability is limited [25] due to lack of efficiency and computational complexity issues. To fully exploit the capabilities of the Transformer-based architecture, ELECTRicity consists of two parts: ***(i) the pre-training process***, which is an unsupervised pre-training process that requires only the aggregated power signal as input, ***(ii) the training process***, in which the pre-trained Transformer model is fine-tuned in a supervised way to predict the electrical consumption of a specific domestic appliance. During the pre-training step, our model consists of a Transformer-based **generator** and a **discriminator**, that cooperate to increase model performance, while using few computational resources. This process lead to a novel, efficient NILM framework, that has the comparative advantages summarized below:

- **ELECTRIcity is capable of learning long-range temporal dependencies.** In seq2seq models, learning temporal dependencies is a demanding task, and often the model forgets the first part, once it completes processing the whole sequence input. ELECTRIcity utilizes attention mechanisms and identifies complex dependencies between input sequence elements regardless of their position.

- **ELECTRIcity can handle imbalanced datasets.** Our work demonstrates that combining the unsupervised pre-training process with downstream task fine-tuning, offers a practical solution for NILM, and handles successfully imbalanced datasets. This is a comparative advantage against the existing state-of-the-art NILM works which, in most cases, require data balancing to achieve good performance.

- **ELECTRIcity is an efficient and fast Transformer.** ELECTRIcity introduces a computationally efficient unsupervised pre-training process through the combined use of a generator and a discriminator. This leads to a significant training time decrease without affecting model performance compared to traditional Transformer architectures.

## Related Work

Deep learning has achieved great success in domains such as computer vision and natural language processing (NLP) [26]. Since 2015, deep neural networks (DNN) have transversed into NILM and the number of the proposed DNN approaches has increased rapidly [27].

Recurrent neural networks (RNN), LSTM, bidirectional LSTM (BiLSTM), and gated recurrent unit (GRU) networks have been firmly established as state-of-the-art approaches in NILM [28]. These techniques take advantage of recurrent mechanisms to identify temporal patterns in power consumption sequences. Recurrent layers utilize feedback connections to capture temporal information in 'memory' and are well suited to sequential power signal data and energy disaggregation tasks. However, RNN lacks the ability to learn long-range temporal dependencies due to the vanishing gradient problem, as the loss function decays exponentially with time [29].

LSTMs rely on memory cells that employ forget, input, and output gates to memorize long-term temporal dependencies [18–20]. Even though LSTMs are successful in several time-series-related tasks, their elaborate gating mechanism may result in increased model complexity. At the same time, computational efficiency is a crucial issue for recurrence-based models and alternative architectures, such as GRU networks, have been developed to alleviate this limitation. These have been widely proposed in NILM [30].

CNN-based architectures have made great progress towards capturing long-range temporal dependencies in time series [22, 23, 31], but require significant model depth to expand their receptive field. Various works have proposed CNN-based solutions that leverage emerging advancements like, for instance, causal or temporal 1D-CNN to address NILM-related challenges [21]. These networks combine causal, dilated convolutions and other model modification techniques, such as residual connections or weights normalization to limit computational complexity without affecting the model's performance. Alternative approaches suggest hybrid CNN-RNN architectures, that benefit from the advantages of both convolutional and recurrent layers. Representative examples of how these hybrid structures can be applied to NILM are [32, 33].

Sequence-to-sequence (seq2seq) models have been widely used for energy disaggregation [27, 28, 34, 35]. These models are particularly successful at machine translation [36], where word sequences are translated from one language to another. By analogy, in the energy disaggregation field, the aggregated sequence is translated through a seq2seq model to the power consumption of a specific domestic appliance. Denoising autoencoders are commonly considered the current state-of-the-art deep learning method for NILM [27, 37]. Apart from seq2seq models, sequence-to-point (seq2point) [38, 39] and sequence-to-subsequence (seq2subseq) [40, 41] methods have also been utilized.

Most of the aforementioned studies deploy a pre-processing strategy to handle data balancing properly. In a NILM framework, the time interval between an appliance being switched on and off is referred to as an activation [28]. Domestic appliances, depending on their household use, may showcase from zero to several activations daily. Usually, the appliance run-time is considerably shorter compared to the time it is switched off, which leads to skewed datasets with sparse appliance activations.

Transformers [42] have rapidly emerged across a wide variety of sequence modeling tasks [43–45], due to their ability to arbitrarily and instantly access information across time, as well as their superior scaling properties compared to recurrent architectures. The main advantage of Transformers stems from the fact that they, in contrast to the aforementioned architectures, process a sequence in parallel in an order-invariant way. Techniques such as positional embeddings and attention masking are an integral part of Transformer-based methodologies [46, 47]. Original Transformers do not rely on past hidden states to capture dependencies. On the contrary, they process a sequence as a whole, mitigating the risk to lose -or 'forget'- past information. As a consequence, Transformers do not suffer from long-range dependency issues, which is the main controversy in RNN. Even though Transformer architectures seem suitable for NILM challenges, their applicability is limited [25, 48] due to efficiency and computational complexity issues.

# 3.2 Methodology

## 3.2.1 Transformer Model Fundamentals

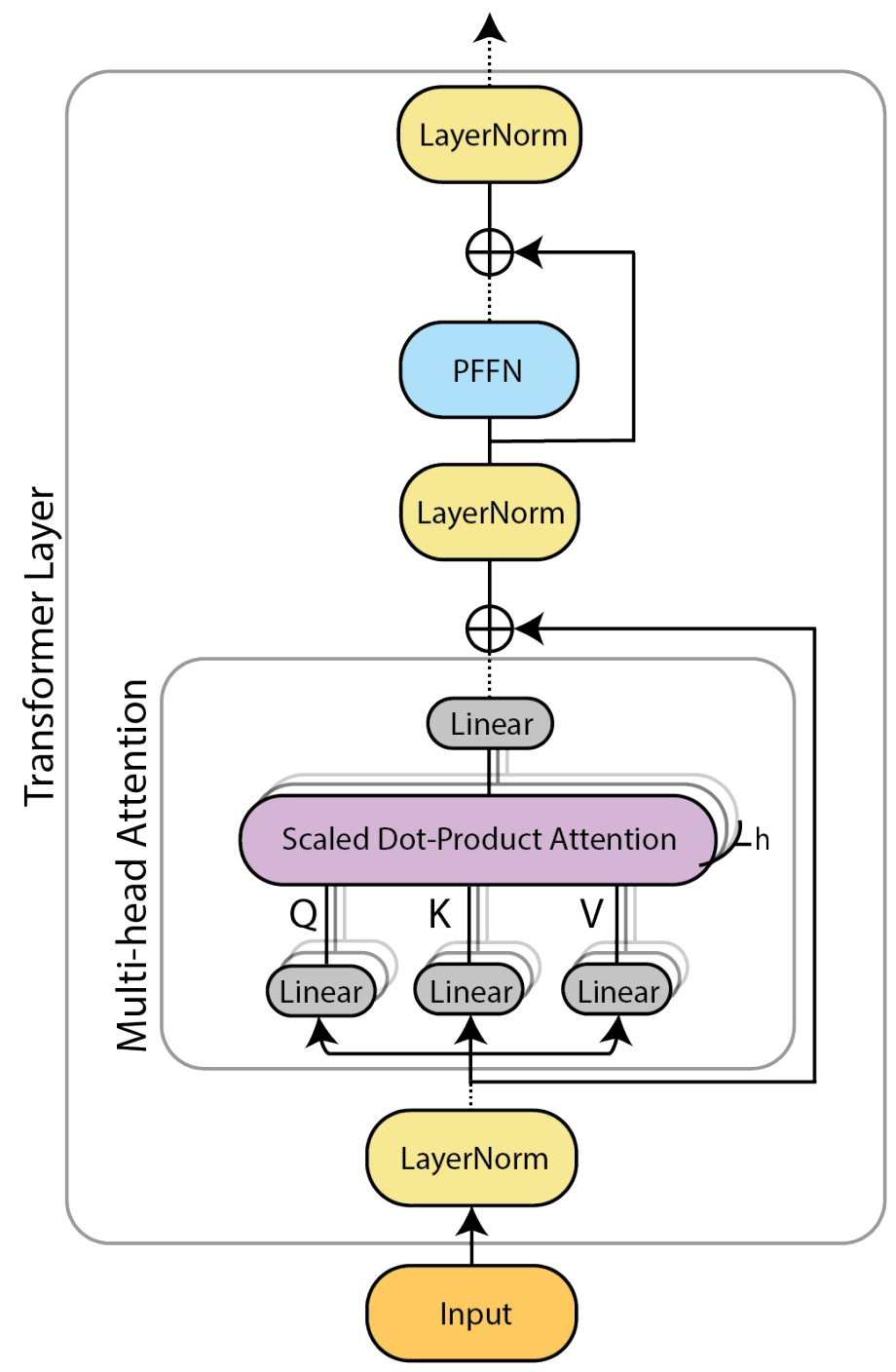

Figure 3.1 Overview of a transformer layer. Dashed lines indicate Dropout regularization

.

The Transformer model [42] consists of two major components: a multi-head attention (MHA) module and a position-wise feed-forward network (PFFN). An overview of the Transformer layer is depicted in Figure 3.1. The input signal is first normalized and fed to the multi-head attention layer, which calculates the attention scores (see Section 3.2.1.1). Then, the attention scores are normalized and passed on to a position-wise feed-forward layer (see Section 3.2.1.2). Residual connections and dropout regularization [49] are introduced to increase the stability of the model. In the following subsections, we shall introduce the two key components (MHA and PFFN) of a Transformer layer.

#### 3.2.1.1 Multi-Head Attention Mechanism

Transformers implement attention mechanism as a Query-Key-Value (QKV) model. Attention consists of a series of linear transformations that process input sequences in an order-invariant way and assign importance weights to each position in the sequence. Thus, single-head dot-product attention mechanism applies linear transformations to the input signal to form query (Q), key (K) and value (V) matrices. Let us denote the input signal as $x \rightarrow \mathsf{R}^{d_b \uparrow d_l}$, where $d_b$ is the batch size and $d_l$ the input length. The linear transformations can be formulated as matrices $W_q \rightarrow R^{d_l \uparrow d_q}$, $W_k \rightarrow \mathsf{R}^{d_l \uparrow d_k}$ and $W_v \rightarrow \mathsf{R}^{d_l \uparrow d_v}$.

$$Q = W_q^T x, \quad K = W_k^T x, \quad V = W_v^T x \tag{3.1}$$

To ease matrix computations, $W_q$, $W_k$ and $W_v$ should have the same size $d_k = d_q = d_v$. Single-head dot product attention (denoted by $A$) is then a matrix multiplication of Q, K and V after a scaling and softmax operation.

$$A(Q,K,V) = softmax(\frac{QK^T}{\sqrt{d_k}})V \tag{3.2}$$

The first term in Equation (3.2) can be viewed as the important weighting of values at all positions of the sequence. Therefore, attention can inherently understand which parts of the sequence are significant to predict the output and ignore parts that are not. This feature is particularly useful when dealing with imbalanced datasets since the respective weight for negative samples can automatically be set to a small value. Attention is an integral part of our proposed model architecture, which is illustrated in Figure 3.2.

Instead of applying a single attention function, Transformers deploy a multi-head attention mechanism. *MHA* is calculated by extending the aforementioned single-head attention mechanism to h dimensions (multiple heads) by concatenating the single-head attention outputs, followed by a linear layer.

$$MHA = Concat(A(Q_i, K_i, V_i)) \quad \downarrow i \rightarrow 1, \dots h \tag{3.3}$$

In literature, multiple single-head attention techniques have been developed (additive attention [36], multiplicative attention [50], dot-product attention [42]). The latter is the most widely used variation.

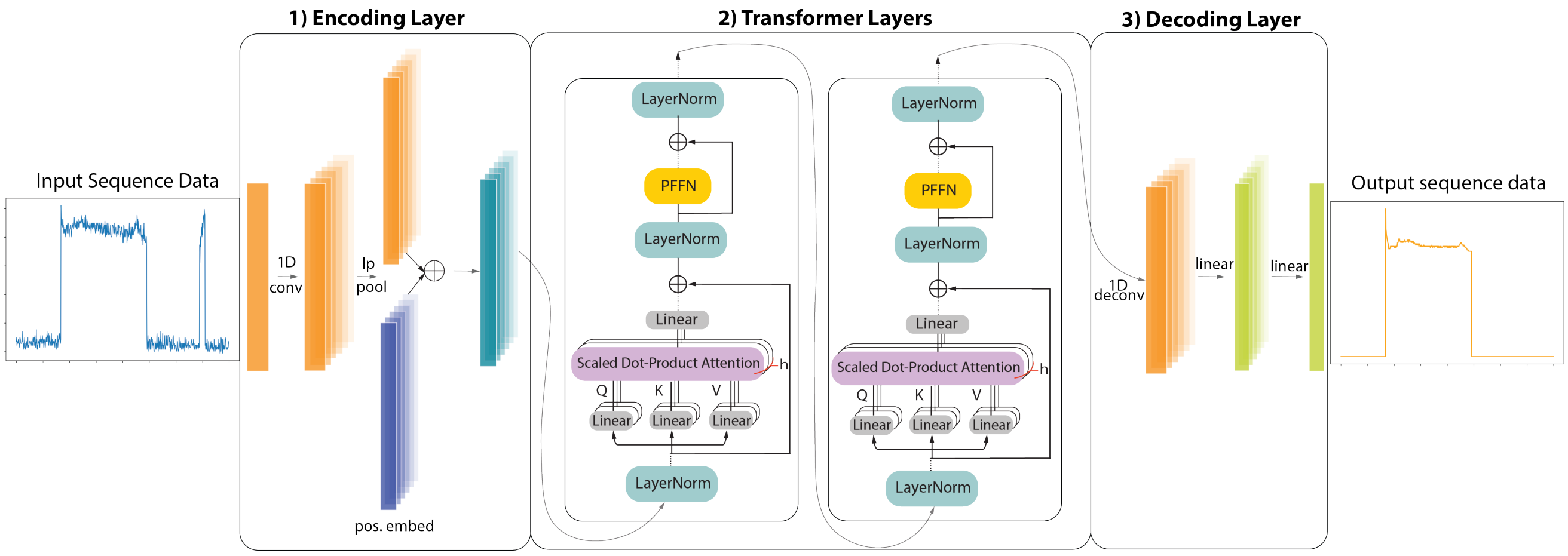

Figure 3.2 Proposed model architecture for ELECTRIcity's generator and discriminator models.

#### 3.2.1.2 Position-Wise Feed-Forward Network

The normalized attention scores are passed on to a position-wise feed-forward layer (*PFFN*), which performs linear transformations with *GELU* activation function [51]. The linear transformations are applied to each position separately and identically, meaning that the transformations use the same parameters for all positions of a sequence and different parameters from layer to layer. Let us denote the attention sub-block output as $a$ and the weight matrices and bias vectors of each linear transformation as $W_1$, $b_1$ and $W_2$, $b_2$ respectively. Then:

$$PFFN(a) = GELU(0, aW_1 + b_1)W_2 + b_2 \tag{3.4}$$

### 3.2.2 ELECTRIcity: An Efficient Transformer for NILM

ELECTRicity is an efficient model training routine for energy disaggregation. ELECTRicity splits model training into a pre-training (Section 3.2.2.1) and a training routine (Section 3.2.2.2). The pre-training step includes an unsupervised model trained with unlabeled data that uses only the aggregate signal and is applied for weight initialization to boost model performance. Here, during pre-training, we introduce the concept of generator and discriminator that is inspired by [52, 53] to improve the efficiency of the proposed model. Then, the model is fine-tuned to handle the signal of an individual appliance [45] using the discriminator model.

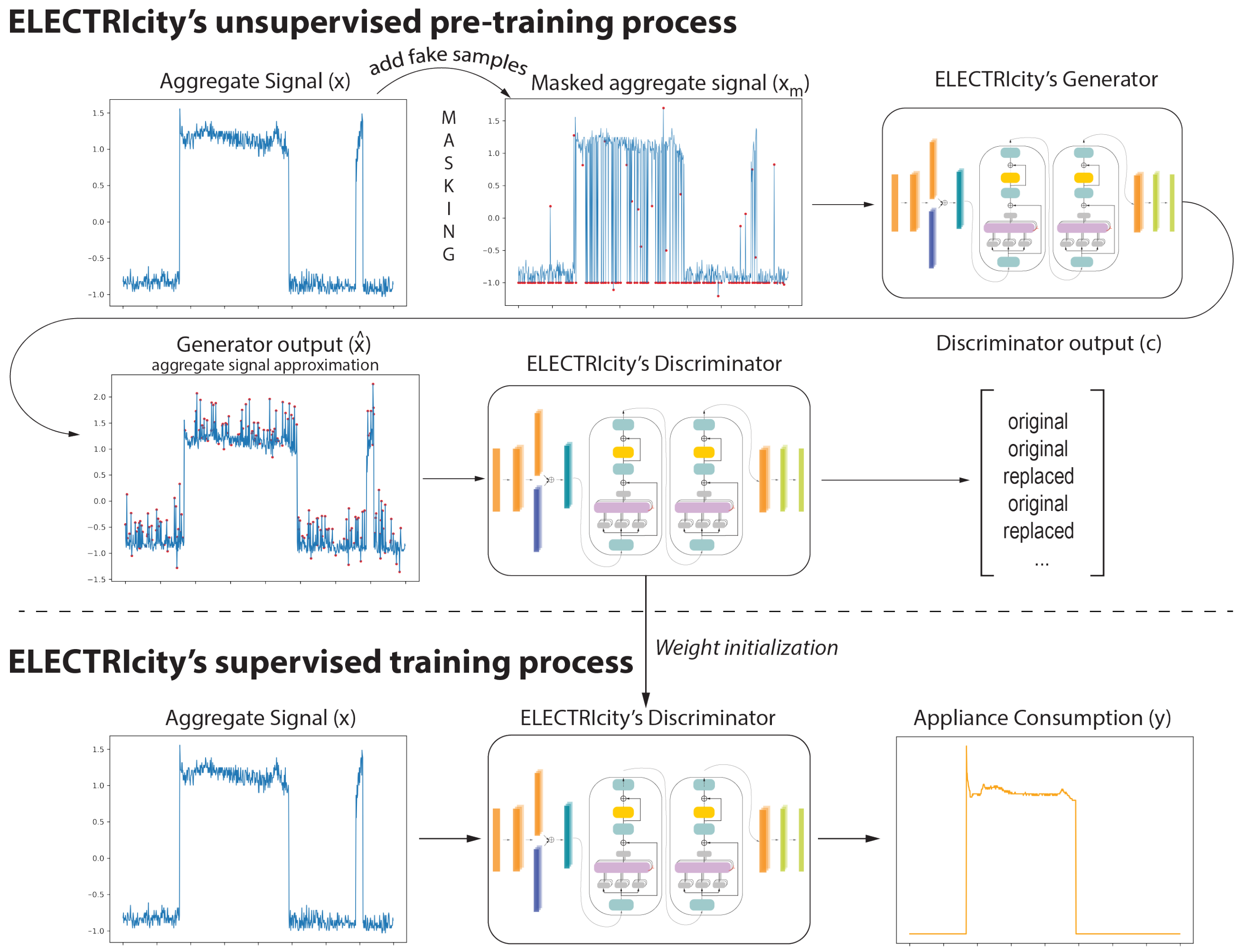

Figure 3.3 Overview of ELECTRIcity's model training routine.

#### 3.2.2.1 ELECTRIcity's Unsupervised Pre-Training Process

It is a common strategy in various Transformer architectures to utilize a model pre-training procedure [25, 45]. In such approaches, the model is pre-trained in an unsupervised way by replacing certain values from the input signal, and it is subsequently fine-tuned to solve any downstream task. Nevertheless, the loss function in such approaches [25, 45] is calculated only considering the replaced positions, meaning that only a small fraction of the data is taken into account for model training. Even though it is an interesting technique, we argue that ignoring most output values is data inefficient and that a more effective strategy could lead to higher performance.

ELECTRicity's efficient pre-training approach is illustrated in Figure 3.3. Training is split into an unsupervised pre-training mechanism and a supervised process. During pre-training, the aggregate signal is masked at random positions with fake samples and the generator tries to reconstruct the original signal. The discriminator has to distinguish which positions of the generator output were fake (replaced) and which correspond to the original signal. During training, the generator is discarded and the discriminator is fine-tuned to predict the individual appliance consumption from the aggregate signal.

Contrary to the traditional Transformer approaches described above, which use a single Transformer model, ELECTRicity consists of two twin Transformers, a generator, and a discriminator. In our approach, a fixed percentage of values in a given aggregate sequence $x \rightarrow \mathsf{R}^N$ is masked/replaced to create a masked aggregate signal $x_m$. 80% of the masked samples are replaced with a predefined value (e.g. $\searrow 1$), 10% with a random value taken from a standard Gaussian distribution, and 10% with the original input value. The generator receives the masked aggregate signal and tries to predict the original signal values at the masking positions and reconstruct the original aggregate sequence. This procedure forces the model to understand the interdependencies of the aggregate sequence without relying on labeled data. The discriminator task is then to receive the generator estimation and understand which samples correspond to the aggregate signal and which were replaced.

To account for the data inefficiency of traditional masked pre-training mechanisms [25], the generator loss function is computed only on the masked portion of the signal, whereas the discriminator loss function utilizes the whole signal. The generator loss function consists of a combination of Mean Squared Error (MSE) and Kullback-Leibler Divergence ($D_{KL}$), while the discriminator loss function implements Binary Cross-Entropy (BCE) loss. To properly formulate the loss functions, let $x \rightarrow \mathsf{R}^N$ be the aggregate signal and $\hat{x} \rightarrow \mathsf{R}^N$ the generator output. Let further $m \rightarrow \mathsf{R}^N$ be a binary mask with M masking positions and $x_m$ be the masked input signal. Finally, let $c$ be the discriminators' output. Then the pre-training loss functions $L_{gen}$ and $L_{disc}$ are:

$$
\begin{aligned}
L_{gen} &= \frac{1}{T}\sum_{i=1}^{M}(\hat{x}_i \searrow x)^2 + D_{KL}(softmax(\frac{\hat{x}}{\mathsf{c}}) \| \, softmax(\frac{x}{\mathsf{c}})) \\
L_{disc} &= \searrow \frac{1}{N}\sum_{i=1}^{N} m_i log(p(c_i)) + (1 \searrow m_i) log(1 \searrow p(c_i))
\end{aligned}
\tag{3.5}
$$

where $\mathsf{c}$ is a hyperparameter to control softmax temperature. From a dataflow perspective, the aggregate signal $x$ is masked to produce $x_m$ that is used as input to the generator. The generator output $\hat{x}$ is passed on to the discriminator which predicts which values correspond to the original aggregate signal and which were replaced. That information is captured in vector $c$. This process can be summarized as:

$$
x \Leftarrow x_m \Leftarrow generator \Leftarrow \hat{x} \Leftarrow discriminator \Leftarrow c \tag{3.6}
$$

#### 3.2.2.2 ELECTRIcity Supervised Training Process

On a high level, the pre-training process can be seen as a task-specific weight initialization technique to boost model performance. During training, the generator is discarded and the discriminator is re-trained to produce the appliance signature. Since, during training, the objective of the model changes, a different loss function is required that fits the energy disaggregation problem. The discriminator loss function is formulated in Equation (3.7).

$$
\begin{aligned}
L(y,s) = {} & \frac{1}{N}\sum_{i=1}^{N}(\hat{y}_i - y_i)^2 \\
& + D_{KL}\left(softmax\left(\frac{\hat{y}}{c}\right) \| softmax\left(\frac{y}{c}\right)\right) \\
& + \frac{1}{N}\sum_{i=1}^{N} log(1 + exp(\hat{s}_i s_i)) + \frac{k}{N}\sum_{i \in O} |\hat{y}_i - y_i|
\end{aligned}
\tag{3.7}
$$

where, k is another hyperparameter that controls the impact of the absolute error from the set $O$ of incorrectly predicted samples and timepoints when the status of the appliance is on. The loss function also considers the ground truth status of the appliance, as well as the on-off status $s$ of the predicted consumption signal. During training, the dataflow is simpler. The aggregate signal $x$ is used as input to the pre-trained discriminator, which outputs the individual appliance consumption signal $y$.

$$
x \Leftarrow discriminator \Leftarrow y \tag{3.8}
$$

## 3.3 Experimental Setup and Results

We use three open-source datasets for results comparison, UK-DALE [54], REDD [55] and Refit [56]. All datasets include electricity measurements from multiple houses and provide both low-and high-frequency data. We focus on low-frequency data and will examine 4 appliance types: (1) Appliances with distant activations and very short activation period (Kettle, Microwave) (2) Appliances with frequent, recurring activations that do not have high power consumption peaks (fridge, fridge-freezer) (3) appliances with distant activations and long activation period (Washing Machine, Dishwasher) and (4) appliances with distant activations and low power consumption peak (TV). It should be noted that UK-DALE and Refit contain significantly more data than REDD and, therefore, more appliance activations.

The data was minimally processed to meet the requirements of Table 3.1. Aggregate and appliance signals were examined at $\frac{1}{6}$ Hz frequency and time gaps shorter than 3 min were forward-filled. No measures were taken to tackle class imbalance, as we would like to test to what extent the models can perform well in real life scenarios when the appliances are turned off most of the time. In the training set, the signals were split in windows of 480 samples (48 min) with a stride of 240 samples for UK-DALE and Refit and 120 samples for REDD. The models were tested on unseen data from a house not included in the training set without window stride. More specifically, in UK-DALE houses 1,3,4 and 5 were used for training and house 2 for testing. In REDD, house 1 was kept for model evaluation and houses 2,3,4,5 and 6 were included in the training set, while in Refit houses 2,3 and 16 were used for training and the models were tested on data coming from house 5.

To validate the performance of our methodology, we utilized several state-of-the-art models that are based on different technologies. More specifically, we adopted two recurrent approaches, GRU+ and LSTM+ [57], a convolutional seq2seq network [38, 58] and a Transformer-based solution [25]. The models were trained on a Google Colab server with an Nvidia Tesla P100 GPU.

Table 3.1 Pre-processing values for the REDD (upper), UK-DALE (middle) and Refit (lower) datasets.

| Appliance | k | Max. Limit | On Thres. | Min. on Duration [s] | Min. off Duration [s] |
|---|---|---|---|---|---|
| Washer | $10^{-3}$ | 500 | 20 | 1800 | 160 |
| Microwave | 1 | 1800 | 200 | 12 | 30 |
| Dishwasher | 1 | 1200 | 10 | 1800 | 1800 |
| Fridge | $10^{-6}$ | 300 | 50 | 60 | 12 |
| Washer | $10^{-2}$ | 2500 | 20 | 1800 | 160 |
| Microwave | 1 | 3000 | 200 | 12 | 30 |
| Dishwasher | 1 | 2500 | 10 | 1800 | 1800 |
| Kettle | 1 | 3100 | 2000 | 12 | 0 |
| Washer | $10^{-2}$ | 2500 | 20 | 70 | 182 |
| TV | 1.5 | 80 | 10 | 14 | 0 |
| Fridge-Freezer | $10^{-6}$ | 1700 | 5 | 70 | 14 |

In ELECTRIcity, both generator and discriminator followed the same architecture (Figure 3.2). Feature extraction was performed with a 1D-convolutional layer with kernel size 5 and a replicate padding of 2 on both sides. Feature extraction was followed by a squared average pooling layer with kernel size and stride 2. On the decoding side, a de-convolutional layer with kernel size 4, stride 2, and padding length 1 was implemented. Both models contain 2 Transformer layers with 2 attention heads each and a hidden size $d_k$ of 64 for the generator and 64 for the discriminator. A Dropout probability of 10% has been adopted in all Dropout layers.

### 3.3.1 Performance Metrics

We recorded four widely used metrics to evaluate model performance. Mean Relative Error (*MRE*), Mean Absolute Error (*MAE*) and Mean Squared Error (*MSE*) (Equation (3.9)) were calculated using the ground truth and estimated appliance signature.

$$MRE = \frac{1}{max(Y)} \sum_{i=1}^{N} |\hat{y}_i - y_i|, \quad MAE = \frac{1}{N} \sum_{i=1}^{N} |\hat{y}_i - y_i|, \quad MSE = \frac{1}{N} \sum_{i=1}^{N} (\hat{y}_i - y_i)^2 \tag{3.9}$$

Accuracy and $F1$ score were also determined to assess if the model can properly address the class imbalance. The on-off status of the device is required and can be computed by comparing the appliance signature with the predefined requirements of Table 3.1. Accuracy is equal to the amount of correctly predicted time points over the sequence length, while $F1$-score is computed according to Equation 3.10, where $TP$ stands for True Positives, $FP$ for False positives and $FN$ for false negatives.

$$F1 = \frac{TP}{TP + \frac{1}{2}(FP + FN)} \tag{3.10}$$

*MRE*, *MAE* and *MSE* indicate the model's ability to correctly infer the individual appliance consumption levels, whereas $F1$-score indicates the model's ability to adequately detect appliance activations in imbalanced data. In our study, $F1$-score is the most important metric, as it captures the model's ability to identify appliance activations and minimize false positives.

### 3.3.2 Results

The experimental results for UK-DALE, REDD and Refit are presented in Tables 3.2, 3.3 and 3.4 respectively, while Figure 3.4 illustrates prediction examples for each examined appliance. Across all datasets, ELECTRIcity outperforms the other models in most of the appliances.

Let us now consider the kettle and microwave appliances. For these appliances, ELECTRIcity showcases a performance increase in terms of F1-Score and, in some cases, a slightly lower MAE across both datasets (UK-DALE, REDD). This can be translated to a better model capability to detect activations, while not always reaching a precise consumption prediction, which can be explained by the high data sparsity due to the timespan of each activation. In these appliances, lighter models in terms of computational complexity (CNN, LSTM+, and GRU+) reach lower performance at a lower training time. It can be argued that there is a tradeoff between performance and computational complexity during training for these appliances. It should be mentioned that ELECTRIcity and the compared models (CNN, LSTM+, and GRU+) present similar computational demands during the testing phase, while ELECTRIcity has a higher performance. A different pre-training strategy, in the sense of using an alternative masking distribution, may lead to a further performance increase. In future work, we will evaluate such approaches to investigate the full capabilities of our model.

Table 3.4 Comparison of ELECTRIcity's model performance to other techniques in the Refit dataset.

| Device | Model | MRE | MAE | MSE | Acc. | F1 | Training Time (min) |
|---|---|---|---|---|---|---|---|
| Washer | GRU+ [57] | 0.089 | 24.60 | 31,082.49 | 0.929 | 0.128 | 17.95 |
| | LSTM+ [57] | 0.098 | 25.76 | 32,958.09 | 0.920 | 0.130 | 22.86 |
| | CNN [38, 58] | 0.096 | 23.58 | 29,383.90 | 0.924 | 0.248 | 28.62 |
| | BERT4NILM [25] | **0.080** | **22.19** | 27,420.48 | **0.939** | 0.188 | 813.25 |
| | ELECTRIcity | 0.089 | 23.67 | **26,465.39** | 0.936 | **0.398** | 217.32 |
| TV | GRU+ [57] | 0.619 | 38.36 | 2539.97 | 0.410 | 0.370 | 26.51 |
| | LSTM+ [57] | 0.657 | 39.35 | 2467.22 | 0.374 | 0.357 | 32.59 |
| | CNN [38, 58] | 0.776 | 19.52 | **980.39** | 0.352 | 0.318 | 41.17 |
| | BERT4NILM [25] | 0.593 | 32.15 | 1769.43 | 0.452 | 0.381 | 1280.10 |
| | ELECTRIcity | **0.278** | **19.29** | 1375.13 | **0.740** | **0.505** | 316.10 |
| Fridge-Freezer | GRU+ [57] | 0.756 | 56.17 | 4773.60 | 0.552 | 0.710 | 17.95 |
| | LSTM+ [57] | 0.730 | 54.92 | **4567.50** | 0.551 | 0.710 | 22.86 |
| | CNN [38, 58] | 0.686 | 58.15 | 5660.37 | 0.561 | **0.713** | 28.62 |
| | BERT4NILM [25] | 0.587 | **50.16** | 5437.78 | **0.623** | 0.674 | 813.25 |
| | ELECTRIcity | **0.586** | 51.08 | 5331.71 | 0.613 | 0.668 | 217.32 |

In the second case of experiments, we have examined the fridge in UK-Dale and fridge-freezer in Refit appliances. When disaggregating the fridge appliance, ELECTRIcity is outperforming most comparison models, but falls short to BERT4NILM [25]. The activations frequency for this appliance is unique, as it exhibits a periodicity that is usually not user-controlled. The fridge turns on when the inside temperature falls under a certain threshold, and turns off when that threshold is reached. Throughout a day, we can assume that the house temperature remains at a certain level, which in turn means that the periodicity of activations is constant and the appliance activates frequently. Therefore, a disaggregation model needs to capture the activation pattern very precisely to reach low regression errors and high classification performance. The masking procedure in the pre-training process of ELECTRIcity aims to model the noisy distributions in the aggregate signal, which is not suitable for constant recurring activations, as in the case of the fridge. On the contrary, the fridge-freezer appliance in the Refit dataset is different than the fridge, as it combines a periodic low-power activation with high consumption peaks stemming from the freezer cooling. Even though ELECTRIcity achieves the best MRE, it does not fully capture the activation pattern behavior, resulting in lower F1 score.

Table 3.2 Comparison of ELECTRIcity's model performance to other techniques in the UK-DALE dataset.

| Device | Model | MRE | MAE | MSE | Acc. | F1 | Training Time (min) |
|---|---|---|---|---|---|---|---|
| Kettle | GRU+ [57] | 0.004 | 12.38 | 28,649.73 | 0.996 | 0.799 | 40.67 |
| | LSTM+ [57] | 0.004 | 11.78 | 28,428.10 | 0.997 | 0.800 | 34.70 |
| | CNN [38, 58] | **0.002** | **6.92** | 16,730.81 | 0.998 | 0.889 | 51.87 |
| | BERT4NILM [25] | 0.003 | 9.80 | 16,291.56 | 0.998 | 0.912 | 697.87 |
| | ELECTRIcity | 0.003 | 9.26 | **13,301.43** | **0.999** | **0.939** | 294.37 |
| Fridge | GRU+ [57] | 0.797 | 31.47 | 1966.50 | 0.750 | 0.673 | 33.35 |
| | LSTM+ [57] | 0.813 | 32.36 | 2058.13 | 0.748 | 0.661 | 34.67 |
| | CNN [38, 58] | 0.726 | 30.46 | 1797.54 | 0.718 | 0.686 | 44.43 |
| | BERT4NILM [25] | **0.683** | **20.17** | **1087.36** | **0.859** | **0.831** | 687.12 |
| | ELECTRIcity | 0.706 | 22.61 | 1213.61 | 0.843 | 0.810 | 428.79 |
| Washer | GRU+ [57] | 0.056 | 21.90 | 27,199.96 | 0.950 | 0.228 | 33.06 |
| | LSTM+ [57] | 0.055 | 23.42 | 32,729.26 | 0.950 | 0.221 | 34.14 |
| | CNN [38, 58] | 0.023 | 15.41 | 25,223.21 | 0.984 | 0.518 | 44.46 |
| | BERT4NIMLM [25] | 0.012 | 4.09 | 4369.72 | 0.994 | 0.775 | 687.74 |
| | ELECTRIcity | **0.011** | **3.65** | **2789.35** | **0.994** | **0.797** | 462.95 |
| Microwave | GRU+ [57] | 0.015 | 7.16 | 8464.09 | 0.994 | 0.131 | 35.18 |
| | LSTM+ [57] | 0.014 | 6.60 | 7917.85 | 0.995 | 0.207 | 37.05 |
| | CNN [38, 58] | 0.014 | 6.44 | 7899.43 | 0.995 | 0.193 | 47.32 |
| | BERT4NILM [25] | 0.014 | 6.53 | 8148.81 | 0.995 | 0.049 | 755.58 |
| | ELECTRIcity | **0.013** | **6.28** | **7594.23** | **0.996** | **0.277** | 518.93 |
| Dishwasher | GRU+ [57] | 0.035 | 28.60 | 43,181.30 | 0.975 | 0.722 | 44.31 |
| | LSTM+ [57] | 0.036 | 28.75 | 42,333.18 | 0.975 | 0.727 | 47.48 |
| | CNN [38, 58] | 0.051 | 41.44 | 80,292.31 | 0.960 | 0.087 | 56.99 |
| | BERT4NILM [25] | **0.026** | **14.11** | **14,676.17** | 0.982 | 0.804 | 859.87 |
| | ELECTRIcity | 0.028 | 18.96 | 24,152.70 | **0.984** | **0.818** | 462.83 |

Table 3.3 Comparison of ELECTRIcity's model performance to other techniques in the REDD dataset.

| Device | Model | MRE | MAE | MSE | Acc. | F1 | Training Time (min) |
|---|---|---|---|---|---|---|---|
| Washer | GRU+ [57] | 0.028 | 35.83 | 87,742.33 | 0.985 | 0.576 | 2.94 |
| | LSTM+ [57] | 0.026 | 35.71 | 89,855.09 | 0.983 | 0.490 | 3.08 |
| | CNN [38, 58] | 0.020 | 35.78 | 94,248.61 | 0.982 | 0.000 | 3.80 |
| | BERT4NILM [25] | 0.021 | 35.79 | 93,217.72 | 0.990 | 0.190 | 59.01 |
| | ELECTRIcity | **0.016** | **23.07** | **44,615.35** | **0.998** | **0.903** | 35.14 |
| Microwave | GRU+ [57] | 0.061 | 18.97 | 23,352.36 | 0.983 | 0.382 | 1.89 |
| | LSTM+ [57] | 0.060 | 18.91 | 24,016.75 | 0.983 | 0.336 | 1.53 |
| | CNN [38, 58] | 0.056 | 18.07 | 24,653.65 | 0.987 | 0.336 | 2.22 |
| | BERT4NILM [25] | **0.055** | 16.97 | 22,761.11 | 0.989 | 0.474 | 27.72 |
| | ELECTRIcity | 0.057 | **16.41** | **17,001.33** | **0.989** | **0.610** | 16.49 |
| Dishwasher | GRU+ [57] | 0.049 | 24.91 | 22,065.08 | 0.962 | 0.341 | 2.76 |
| | LSTM+ [57] | 0.050 | 25.09 | 22,297.01 | 0.961 | 0.350 | 2.98 |
| | CNN [38, 58] | 0.041 | 25.28 | 23,454.64 | 0.962 | 0.000 | 4.45 |
| | BERT4NILM [25] | **0.038** | **19.67** | **15,488.62** | **0.974** | 0.580 | 59.24 |
| | ELECTRIcity | 0.051 | 24.06 | 19,853.05 | 0.968 | **0.601** | 35.08 |

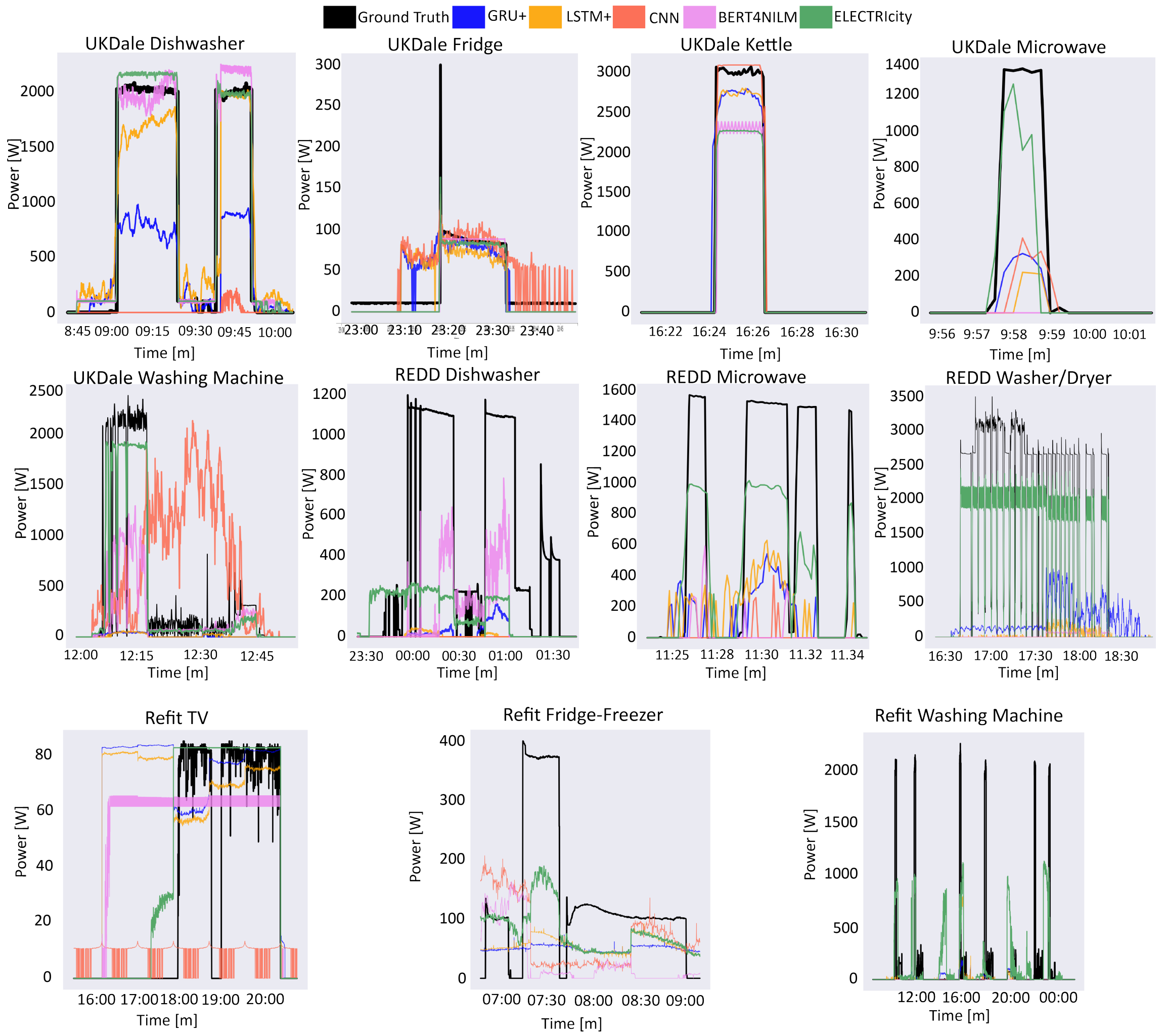

Figure 3.4 Comparison between ground truth appliance consumption signal and comparative model outputs for all examined appliances.

Next, we examine appliances with sparse, but longer duration activations (Washing Machine, Dishwasher), where ELECTRIcity showcases superior performance compared to the other models. For the washing machine, ELECTRIcity has better performance both in regression and in classification metrics. This performance increase is especially evident in the REDD dataset, where the F1 score is approximately 40% better than the second-best performing model. As for the dishwasher, its activation pattern is different than the washing machine, and contains more major fluctuations. ELECTRIcity produces a higher F1 score in both datasets, albeit with a lower MAE. This is due to the fact that the pre-training process of ELECTRIcity is suitable for modeling abnormal noisy distributions in the aggregate signal, which fits the activation profile of this appliance category. At the same time, ELECTRIcity requires 55% less training time than the second-best performing model (BERT4NILM) for the washing machine and 45% for the dishwasher, confirming the efficiency increase of our approach. We can therefore conclude that ELECTRIcity is the most suitable model for disaggregation of the washing machine and the dishwasher. In addition to the aforementioned appliances, we evaluate the disaggregation performance of an entertainment appliance (television). Entertainment appliances have particular disaggregation interest since they can be one of the main sources of energy saving for a domestic household. The television consumption pattern is different from the appliances examined so far, as the activations are distant and have a lower power consumption. Therefore, it is easier for the activations to be "lost" in the aggregate signal. However, our approach outperforms the other models both in regression and classification metrics, while requiring 75% less training time than the second-best performing model. This finding is very interesting and paves the way for evaluating ELECTRIcity on other entertainment appliances.

To summarize the above findings across all datasets, ELECTRIcity exhibits an average comparative performance increase of 9.03%, 5.38% and 23.59% in terms of MRE, MAE and MSE respectively, as well as an increase of 5.10% and 27.68% in terms of accuracy and F1-score to the second-best performing model [25], thus confirming the superiority of our approach. Finally, we examine the performance advantages that the pre-training procedure yields in terms of training time between the two Transformer-based models (ELECTRIcity and BERT4NILM). The total amount of training time per appliance can be seen in Figure 3.5. On average, ELECTRIcity required approximately 50% less training time than BERT4NILM using the same model size and hyperparameters. Overall, the introduction of a more efficient pre-training technique that is not limited to a percentage of the data leads to both performance and training time improvements, which makes ELECTRIcity a fast and efficient Transformer architecture for energy disaggregation.

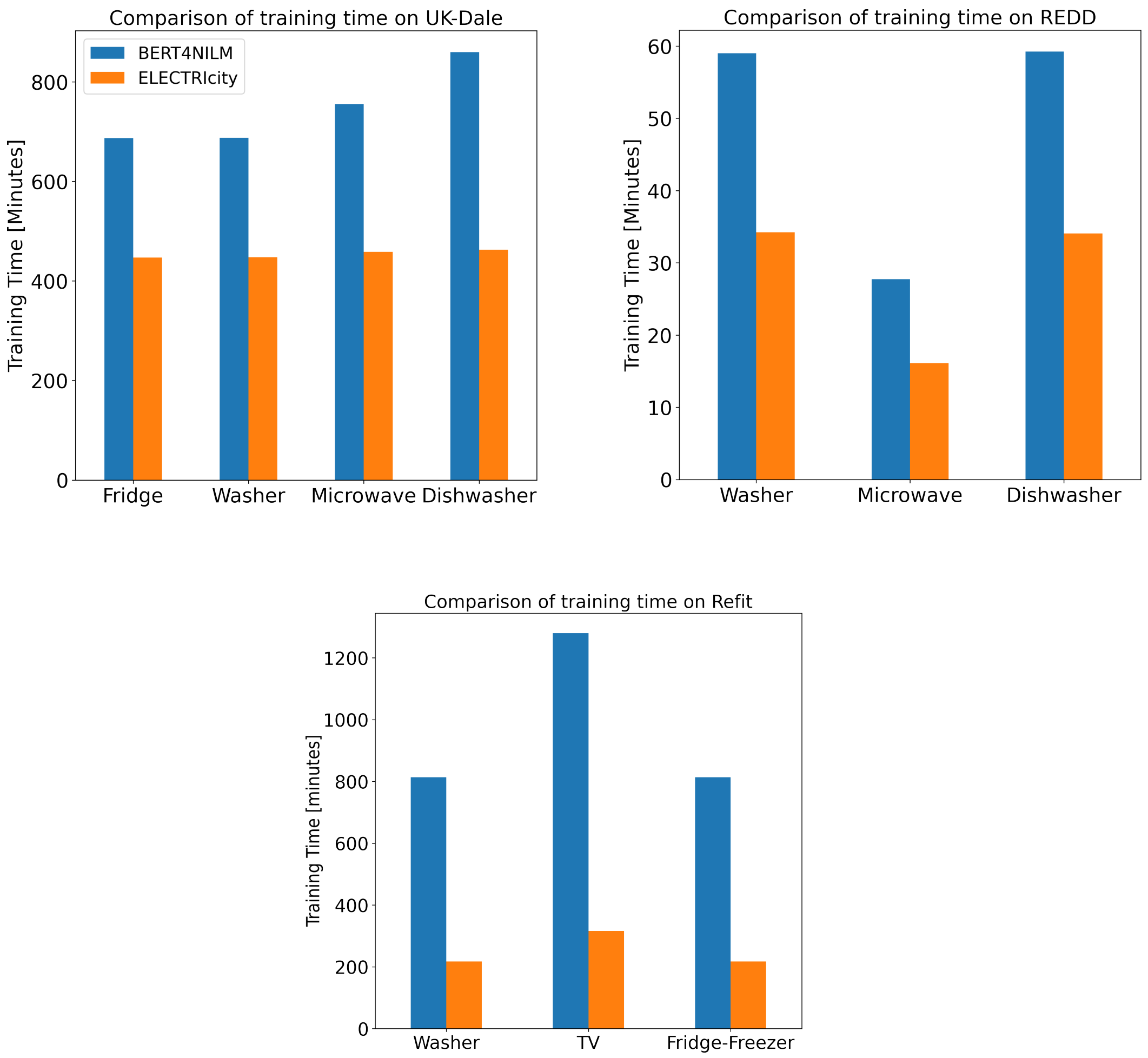

Figure 3.5 Training time comparison between transformer architectures on UK-Dale, REDD and Refit datasets.

## 3.4 Conclusions

In this Chapter, we introduced ELECTRIcity, an efficient fast Transformer-based architecture for energy disaggregation. ELECTRIcity outperforms state-of-the-art models in both examined datasets without requiring any data balancing. Averaging across all devices, ELECTRIcity achieves a performance boost across both datasets. The most significant increase can be showcased through the MSE and F1-score, where ELECTRIcity attains an average comparative increase of 23.59% and 27.68% respectively against the second-best performing model BERT4NILM[25]. At the same time, ELECTRIcity requires 50% less training time than BERT4NILM, making our approach superior in both performance and computational efficiency.

However, the performance evaluation of our approach has highlighted some limitations. In appliances with sparse and short activations, the increased training time of ELECTRIcity may not be always justified, compared to models with lower computational demands during the training phase. The disaggregation performance of the model, even though it outperforms the other comparative models, needs to be enhanced to solidify the preference toward ELECTRIcity, especially in cases where the pre-training masking procedure fails to model the noise distribution of the aggregate signal. At the same time, ELECTRIcity offers a great opportunity to improve the performance of NILM on appliances such as fridges or fridge-freezers, where the activation behavior is recurring at a similar consumption level. The results on entertainment appliances with small power consumption (and thus difficult to be disaggregated), such as the television, are very promising and open further research opportunities in that direction. Finally, we believe that our approach, which can work with minimal data pre-processing, is a big step towards the large-scale integration of NILM techniques in domestic households.

# Chapter 4

# Performance-aware NILM model optimization for edge deployment

## 4.1 Introduction

Non-Intrusive Load Monitoring (NILM) refers to the process of analyzing the aggregated energy consumption of a residential building to infer the individual consumption pattern of domestic appliances [59]. In recent years, NILM approaches have transversed from statistical analysis methods to deep learning techniques due to their superior performance capabilities. However, most of the deep learning NILM approaches are designed to be deployed in a central server instead of performing inference on the edge due to the increased computational needs [60–62]. This design methodology assumes data transfer from the data source, i.e., the domestic house, to an external entity and impacts the wider deployment scalability of NILM frameworks. Central data storage has increased costs for the service provider since the accumulation of large amounts of data requires an expanded storage infrastructure. In addition, performing inference centrally usually requires more computational resources, thus increasing the energy required to run the service and increasing the carbon footprint of the solution. Finally, apart from the heavy reliance on a stable internet connection for data transmission, privacy concerns arise since sensitive customer information can be inferred [63]. It can therefore be argued that a transition to deploying NILM algorithms on the edge (i.e., at each domestic house equipped with a smart meter and a device with restricted processing power) is a more attractive solution that alleviates the issues of central data processing.

### Our Contribution

In this Chapter, we propose a performance-aware NILM optimization framework for edge deployment that takes into account the edge device characteristics [64]. Our approach considers multiple hardware limitations and, depending on the deployment scenario, employs a different model optimization technique to efficiently preserve the limited edge device resources, resulting in an efficient resource management scheme. The basic contributions of our work are summarized below:

- **NILM green computing edge-inference framework**: We propose an edge inference framework for NILM that utilizes multiple model optimization techniques, taking into account the edge device hardware characteristics to enable an efficient edge green computing scheme.
- **Model optimization metric**: We introduce Pruning Gain, an objective model optimization metric for NILM algorithms that describes the trade-off between model performance and computational complexity.
- **Performance-aware NILM model edge optimization**: We present performance-aware pruning, an iterative algorithm to determine which model parameters can be removed from the network without severely impacting model performance.
- **Application specific NILM model edge optimization**: We explore the impact of model optimization on various NILM techniques (CNN, LSTM, Transformers) for different appliances, and we experimentally prove that, depending on the application scenario, a different level of model optimization for resource management is tolerable from a model performance perspective.

The rest of the chapter is organized as follows. First, we present an overview of the existing work for deploying NILM algorithms on edge devices. Section 4.2 mathematically formulates the problem of performance-aware NILM model optimization and describes in detail the proposed NILM edge optimization framework. Finally, Section 4.3 presents the experimental setup and results, while Section 4.4 summarizes the main outcomes of the chapter and potential future steps.

## Related Work

Since its official problem formutation [13], NILM has received increasing research interest, backed by the expanded availability of smart meter data. Earlier NILM approaches were based on signal processing techniques, such as Graph Signal Processing [65, 66] and Hidden Markov Models [67, 68]. Since 2015, the research focus has shifted towards utilizing deep learning techniques on low-frequency data, with models such as Denoising Autoencoders [27, 69], Recurrent Neural Networks [30, 39] and Convolutional Neural Networks [21, 33, 38] being successfully applied for energy disaggregation. Due to the rapid advancements in deep learning, state-of-the-art network architectures such as U-net [70], Generative Adversarial Networks [71, 72] and Transformers [24, 25] have been employed to advance NILM.

Recently, progress has been made towards the deployment of NILM and other energy-related applications on edge devices, either as part of a Home Energy Management System (HEMS) [73] or as standalone applications [74]. NILM edge inference does not require the transmission of data to an external server and, therefore, alleviates the aforementioned issues of central data processing. Approaches to deploying NILM models on the edge have been proposed, both on embedded computers, such as Raspberry Pi, and on more resource-constrained devices. Deployment on a Raspberry Pi has been proposed [75, 76], but the deployed models either require additional metadata, such as room occupancy or utilize high-frequency features for energy disaggregation, which increases data acquisition costs. In addition, NILM models on more resource-constrained devices, such as microcontrollers [77, 78] and FPGA [79] have also been proposed, but the respective models only consider appliance state classification instead of regression and require high-frequency data to operate.

Despite the great success of deep learning in diverse applications, neural networks often possess a vast number of parameters, leading to significant challenges in deploying deep learning systems to a resource-limited device [80, 81]. The deployment of sensing devices with higher computational power has been investigated [82, 83], but the devices have high cost and high power demands, thus making them impractical for commercialization [77]. Therefore, edge inference requires compression and optimization of NILM deep learning models to account for the limited computational resources. Quantization, parameter pruning, low-rank factorization, and knowledge distillation [84], as well as combinations of one or more techniques [85] are the main approaches employed in the literature. Even though NILM-related deep learning applications have utilized state-of-the-art architectures [18, 24, 86], research on the constraints and methodology for deploying NILM deep learning models on edge devices remains limited. In [87], the quantization of a sequence-to-point (seq2point) convolutional neural network (CNN) [38] from 32-bit float model weights to 8-bit integer weights is applied. The application of multiple pruning approaches on the same seq2point [38] model has also been investigated [88], and the methods have been tested on 2 appliances from the REFIT [56] dataset. Finally, [89] explores model compression of a multi-class seq2point CNN using pruning and tensor decomposition, while the evaluation is performed for three appliances from the REDD dataset [55].

Even though the aforementioned works can be considered as an initial entry-point towards low-frequency (⇒ 1Hz) NILM inference on edge devices, there are several limitations. First, these papers [87–90] do not take into consideration the hardware characteristics of edge devices. This can be an issue in quantization approaches, where some quantization protocols are applicable only to specific chip architectures. Second, all papers employ compression approaches on a specific model architecture (seq2point CNN). Seq2point models are less computationally efficient than sequence-to-sequence models (seq2seq), since they produce only one timepoint prediction instead of a whole window in the testing phase. As a result, significantly more forward pass iterations are required to produce the same number of outputs seq2seq models, which leads to a noteworthy increase in energy consumption. In addition, the effects of model compression on different model architectures, such as recurrent neural networks or Transformers, have not been investigated. Furthermore, the works that investigate more than one model compression strategy do not explore the impact of their combination on model performance, which can result in the optimization of different model aspects. Finally, pruning is applied on an arbitrary basis, and no framework has been proposed to interconnect performance loss after compression with model complexity. A summary of the aforementioned limitations of existing literature can be found in Table 4.1.

Table 4.1 Summary of existing literature for edge NILM

| Work | Deployment Device | Model | Compression Approach | Limitations |
|---|---|---|---|---|
| Uttama et. al. [75] | Raspberry Pi | Combinatorial optimization | state complexity reduction | requires additional metadata (room occupancy information) |
| Xu et. al. [76] | Raspberry Pi | Support Vector Machine | Feature reduction | requires high frequency data |
| Tabanelli et. al. [77] | Micro-controller | Random Forest | Feature reduction | -Requires high frequency data<br>-event-based NILM (classification) |
| Tabanelli et. al. [78] | Micro-controller | Random Forest | Feature reduction | |
| Hernandez et. al. [79] | FPGA | Hardware-oriented | - | |
| Ahmed et.al. [87] | | seq2point CNN | Quantization | - Hardware characteristics are not taken into account<br>-Seq2point requires more forward passes than seq2seq; computationally intensive<br>- Only 1 model architecture is considered<br>- Different model compression approaches are not jointly investigated<br>- Compression is conducted arbitrarily and not connected to performance loss |
| Barber et.al. [88] | - | seq2point CNN | Pruning | |
| Kukunuri et.al. [89] | | seq2point CNN | Pruning, Tensor decomposition | |

# 4.2 Methodology

## 4.2.1 Problem Statement

Under a NILM framework [13], we can assume that the aggregate consumption signal $x$ of a domestic house with $M$ operational appliances, at any time point $t$, equals to the sum of the individual appliance consumption loads $y_i$, $i = 1 \ldots M$, plus a noise term s [14] (see Eq. 2.2).

To extract the consumption signal of a selected appliance $a \rightarrow \{1, \ldots, M\}$, NILM approaches are designed to filter out all non-relevant appliance consumption signals $y_i \downarrow i$ ⇞ $a$. Depending on the chosen appliance $a$, the power signal $y_a$ may showcase different statistical characteristics in terms of peaks, sparsity, or duration of appliance activations, which is defined as the consecutive time interval that the appliance is turned on. As a result, not only may different model architectures have different sensitivity to model optimization approaches, but also the same model, trained to disaggregate different appliances, may showcase different behavior related to compression. It can therefore be argued that the optimization strategy must be bound to model performance, with the goal of finding an equilibrium between model complexity and performance loss.

Two different NILM infrastructure setups for central and edge device deployment are illustrated in Figure 4.1. Central inference requires upload and download of consumption data to a central processing entity, as well as increased data storage capacity and computational power. On the contrary, performing inference on the edge devices alleviates these limitations and only requires the compression of the models and their deployment on the edge device, while data exchange takes place only between the edge device and the domestic house.

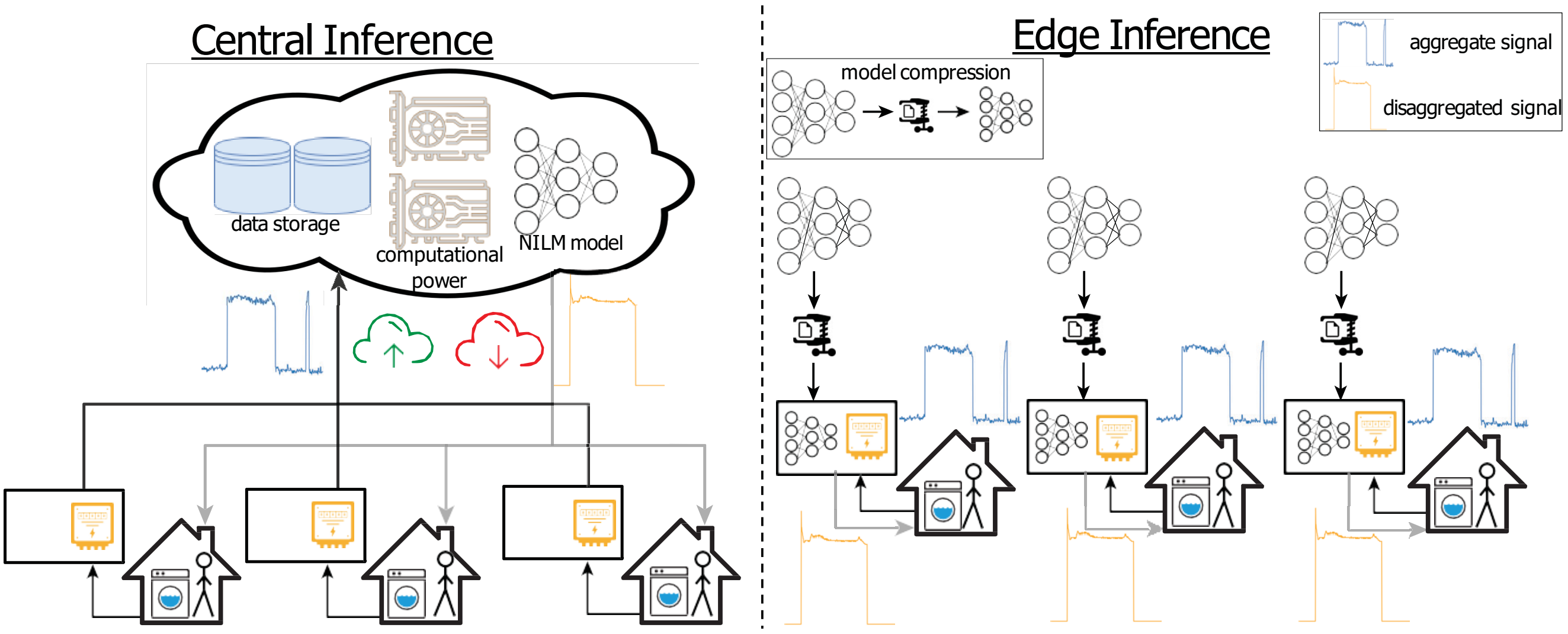

Figure 4.1 Overview of the required infrastructure setup to perform inference centrally vs on the edge.

To mathematically formulate the aforementioned approach, let a NILM model $f(x;w) : X \Leftarrow Y, w \rightarrow R^N$ have a performance $P_f$. Our goal is to obtain a lower-dimensionality model $h(x;\theta) : X \Leftarrow Y$, where $\theta$ is some transformation of $w$, i.e. $\theta = T(w), \theta \rightarrow R^C, C < N$, to perform the same task with performance $P_h$. In other words, we are trying to minimize the following function:

$$L = min(/f(x;w) \searrow h(x;\theta)/) \tag{4.1}$$

However, Equation 4.1 does not take into consideration the performance loss that occurs as a consequence of model optimization. The model should be optimized to match the deployment criteria only to the level that the performance loss is acceptable. Therefore, Equation 4.1 needs to be constrained with the condition that performance loss must not fall below a tolerance threshold $\ni$ . Therefore, a performance-aware model compression framework can be written as:

$$L_{pa} = min(/f(x;w) \searrow h(x;\theta)/) \;\; s.t. \;\; P_f \searrow P_h < \ni \tag{4.2}$$

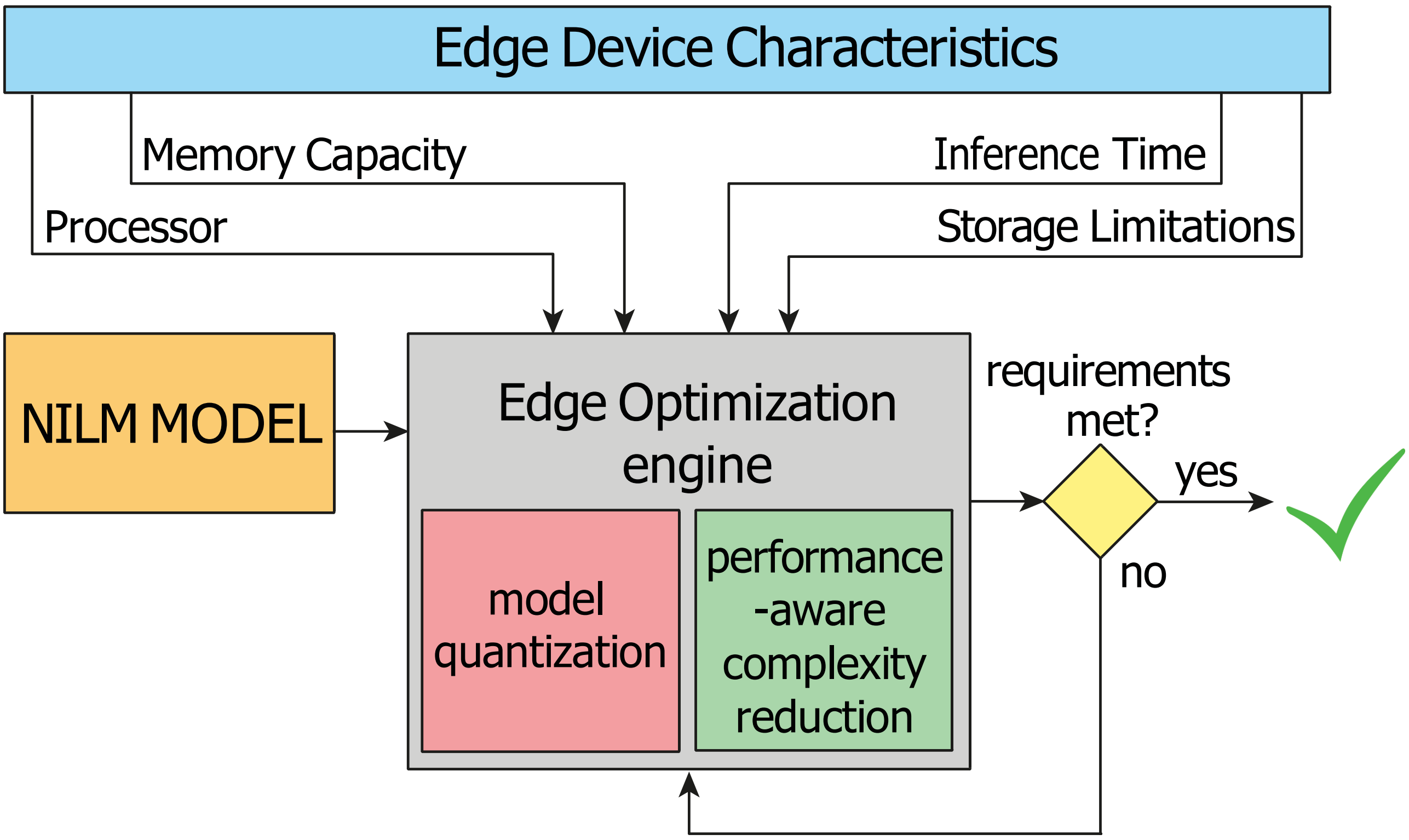

Figure 4.2 High-level overview of the proposed NILM edge optimization framework

### 4.2.2 A green edge resource management framework for NILM

Our proposed green computing framework for NILM model edge optimization is illustrated in Figure 4.2. The backbone of our approach is the edge optimization engine, which is responsible for the optimization of a NILM model depending on the edge deployment requirements. Since resource limitations of the edge device may vary, the optimization engine first receives the edge device characteristics, as well as any additional restrictions imposed by the user. Then, the trained NILM model to be deployed is analyzed, and an optimization strategy is set. The optimization strategy can either be static to reduce the model's storage requirements through model quantization or performance-aware to apply complexity reduction through weight pruning. Performance-aware optimization is defined as the removal of insignificant model weights not arbitrarily but by taking into consideration the respective impact on model performance. In this case, complexity reduction is performed incrementally until the edge deployment requirements are met, under the condition that the trade-off between performance loss and complexity reduction is satisfactory.

An overview of the optimization approaches employed for memory and complexity reduction is depicted in Figure 4.3. We explore model quantization by performing MinMax quantization of the model weights, as well as histogram quantization for the activation function outputs to minimize performance loss. We also integrate magnitude pruning in our approach to remove weights with small L1-norm that contribute minimally to the model's predictions. The following sections provide a detailed description of these techniques, as well as the proposed performance-aware iterative complexity reduction scheme.

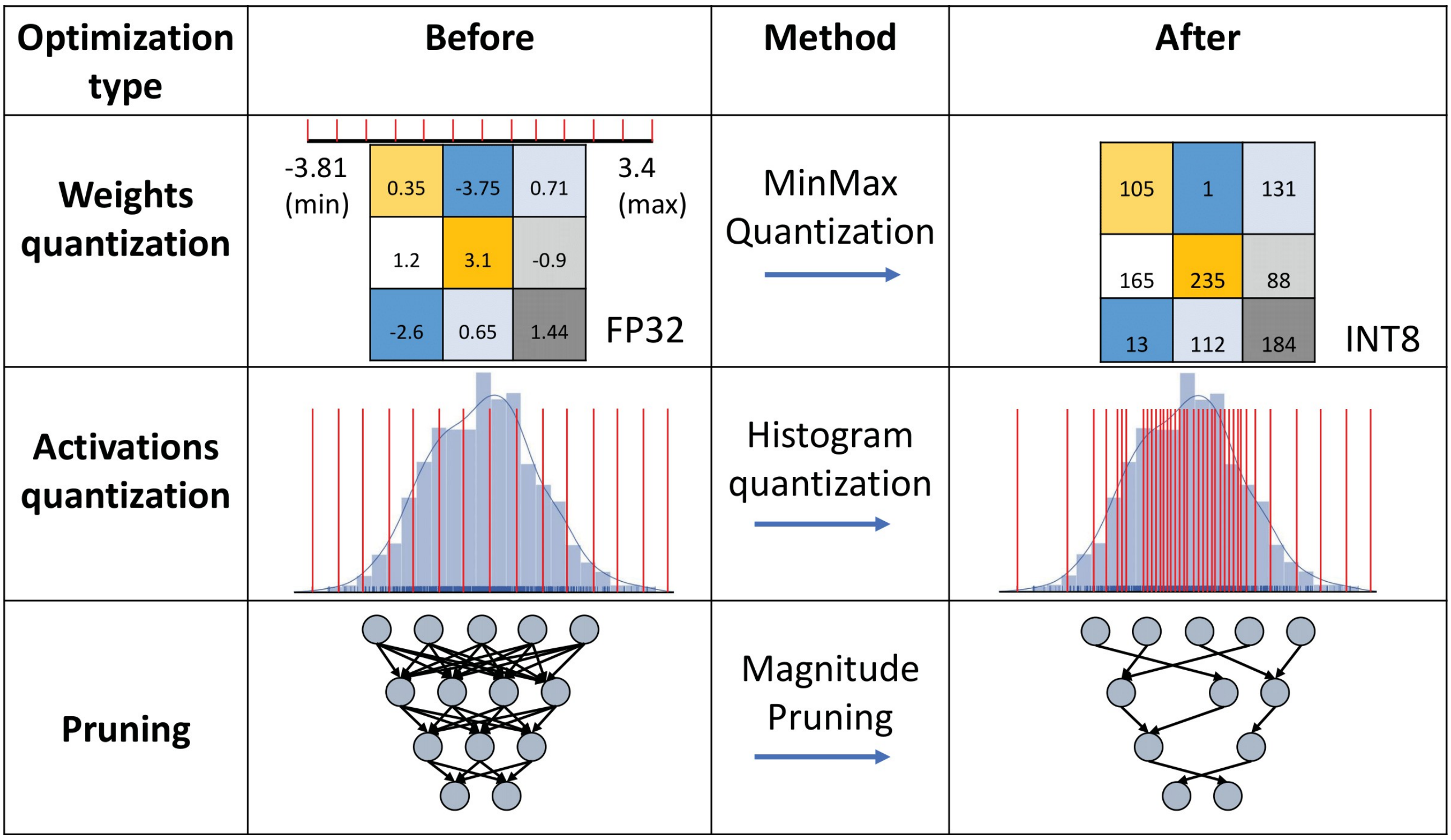

Figure 4.3 Overview of the model optimization methods adopted in this study.

#### 4.2.2.1 Model weights quantization

Model quantization refers to the process where the model's weight type is changed to lower numerical precision to limit the storage and memory space required for the model. In essence, quantization can be formulated as an irreversible mapping function $Q : w \rightarrow R^N \Leftarrow w^{\leftrightarrow} \rightarrow R^Z, Z \Downarrow N$ that maps the model weights $w$, stored in a floating point format, to an integer representation $w^{\leftrightarrow}$. The value range of $w$ is divided into bins and each value $w_i$ is mapped to the integer representing the corresponding bin.

Quantization is either executed post-training, meaning that an already trained model is compressed or during training, in the sense that the quantized version of the model is taken into account when the model is trained (quantization-aware training). In this work, we focus on post-training model quantization and apply a calibration phase on an indicative dataset, during which the quantization parameters are fine-tuned, resulting in a more accurate representation of the initial model weights $w$. This additional calibration step also allows for the quantization of activation function outputs.

To quantize the models, we quantize both model weights and activation outputs to further avoid floating point multiplication operations [91]. Since the activation outputs are fed to the next layer, a more sensitive quantization approach is required to minimize model performance degradation. Therefore, we have opted for min-max uniform quantization for model weights and histogram quantization for the activation outputs, where the activation values are recorded, and a different range per bin is assigned, depending on the corresponding probability distribution. An illustration of the different quantization approaches can be seen in Figure 4.3.

#### 4.2.2.2 Model complexity reduction

An alternative methodology for optimizing a deep learning model is the removal of synaptic connections between model layers. This process, which is commonly referred to in the literature as model pruning, assumes that a deep learning network is over-parameterized and incorporates a subnetwork that contains most of the information [92]. In other words, model pruning is an approach to transform a model's weights $w \rightarrow R^N$ to a lower dimensionality representation $w^{\leftrightarrow} \rightarrow R^M, M < N$ by removing non-informative model connections.

Different techniques on how to optimally remove model connections with minimal information loss have been proposed in the literature. Similar to quantization, pruning can either be applied post-training [93] or in a compression-aware training scheme [94]. The removal of weights is performed either on the overall set of model weights or by eliminating predetermined architectural blocks, such as convolutional filters [95]. In addition, different pruning approaches remove weights by evaluating different metrics, such as weights magnitude, gradients magnitude, intra-layer mutual information, or even by introducing a learnable pruning threshold [96–98].

In this work, we implement magnitude pruning and remove the model connections with the smallest contribution to the model output. Let $w = \{w_i \downarrow i = 1, \ldots, N\} \rightarrow R^N$ be a vector containing all model parameters. Then the magnitude-pruned vector $w^{\leftrightarrow}$ is expressed in Equation 4.3:

$$w^{\leftrightarrow} \Leftrightarrow w, w^{\leftrightarrow} := \left\{ \rightarrow w : \frac{F(\| w_i \|)}{\sum_{j=1}^{N} F(\| w_j \|)} < p_{thres} \right\} \tag{4.3}$$

$F(\| w_i \|)$ signifies the cumulative distribution function of weight magnitudes. In other words, after magnitude pruning, we only keep the weights with the highest $1 \searrow p_{thres}\%$ magnitudes and discard the rest. Even though magnitude pruning is usually executed only once in post-training pruning, in the next section, we present an iterative variation that calculates the optimal $p_{thres}$ bound to the resulting model performance.

#### 4.2.2.3 Iterative performance-aware green resource management algorithm

Magnitude pruning removes a percentage of a model's lowest $L^1$-norm connections, according to a specified threshold $p_{thres}$. However, finding the optimal pruning threshold $p_{opt}$ that represents the optimal tradeoff between model complexity and performance is often a tedious procedure that requires multiple experimentations, whose evaluation is, in most cases, subjective. Therefore, we propose Performance-Aware Optimized Pruning (PAOP), an iterative algorithm to determine the optimal pruning threshold for NILM models. Optimality must be bound in terms of performance, as stated in Equations 4.2. Consequently, finding the optimal pruning threshold $p_{opt}$ requires an objective metric that incorporates both the performance degradation of the reduced model and the gain in terms of parameter reduction. Therefore, the metrics utilized for model performance evaluation need to be first defined. Since seq2seq disaggregation is primarily a regression task and secondarily a classification task, we record three widely used metrics for model evaluation, namely, Mean Absolute Error (MAE) and Mean Relative Error (MRE) for regression evaluation and F1-score for classification performance, as shown in Equations 3.9 and 3.10.

For different pruning thresholds $p_{thres}$, the model performance on the test set will change. At the same time, each metric should not be evaluated independently. Instead, all metrics should be combined in a single term. Taking into consideration all the aforementioned considerations, we propose the Pruning Gain metric (PG) to quantify the tradeoff between model complexity and performance, which is formulated in Equation 4.4.

$$PG = \frac{MAE_b\ MRE_b\ F1_p\ N_{param,b}}{MAE_p\ MRE_p\ F1_b\ N_{param,p}} \tag{4.4}$$

Pruning Gain measures the pruning-related change in a given metric as the ratio of the baseline performance of the model to the performance of the model after pruning. For metrics where a lower score is better, the terms of the ratio need to be reverted (baseline/pruned). For each metric, we record the ratio of baseline model performance (subscript b) and the model performance after pruning (subscript p), and multiply it by the ratio of change in the number of parameters of the original model and the redacted version. The idea behind PG is to combine the increase or decrease of the metrics recorded to evaluate model performance with the reduction in model size in a multiplicative way. This approach was selected to emphasize the sensitivity of changes in model performance, as an averaging operation of the individual terms would lead to the phenomenon where a positive change in one metric may envelop negative changes in the other ones. Even though the separate metrics are in different scales, the ratio of each metric captures the relative change between the baseline and the pruned version, which regularizes each ratio separately. No change results in a ratio of 1. A PG score greater than 1 means that the performance loss from removing model weights is beneficial, whereas a score smaller than 1 signifies that the performance drop was more significant than the model compression achieved. Therefore, the proposed metric captures the relative changes between the metrics and can be used to decide whether the impact of pruning on the NILM model was negligible or not.

Utilizing PG, we are now able to perform iterative magnitude pruning to optimally compress a NILM model. To take the hardware characteristics of the edge device into account, the expert needs to define computational cost goals depending on the deployment scenario. Then, iterative model optimization can begin. First, we define a selected range of pruning threshold percentages $[0, p_{max}]$ that should be taken into consideration, as well as an iteration step $p_{step}$. Then, for each pruning threshold $p$, we calculate the performance metrics, as well as the Pruning Gain PG. If Pruning Gain is, for the given pruning percentage, higher than 1, then we assume that the reduction of the weights dimensionality was beneficial and that the model can be further compressed, in which case we increment the pruning threshold with $p_{step}$%. We continue the aforementioned loop until the Pruning Gain falls below 1, where the iteration stops. If the computational cost goals were met, then the previous pruning percentage $p$ is selected as the optimal pruning for the given model. Otherwise, the model is not deployable on the edge device. The iterative algorithm is summarized in Algorithm 1.

**Algorithm 1** Performance-aware green resource management algorithm

**Input:** $cost_{goal}$ Expert-defined computational cost goals (MFLOPs)
**Define** $p$ : pruning percentage
**Define**: $p_{max}$, $p_{step}$, $p_{opt}$
**for** p in range (0, $p_{step}$, $p_{max}$) **do**
  Calculate performance metrics (MAE, MRE, F1)
  Calculate pruning gain PG
  **if** $PG > 1$ **then**
    $p_{opt}$ =p
  **else if** $PG < 1$ **then**
    Calculate $cost_{new}$
    **if** $cost_{new} \Rightarrow cost_{goal}$ **then**
      break; optimal model found
    **else**
      break; model is not deployable
    **end if**
  **end if**
**end for**

## 4.3 Experimental setup and results

### 4.3.1 Experimental Setup

The methodology to optimize a model for edge inference should depend on the application scenario and the hardware limitations inherent to the edge device. The deployment of NILM models on the edge can be achieved by connecting a smart meter that records the aggregate consumption with a Raspberry Pi 3 Model b single-board computer. Raspberry Pi is one of the most popular edge devices in IoT systems and is commonly used as a gateway to enable the deployment of AI applications in real-world settings [99]. Therefore, we have designed our methodology and experiments to use a Raspberry Pi 3 as the edge device. Raspberry Pi's run on an ARM architecture and have limited storage space and computational power, but are easy to install and use. Their hardware characteristics can be found in Table 4.2.

Table 4.2 Technical specifications of a Raspberry Pi 3 Model b

| | |
|---|---|
| **Architecture** | ARM |
| **Processing power** | Quad Core 1.2GHz Broadcom BCM2837 64bit CPU |
| **Memory size** | 1GB RAM |
| **Connectivity** | Ethernet, WLAN |
| **Storage** | SD Card |

The architecture to deploy the optimized models on the edge device is depicted in Figure 4.4. The edge solution consists of 3 different services responsible for data collection and the NILM inference service, which processes the collected data and produces the disaggregation results. The components of the data collection process are described below:

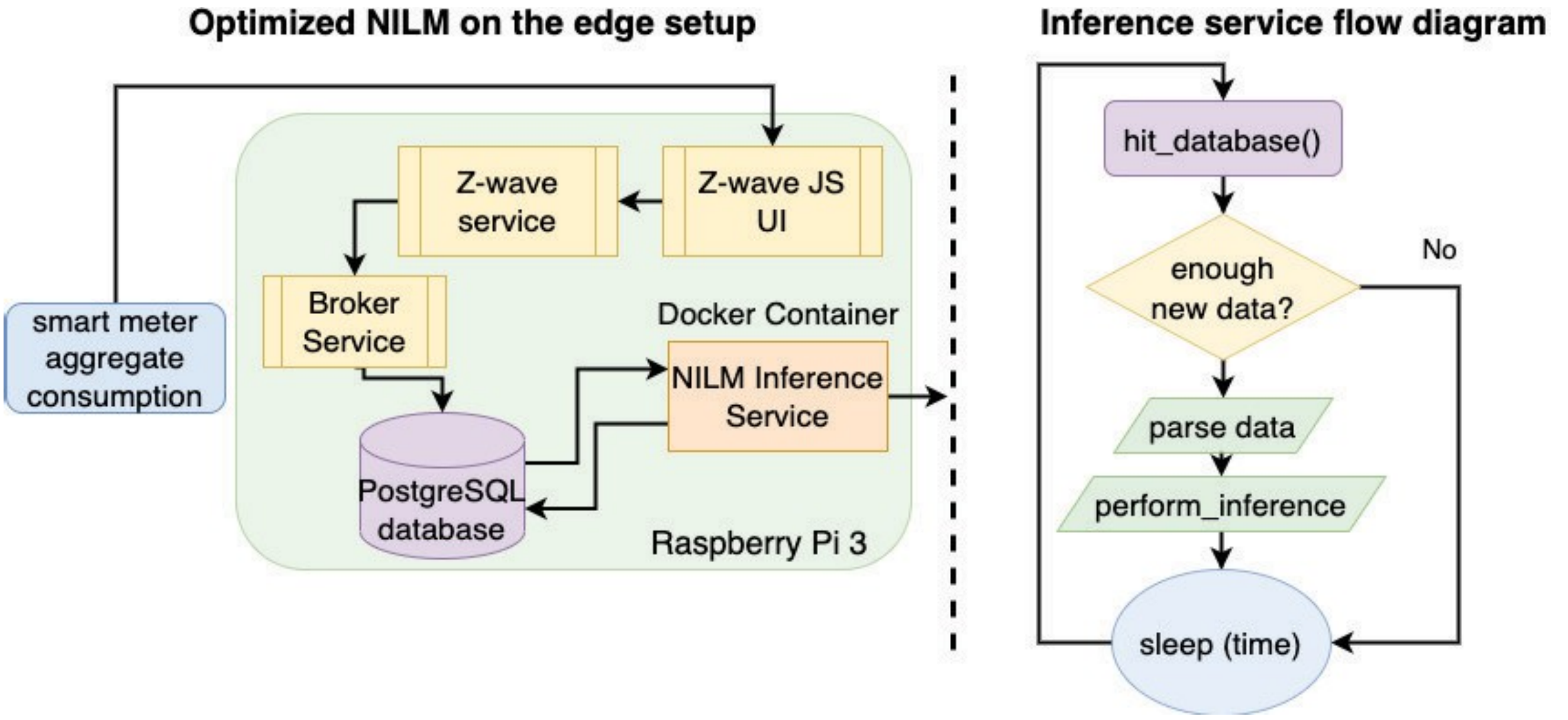

Figure 4.4 Overview of the proposed edge NILM deployment architecture. The edge setup consists of services for data collection and NILM inference (left), which performs disaggregation once enough data are collected (right).

- **Z-Wave JS UI** is an open-source dockerized service that communicates with the aggregate consumption smart meter through Z-wave protocol and forwards the collected data to the Z-Wave service through the MQTT protocol.
- **Z-wave-service** is a custom service that receives the collected data from the Z-Wave JS UI through MQTT protocol and forwards them to the data broker service through an API.
- **DataBroker-service** is responsible for receiving the collected data from Z-wave service and communicating with the PostgreSQL database. DataBroker service is also responsible to update (saving and deleting) the collected data in the existing database.

The **NILM inference service** is included in a Docker container that runs continuously on the edge device. This service communicates directly with the database after a specified time interval and checks if enough data are collected to produce the disaggregation results.

Depending on the processor's architecture, there are two main quantization backend libraries that can be used, namely FGBEMM [100] and QNNPACK [101]. The term backend refers to reduced precision tensor matrix math libraries that are utilized during model compression. FBGEMM can be used to quantize a model to run on x86 architectures, while QNNPACK supports ARM processor architectures. Since the Raspberry Pi processing unit is based on an ARM architecture, we have chosen QNNPACK as the quantization backend.

To evaluate our approach, we conducted experiments on different appliances from UK-Dale [54], and REDD [55] datasets. Both datasets consist of aggregate and appliance level energy consumption measurements from five different houses in the United Kingdom and six different houses in the United States, respectively. UK-Dale was generated at a sample rate of 1 Hz for the aggregate and 1/6 Hz for individual appliances while REDD was monitored at a sample rate of 1 Hz for the aggregate consumption and 1Hz for the plug-level data. The data were resampled at a sample rate of 1/6 Hz and pose the appliance characteristics described in Table 4.3. The models were tested on unseen data from houses not included in the training set, as shown in Figure 4.5. The reason for testing the models on a house not used in the training set is due to the core concept of NILM; if a house has smart meter data to record appliance consumption, there is no point in deploying a NILM algorithm to infer them, since they are already available to the consumer. Therefore, the proposed approach is to perform NILM on smart meter aggregate readings from a house using pre-trained models, for which ground truth in terms of submetering was available for training on a centralized server, e.g. using publicly available datasets. In UK-Dale, we focused on four appliances (washer, kettle, fridge, dishwasher), while in REDD on three appliances (microwave, washer, dishwasher). The set of appliances selected represents single-state and multi-state appliances with variable load fluctuations.

Table 4.3 Appliance characteristics for UK-Dale and REDD datasets.

| Dataset | Appliance | Max Limit [W] | On Thresh. [W] | Min. On Duration [s] | Min. Off Duration [s] |
|---|---|---|---|---|---|
| UK-Dale | Kettle | 3100 | 2000 | 12 | 0 |
| | Washer | 2500 | 20 | 1800 | 160 |
| | Fridge | 300 | 50 | 60 | 12 |
| | Dishwasher | 2500 | 10 | 1800 | 1800 |
| REDD | Microwave | 1800 | 200 | 12 | 30 |
| | Dishwasher | 1200 | 10 | 1800 | 1800 |
| | Washer-Dryer | 500 | 20 | 1800 | 160 |

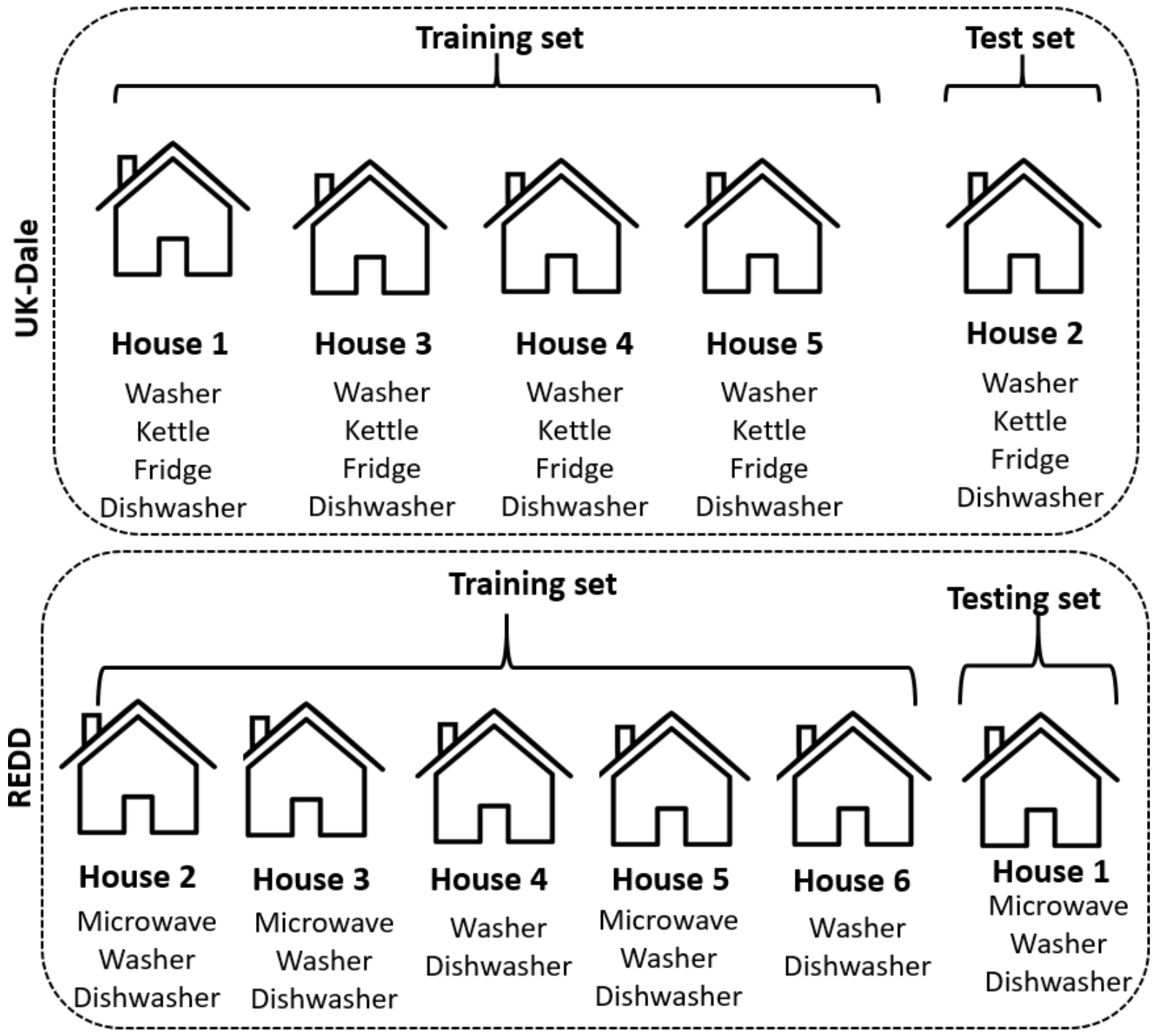

Figure 4.5 Train-test split for UK-Dale and REDD datasets. In UK-Dale, houses 1,3,4 and 5 were used for training and house 2 for testing, while in REDD house 2,3,4,5 and 6 were included in the training set and house 1 was kept for model evaluation.

To diversify our experimental evaluation and test the generalization capabilities of our performance-aware edge inference optimization framework, the models are based on different architectural philosophies. In particular, one convolutional neural network [38], two recurrent architectures (LSTM [15], GRU [57]), and a Transformer-based model [24] were chosen for the evaluation of our approach, and their architectural representations are illustrated in Figure 4.6. The upper left subfigure describes a convolutional neural network, whereas the next two correspond to reccurent architectures with different gating mechanisms (LSTM, GRU). Finally, the lower right subfigure presents the Transformer-based architecture. Even though all models initially employ a 1-D convolutional filter for feature extractions, the intermediate part of the model structure varies significantly. The models were purposely trained and evaluated on unbalanced data. The reason for not balancing the dataset is that, to mitigate the negative aspects of central data storage, model training should take place in a federated manner on edge devices, where the possibility of data balancing is limited by hardware constraints. We envision our work as part of a wider NILM framework that enables the transition from central data processing to all computations occurring on the edge to increase the privacy of customers.

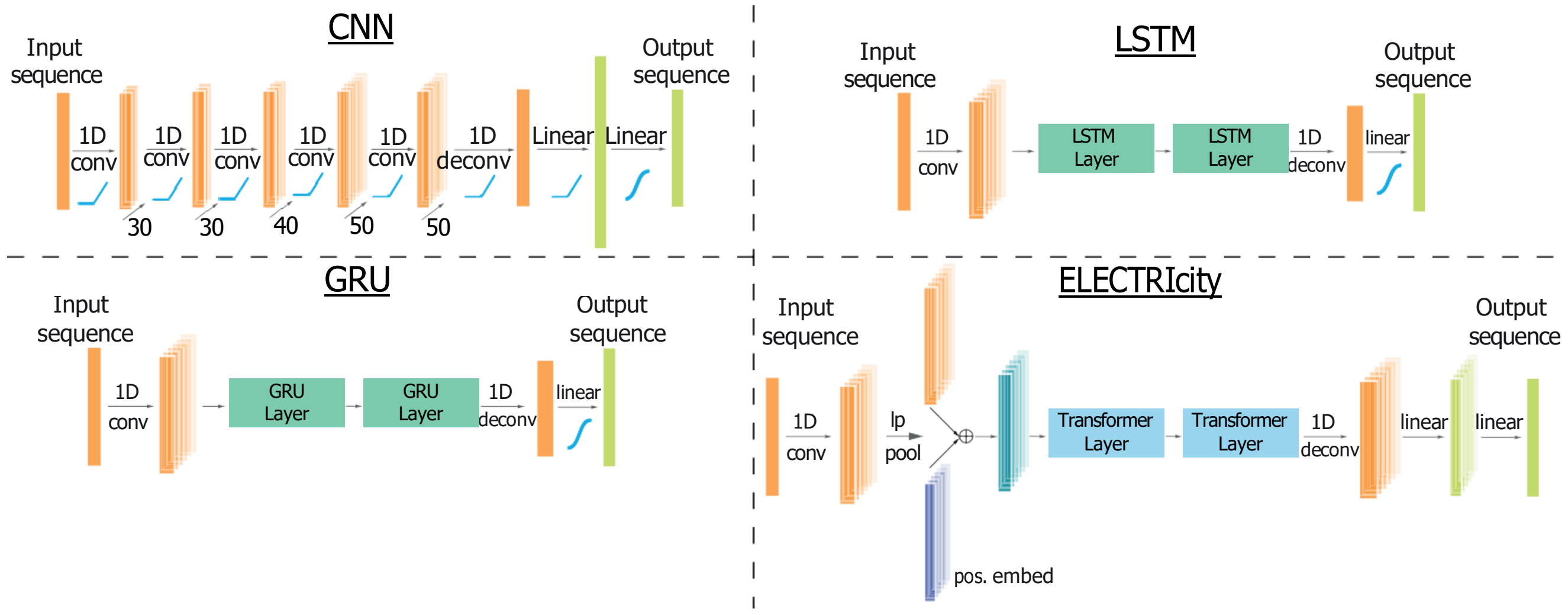

Figure 4.6 Model architecture for the four NILM models that were used to evaluate the impact of model compression on model performance and complexity.

In our analysis, we diversify between three edge deployment scenarios based on different edge device limitations. First, the edge device has limited storage capacity, and the edge optimization engine employs model quantization to limit the required storage space of the model. In the second scenario, the limitation is based on the edge device's processing power, and we optimize the models with Performance-Aware Optimized Pruning (PAOP) to reduce their computational complexity. Finally, we investigate the optimization scenario where the edge devices have limited both storage space and computational power. We apply a combination of performance-aware optimized pruning to reduce the number of floating point operations during a forward pass, followed by weight quantization to reduce storage requirements. We call the combination of both techniques Performance-Aware Pruning and Quantization (PAOPQ). In the utilization of the proposed performance-aware schemes, the model complexity reduction ranged between [0, 70] %, with an increment step of 5%. All optimization experiments were performed on an Apple Macbook M1 Pro to take advantage of the ARM CPU architecture to accurately simulate the deployment of the aforementioned models on a real-world setting with Raspberry Pi edge devices.

Table 4.4 Size on disk before and after model quantization

| Model | Size on disk (MB) | |
|---|---|---|
| | Original | Quantized |
| CNN | 4.0 | 0.98 |
| LSTM | 4.6 | 1.15 |
| GRU | 3.6 | 0.90 |
| ELECTRIcity | 7.8 | 2.89 |

### 4.3.2 Results

#### 4.3.2.1 Scenario 1: Limited storage capacity

The first step in our analysis is to examine how the aforementioned models were impacted by weight quantization. As can be seen in Table 4.4, quantization of model weights leads to a significant 75% reduction in the size required to store the model on the disk. At the same time, the effect of quantization on the disaggregation performance, which is presented in Tables 4.5, 4.6 and 4.7, varies across model architectures and appliances. In UK-Dale results, recurrent neural networks (LSTM, GRU) showcase minimal performance degradation when disaggregating the kettle and fridge, with a performance reduction of less than 0.5% for all metrics. However, they are sensitive to weight quantization for appliances with sparse and long appliance activations, such as the washing machine and the dishwasher. The CNN model has a small performance loss consistent across appliances, while the Transformer-based model (ELECTRIcity) is robust to quantization, showcasing a minimal performance degradation averaging across all appliances (-0.01% MAE, -1.09% MRE and -0.29% F1). The effects of quantization presented in UK-Dale are very similar when the quantized models are evaluated on the REDD dataset. Recurrent as well as Transformer architectures present a minimal performance degradation across all the tested appliances. In some cases, quantization could also lead to a slight improvement in disaggregation results as it happens in LSTM, GRU, and Electricity models on microwave appliance as well as on Electricity model on washer with an average improvement of 10.80% MAE, 8.57% MRE and 2.2% F1.

Table 4.5 Performance metrics of optimized models - UK-Dale (Kettle, Fridge)

| Appliance | Model | Approach | MAE | MRE | F1 |
|---|---|---|---|---|---|
| Kettle | CNN | Baseline | 7.03 | 0.0025 | 0.89 |
| | | Quantized | 7.55 | 0.0026 | 0.88 |
| | | PAOP | 8.71 | 0.0030 | 0.85 |
| | | PAOPQ | 8.97 | 0.0031 | 0.85 |
| | LSTM | Baseline | 12.18 | 0.0043 | 0.80 |
| | | Quantized | 12.20 | 0.0043 | 0.79 |
| | | PAOP | 13.74 | 0.0048 | 0.76 |
| | | PAOPQ | 13.74 | 0.0048 | 0.76 |
| | GRU | Baseline | 12.93 | 0.0046 | 0.79 |
| | | Quantized | 12.93 | 0.0046 | 0.79 |
| | | PAOP | 14.08 | 0.0050 | 0.77 |
| | | PAOPQ | 14.08 | 0.0050 | 0.77 |
| | ELECTRIcity | Baseline | 9.26 | 0.0032 | 0.94 |
| | | Quantized | 9.26 | 0.0032 | 0.94 |
| | | PAOP | 10.17 | 0.0036 | 0.92 |
| | | PAOPQ | 10.18 | 0.0036 | 0.92 |
| Fridge | CNN | Baseline | 31.86 | 0.78 | 0.64 |
| | | Quantized | 34.45 | 0.75 | 0.62 |
| | | PAOP | 37.41 | 0.63 | 0.65 |
| | | PAOPQ | 39.45 | 0.63 | 0.64 |
| | LSTM | Baseline | 32.85 | 0.82 | 0.63 |
| | | Quantized | 33.07 | 0.82 | 0.62 |
| | | PAOP | 35.72 | 0.86 | 0.54 |
| | | PAOPQ | 35.95 | 0.87 | 0.54 |
| | GRU | Baseline | 31.32 | 0.80 | 0.67 |
| | | Quantized | 31.44 | 0.80 | 0.66 |
| | | PAOP | 34.21 | 0.84 | 0.59 |
| | | PAOPQ | 34.38 | 0.84 | 0.58 |
| | ELECTRIcity | Baseline | 23.10 | 0.71 | 0.80 |
| | | Quantized | 23.08 | 0.71 | 0.80 |
| | | PAOP | 27.61 | 0.76 | 0.73 |
| | | PAOPQ | 27.52 | 0.76 | 0.73 |

Table 4.6 Performance metrics of optimized models - UK-Dale (Washer, Dishwasher)

| Appliance | Model | Approach | MAE | MRE | F1 |
|---|---|---|---|---|---|
| Washer | CNN | Baseline | 10.45 | 0.02 | 0.58 |
| | | Quantized | 13.64 | 0.02 | 0.56 |
| | | PAOP | 9.77 | 0.02 | 0.59 |
| | | PAOPQ | 12.43 | 0.02 | 0.57 |
| | LSTM | Baseline | 21.88 | 0.05 | 0.24 |
| | | Quantized | 23.51 | 0.11 | 0.13 |
| | | PAOP | 14.59 | 0.10 | 0.12 |
| | | PAOPQ | 23.51 | 0.11 | 0.13 |
| | GRU | Baseline | 19.29 | 0.05 | 0.24 |
| | | Quantized | 20.02 | 0.08 | 0.18 |
| | | PAOP | 9.38 | 0.05 | 0.10 |
| | | PAOPQ | 20.01 | 0.08 | 0.18 |
| | ELECTRIcity | Baseline | 3.65 | 0.01 | 0.85 |
| | | Quantized | 3.66 | 0.01 | 0.84 |
| | | PAOP | 5.21 | 0.01 | 0.73 |
| | | PAOPQ | 4.58 | 0.01 | 0.76 |
| Dishwasher | CNN | Baseline | 41.29 | 0.04 | 0.09 |
| | | Quantized | 41.30 | 0.04 | 0.09 |
| | | PAOP | 41.38 | 0.05 | 0.08 |
| | | PAOPQ | 41.37 | 0.05 | 0.08 |
| | LSTM | Baseline | 31.25 | 0.03 | 0.65 |
| | | Quantized | 32.74 | 0.13 | 0.28 |
| | | PAOP | 33.01 | 0.04 | 0.57 |
| | | PAOPQ | 32.71 | 0.13 | 0.28 |
| | GRU | Baseline | 31.43 | 0.03 | 0.62 |
| | | Quantized | 31.53 | 0.05 | 0.53 |
| | | PAOP | 39.75 | 0.04 | 0.45 |
| | | PAOPQ | 31.55 | 0.05 | 0.53 |
| | ELECTRIcity | Baseline | 18.96 | 0.03 | 0.82 |
| | | Quantized | 18.93 | 0.03 | 0.81 |
| | | PAOP | 24.75 | 0.03 | 0.79 |
| | | PAOPQ | 24.40 | 0.04 | 0.71 |

Table 4.7 Performance metrics of optimized models-REDD

| Appliance | Model | Approach | MAE | MRE | F1 |
|---|---|---|---|---|---|
| Microwave | CNN | Baseline | 17.49 | 0.0558 | 0.38 |
| | | Quantized | 17.29 | 0.0558 | 0.37 |
| | | PAOP | 18.16 | 0.0562 | 0.21 |
| | | PAOPQ | 17.27 | 0.0557 | 0.37 |
| | LSTM | Baseline | 34.86 | 0.0923 | 0.31 |
| | | Quantized | 19.69 | 0.0622 | 0.33 |
| | | PAOP | 35.03 | 0.1065 | 0.28 |
| | | PAOPQ | 17.04 | 0.3156 | 0.03 |
| | GRU | Baseline | 19.35 | 0.0618 | 0.37 |
| | | Quantized | 19.35 | 0.0618 | 0.38 |
| | | PAOP | 18.84 | 0.0588 | 0.24 |
| | | PAOPQ | 5.44 | 0.0348 | 0.18 |
| | ELECTRIcity | Baseline | 17.45 | 0.0562 | 0.42 |
| | | Quantized | 17.46 | 0.0562 | 0.43 |
| | | PAOP | 20.23 | 0.0667 | 0.52 |
| | | PAOPQ | 11.16 | 0.0198 | 0.29 |
| Washer-Dryer | CNN | Baseline | 5.83 | 0.0283 | 0.24 |
| | | Quantized | 3.83 | 0.0125 | 0.00 |
| | | PAOP | 5.92 | 0.0346 | 0.22 |
| | | PAOPQ | 3.83 | 0.0125 | 0.12 |
| | LSTM | Baseline | 6.17 | 0.0288 | 0.21 |
| | | Quantized | 6.19 | 0.0289 | 0.21 |
| | | PAOP | 6.16 | 0.0287 | 0.20 |
| | | PAOPQ | 6.20 | 0.0299 | 0.20 |
| | GRU | Baseline | 5.38 | 0.0342 | 0.19 |
| | | Quantized | 5.39 | 0.0344 | 0.18 |
| | | PAOP | 5.61 | 0.0348 | 0.16 |
| | | PAOPQ | 5.39 | 0.0344 | 0.18 |
| | ELECTRIcity | Baseline | 15.98 | 0.0204 | 0.31 |
| | | Quantized | 15.98 | 0.0208 | 0.32 |
| | | PAOP | 11.17 | 0.0197 | 0.29 |
| | | PAOPQ | 11.16 | 0.0198 | 0.29 |
| Dishwasher | CNN | Baseline | 30.23 | 0.0784 | 0.12 |
| | | Quantized | 28.65 | 0.0768 | 0.12 |
| | | PAOP | 30.88 | 0.0810 | 0.06 |
| | | PAOPQ | 29.54 | 0.0745 | 0.06 |
| | LSTM | Baseline | 34.86 | 0.0923 | 0.31 |
| | | Quantized | 34.83 | 0.0938 | 0.30 |
| | | PAOP | 35.03 | 0.1065 | 0.28 |
| | | PAOPQ | 34.98 | 0.1066 | 0.27 |
| | GRU | Baseline | 49.33 | 0.0942 | 0.30 |
| | | Quantized | 49.37 | 0.0948 | 0.30 |
| | | PAOP | 48.37 | 0.0966 | 0.30 |
| | | PAOPQ | 49.04 | 0.0991 | 0.29 |
| | ELECTRIcity | Baseline | 17.35 | 0.0498 | 0.63 |
| | | Quantized | 17.36 | 0.0496 | 0.61 |
| | | PAOP | 11.17 | 0.0197 | 0.29 |
| | | PAOPQ | 21.67 | 0.048 | 0.57 |

Table 4.8 Baseline model parameters and optimal pruning threshold, as obtained by applying Algorithm 1

| Dataset | Appliance | Model | Baseline Parameters (k) | Optimal pruning threshold $p_{opt}$ (%) | | MFLOPs | | |
|---|---|---|---|---|---|---|---|---|
| | | | | *PAOP* | *PAOPQ* | *Baseline* | *PAOP* | *PAOPQ* |
| UK-Dale | Kettle | CNN | 996.5 | 40 | 5 | 18.33 | 13.75 | 17.41 |
| | | LSTM | 1141.7 | 30 | 30 | 16.71 | 12.50 | 12.50 |
| | | GRU | 887.5 | 25 | 25 | 12.44 | 10.17 | 10.17 |
| | | ELECTRIcity | 1938.4 | 25 | 25 | 586.70 | 452.08 | 452.08 |
| | Fridge | CNN | 996.5 | 70 | 70 | 18.33 | 10.00 | 10.00 |
| | | LSTM | 1141.7 | 30 | 30 | 16.71 | 11.67 | 11.67 |
| | | GRU | 887.5 | 25 | 25 | 12.44 | 8.67 | 8.67 |
| | | ELECTRIcity | 1938.4 | 40 | 40 | 586.70 | 376.43 | 376.43 |
| | Washer | CNN | 996.5 | 10 | 5 | 183.3 | 15.79 | 17.41 |
| | | LSTM | 1141.7 | 60 | 5 | 16.71 | 7.50 | 15.86 |
| | | GRU | 887.5 | 70 | 5 | 12.44 | 3.72 | 11.94 |
| | | ELECTRIcity | 1938.4 | 60 | 50 | 586.70 | 268.18 | 283.73 |
| | Dishwasher | CNN | 996.5 | 30 | 30 | 18.33 | 15.20 | 15.20 |
| | | LSTM | 1141.7 | 35 | 5 | 16.71 | 10.80 | 15.86 |
| | | GRU | 887.5 | 70 | 5 | 12.44 | 3.71 | 11.94 |
| | | ELECTRIcity | 1938.4 | 35 | 55 | 586.70 | 398.13 | 309.48 |
| Redd | Microwave | CNN | 996.5 | 70 | 5 | 18.33 | 10.45 | 17.82 |
| | | LSTM | 1141.7 | 5 | 30 | 16.71 | 15.73 | 11.79 |
| | | GRU | 887.5 | 45 | 45 | 12.44 | 7.60 | 7.60 |
| | | ELECTRIcity | 1938.4 | 70 | 70 | 586.70 | 220.04 | 220.04 |
| | Washer-Dryer | CNN | 996.5 | 60 | 5 | 18.33 | 11.65 | 17.80 |
| | | LSTM | 1141.7 | 5 | 5 | 16.71 | 15.72 | 15.72 |
| | | GRU | 887.5 | 70 | 5 | 12.44 | 3.85 | 11.81 |
| | | ELECTRIcity | 1938.4 | 35 | 35 | 586.70 | 402.62 | 402.62 |
| | Dishwasher | CNN | 996.5 | 60 | 60 | 18.33 | 11.94 | 11.94 |
| | | LSTM | 1141.7 | 5 | 5 | 16.71 | 15.73 | 15.73 |
| | | GRU | 887.5 | 20 | 15 | 12.44 | 10.40 | 10.91 |
| | | ELECTRIcity | 1938.4 | 35 | 35 | 586.70 | 403.27 | 403.27 |

#### 4.3.2.2 Scenario 2: Limited processing power

Next, we would like to evaluate how the models are affected by PAOP. Applying the proposed iterative algorithm to find the optimal pruning threshold, the average number of model parameters can be decreased by 40.93 % in UK-Dale and by 40 % in REDD dataset. The optimal pruning threshold for each model and appliance, as well as the number of baseline parameters, are illustrated in Table 4.8. By comparing the obtained optimal pruning threshold with the performance metrics, as described in Tables 4.5 and 4.5, we observe that, in cases where the baseline model does not perform well, indicated by the low F1 score and the high MRE, our algorithm concludes to suggest the highest pruning percentage $p_{max}$, whereas in cases where the model performs well, the suggested optimal pruning threshold coincides with a plausible value close to the average. An indicative example of this finding is illustrated during the pruning of the LSTM model for the disaggregation of the washing machine in the UK-Dale dataset. Since the baseline performance is suboptimal, the ratio of baseline performance to performance after pruning is very sensitive to change and even though the MRE rises by 4.3% in absolute value, the relative change is ↘83.22%. At the same time, however, the MAE is 33.28% better than the baseline, which can be explained by the fact that artifacts in the predicted appliance signature are no longer being produced, and, multiplied with the ratio of model parameter reduction, leads to a positive Pruning Gain value. In the example of ELECTRIcity for the dishwasher appliance, we observe that 35 % of the model weights are removed without notable affecting the model's disaggregation performance. Overall, it can be concluded that the utilization of our performance-aware model compression strategy can reduce the computational complexity of a NILM model without significantly affecting its performance. The complexity reduction is validated through the reduced number of floating point operations (FLOPs) required to perform a forward pass, as can be seen in Table 4.8. On average, PAOP reduces the FLOPs of a NILM model by 36.3% in UK-Dale and by 31.8% in REDD.

#### 4.3.2.3 Scenario 3: Limited storage capacity and processing power

The last experiment that was conducted was the combination of both aforementioned model compression approaches (PAOPQ). To calculate the optimal pruning threshold in this case, the model performance was evaluated after both schemes were applied. Therefore, the optimal threshold obtained is different than in the case of pruning (see Table 4.8). It can be easily noticed that the combination of both techniques tolerates significantly lower pruning percentages for most models. On average, the optimal pruning threshold is 37.4% lower in UK-Dale and 26.25% in REDD, compared to when weights quantization is not utilized. Therefore, it can be concluded that, without the proposed performance-aware optimization scheme, the performance degradation of the models would be significantly higher. The integration of both techniques in our scheme results in 75% less size on disk for both UK-Dale and REDD and, on average, 25.62% fewer model parameters and 22% fewer FLOPs for UK-Dale and 21.51% fewer model parameters and 21.09% fewer FLOPs for REDD, thus reducing both storage requirements and model computational complexity. Another interesting finding is that the optimal thresholds for ELECTRIcity remain the same as in the case of applying only parameter pruning and, in the case of the dishwasher, the model tolerates a higher pruning percentage, meaning that Transformer-based architectures are more robust to model compression than convolution-based and recurrent modeling approaches.

An illustration of the Pruning Gain metric values during the application of the algorithm to find the optimal pruning threshold $p_{thres}$ is illustrated in Figure 4.7. The upper part of the figure showcases the Pruning Gain distribution when only model pruning is applied, while the lower part demonstrates the distribution when both pruning and quantization are selected. We observe that the curves of the pruning gain are significantly different depending on the chosen model compression approach and that, if the model does not have a strong baseline performance (GRU), the Pruning Gain is gradually increasing with further parameter reduction. The difference in both plots validates the observation that model performance is more sensitive to combining both compression approaches and that the proposed metric can accurately quantify the tradeoff between model performance and computational complexity.

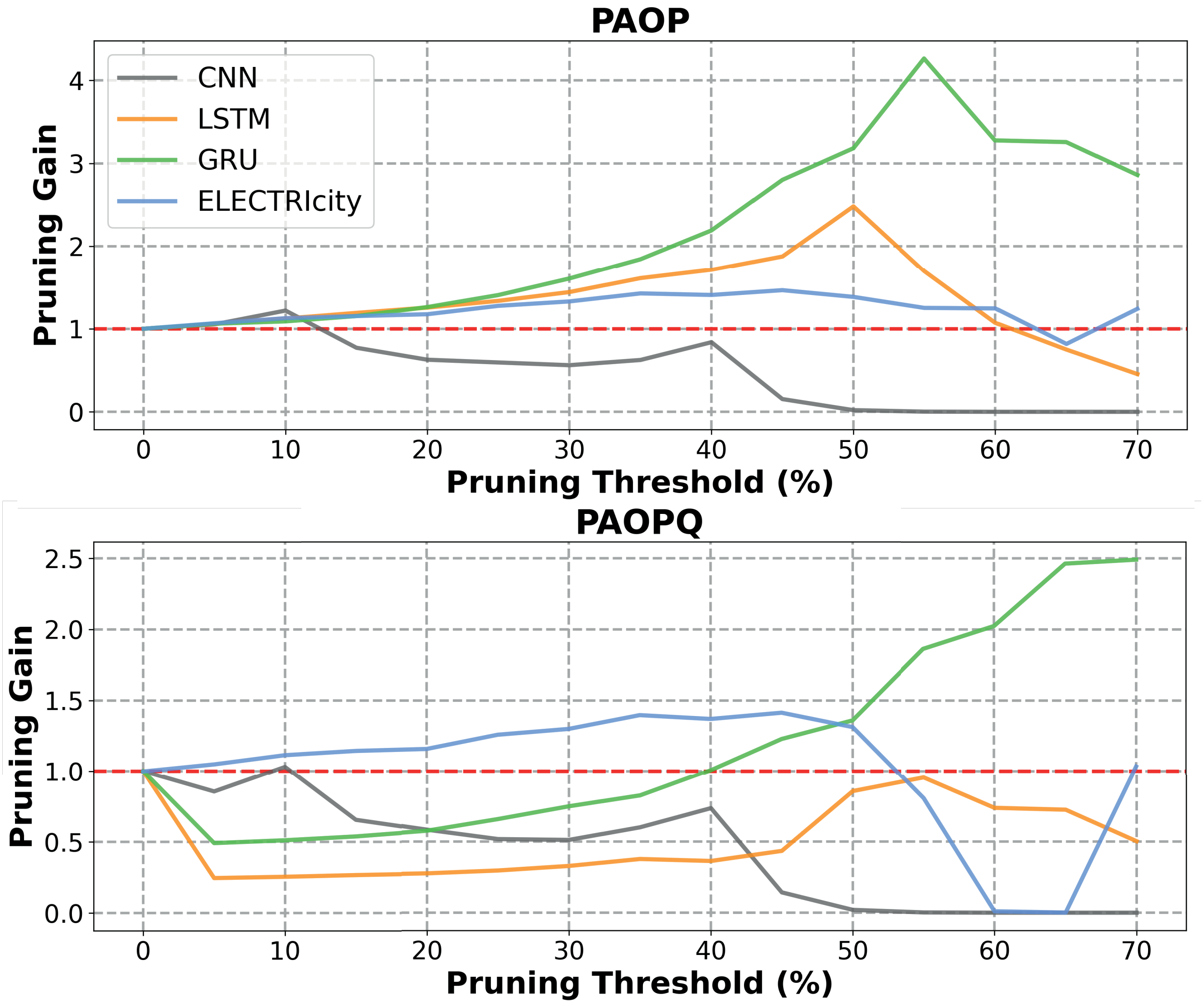

Figure 4.7 Pruning Gain plot over different pruning thresholds during the application of proposed iterative Algorithm 1 on the washing machine when the compression method is pruning (upper) vs pruning & quantization (lower).

Finally, we recorded the $CO_2$ emissions of each optimized model through the CodeCarbon Python library, and the results are presented in Table 4.9. All optimization approaches reduce $CO_2$ emissions by ↖ 17%. The only exception can be found for recurrent architectures (LSTM, GRU), where the $CO_2$ emissions are higher when quantization is involved. This can be explained by the increased complexity of recurrent layer computations, where multiple activation functions are utilized inside each memory cell. Since the outputs of activation functions are dynamically quantized during inference, the increased energy needs to perform the quantization is justifiable.

Table 4.9 $CO_2$ Emissions[g] for the baseline model and each optimization method for edge deployment

| Dataset | Appliance | Model | Baseline | Quantization | PAOP | PAOPQ |
|---|---|---|---|---|---|---|
| UK-Dale | Kettle | CNN | 0.072 | 0.061 | 0.071 | 0.061 |
| | | LSTM | 0.083 | 0.178 | 0.081 | 0.177 |
| | | GRU | 0.070 | 0.185 | 0.068 | 0.186 |
| | | ELECTRIcity | 1.240 | 1.059 | 1.209 | 1.068 |
| | Fridge | CNN | 0.060 | 0.049 | 0.059 | 0.049 |
| | | LSTM | 0.068 | 0.145 | 0.065 | 0.145 |
| | | GRU | 0.057 | 0.151 | 0.053 | 0.150 |
| | | ELECTRIcity | 1.246 | 1.014 | 1.134 | 1.025 |
| | Washing Machine | CNN | 0.060 | 0.050 | 0.057 | 0.050 |
| | | LSTM | 0.067 | 0.145 | 0.065 | 0.145 |
| | | GRU | 0.056 | 0.150 | 0.055 | 0.150 |
| | | ELECTRIcity | 1.257 | 1.016 | 1.157 | 1.014 |
| | Dishwasher | CNN | 0.060 | 0.049 | 0.059 | 0.049 |
| | | LSTM | 0.067 | 0.144 | 0.065 | 0.145 |
| | | GRU | 0.056 | 0.150 | 0.055 | 0.150 |
| | | ELECTRIcity | 1.235 | 1.030 | 1.187 | 1.052 |
| REDD | Microwave | CNN | 0.003 | 0.001 | 0.003 | 0.002 |
| | | LSTM | 0.002 | 0.002 | 0.002 | 0.002 |
| | | GRU | 0.002 | 0.001 | 0.002 | 0.001 |
| | | ELECTRIcity | 0.051 | 0.032 | 0.051 | 0.032 |
| | Washer-Dryer | CNN | 0.003 | 0.001 | 0.003 | 0.001 |
| | | LSTM | 0.002 | 0.002 | 0.002 | 0.002 |
| | | GRU | 0.002 | 0.001 | 0.002 | 0.001 |
| | | ELECTRIcity | 0.051 | 0.032 | 0.051 | 0.032 |
| | Dishwasher | CNN | 0.003 | 0.001 | 0.003 | 0.002 |
| | | LSTM | 0.002 | 0.002 | 0.002 | 0.002 |
| | | GRU | 0.002 | 0.001 | 0.002 | 0.001 |
| | | ELECTRIcity | 0.051 | 0.032 | 0.051 | 0.032 |

### 4.3.3 Discussion

As already mentioned, our approach suffers from certain limitations. First, we have observed that models that do not showcase good baseline performance tend to be overpruned by our proposed iterative algorithm during the search for the optimal pruning threshold. Even though such models should not be deployed to perform inference, as the insights that the consumer will get regarding energy consumption will not be accurate, our approach should still take into consideration such cases. The second limitation concerns the fact that, in our approach, the models are optimized for each different appliance. Therefore, to perform energy disaggregation for multiple appliances, the deployment of multiple NILM models is required. However, we have experimented with model optimization for all appliances simultaneously, as shown in Figure 4.8, and have found that optimizing the model for all appliances at the same time leads to over/underpruning and impacts the achievable performance. Averaging across all appliances and models, this approach would lead to a performance loss of 8.92% MAE, 7.32% MRE and 12.20% F1.

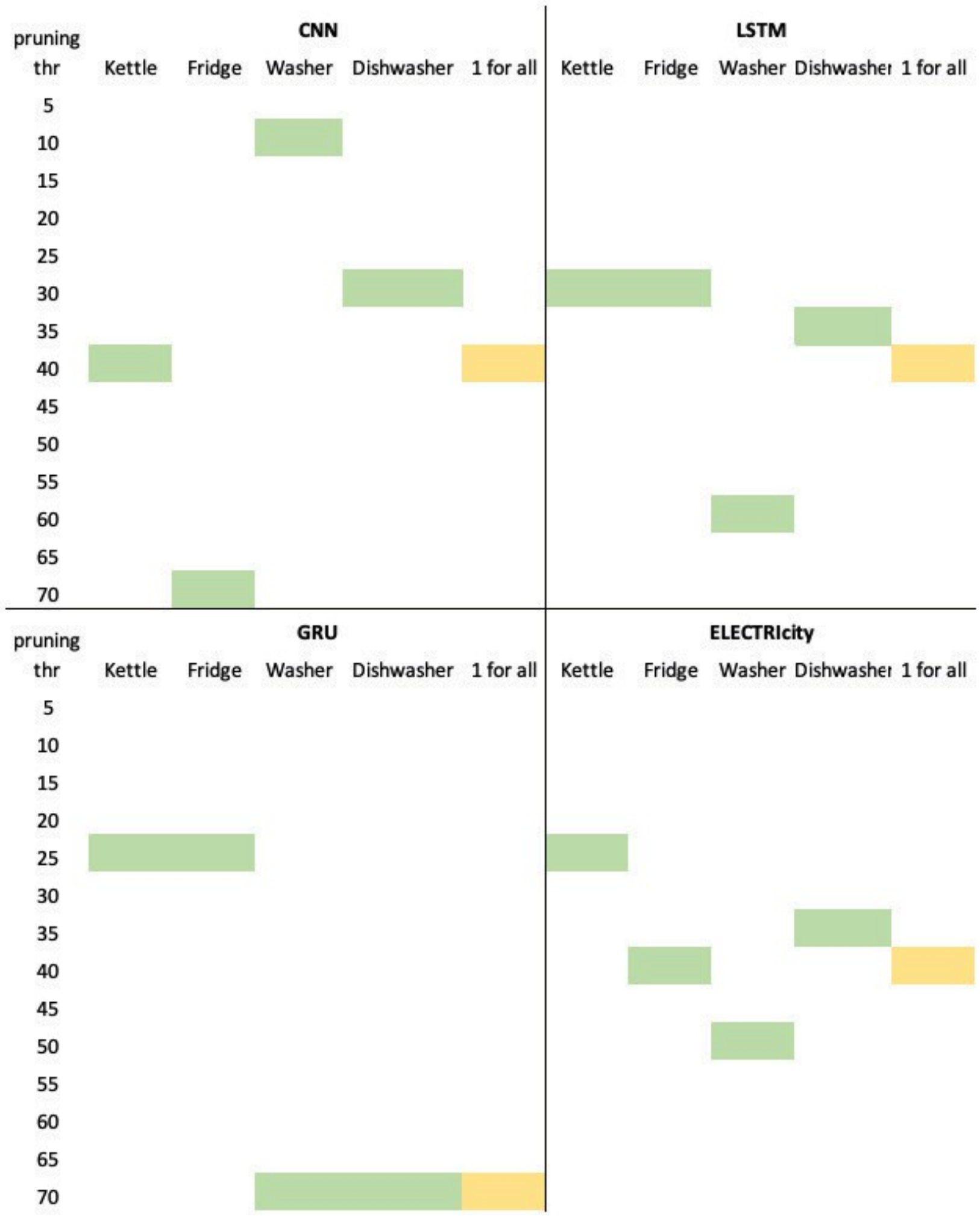

| Model | Appliance | | MAE | MRE | F1 |
|---|---|---|---|---|---|
| CNN | Kettle | single | 8.71 | 0.003 | 0.852 |
| | | 1 for all | 8.71 | 0.003 | 0.852 |
| | Fridge | single | 37.41 | 0.627 | 0.647 |
| | | 1 for all | 30.74 | 0.754 | 0.665 |
| | Washer | single | 9.77 | 0.021 | 0.591 |
| | | 1 for all | 10.98 | 0.033 | 0.468 |
| | Dishwasher | single | 41.38 | 0.048 | 0.083 |
| | | 1 for all | 41.80 | 0.079 | 0.060 |
| LSTM | Kettle | single | 13.74 | 0.005 | 0.763 |
| | | 1 for all | 16.28 | 0.006 | 0.697 |
| | Fridge | single | 35.72 | 0.863 | 0.545 |
| | | 1 for all | 39.64 | 0.916 | 0.393 |
| | Washer | single | 14.59 | 0.095 | 0.123 |
| | | 1 for all | 20.27 | 0.053 | 0.229 |
| | Dishwasher | single | 33.01 | 0.042 | 0.567 |
| | | 1 for all | 34.19 | 0.044 | 0.535 |
| GRU | Kettle | single | 14.08 | 0.005 | 0.767 |
| | | 1 for all | 28.53 | 0.010 | 0.000 |
| | Fridge | single | 34.21 | 0.842 | 0.590 |
| | | 1 for all | 45.47 | 1.000 | 0.000 |
| | Washer | single | 9.38 | 0.049 | 0.098 |
| | | 1 for all | 9.38 | 0.049 | 0.098 |
| | Dishwasher | single | 39.75 | 0.045 | 0.447 |
| | | 1 for all | 39.75 | 0.045 | 0.447 |
| ELECTRIcity | Kettle | single | 10.17 | 0.004 | 0.923 |
| | | 1 for all | 15.01 | 0.005 | 0.885 |
| | Fridge | single | 27.61 | 0.764 | 0.730 |
| | | 1 for all | 27.61 | 0.764 | 0.730 |
| | Washer | single | 5.21 | 0.014 | 0.734 |
| | | 1 for all | 4.10 | 0.012 | 0.835 |
| | Dishwasher | single | 24.75 | 0.030 | 0.794 |
| | | 1 for all | 29.09 | 0.030 | 0.775 |

Figure 4.8 Comparison of optimal pruning threshold and performance between optimizing the models for each individual appliance (green) vs jointly (yellow). Optimizing the models on the joint set of appliances (1 for all) leads to subpar optimization.

## 4.4 Conclusions

In this chapter, we have proposed an efficient, performance-aware model optimization framework for edge deployment of NILM models that takes into account the edge device characteristics. We have explored three different deployment limitations, for which optimization of different model aspects is required. Additionally, we proposed an objective model optimization metric and a performance-aware model complexity reduction algorithm that constrains model optimization on performance loss. Experimental results validate that our proposed method to bind model performance with model compression, instead of performing it arbitrarily, allows for the combined utilization of more than one compression approaches on the same model without significantly affecting model performance, thus enabling the efficient deployment of NILM models on edge devices.

# Chapter 5

# ContiNILM: A continual learning scheme for Non-Intrusive Load Monitoring

## 5.1 Introduction

Non-Intrusive Load Monitoring (NILM) describes the extraction of the individual consumption pattern of an appliance by analyzing the aggregated consumption [59]. Recent works have utilized state-of-the-art deep learning approaches to tackle the NILM problem [18, 24, 103], remarkably advancing the maturity of NILM applications. However, the performance of NILM algorithms is heavily dependent on the underlying data, which can present significant power distribution deviations due to different contextual characteristics between houses [15, 104]. It can therefore be argued that the trained models should not remain static, but be continuously retrained to adapt to new incoming data conditions.

Continual learning has emerged as a popular research domain in research years to address deep learning model adaptability issues [105–108]. Continual learning broadly refers to the process of a model continuously learning over time across varying domains, tasks, or data distributions [109]. The performance of a NILM model may differ from household to household, or showcase post-deployment concept drift, a phenomenon where the statistical properties of the target variable change over time in unforeseen ways. Therefore, the integration of continual learning techniques in NILM models is an essential step to address performance issues stemming from alterations in the appliance consumption pattern.

However, continual learning approaches suffer from three main limitations [109]. The most commonly observed issue is catastrophic forgetting, which relates to the case where adapting the model to learn the characteristics of the new data distribution results in poor performance in previous data [110–112]. In addition, a naive approach for continual learning would be to jointly store old and new data and train a model on the enhanced dataset from scratch. However, this solution assumes infinite storing space for data, as well as abundant computational resources, thus making this approach impractical and, sometimes, infeasible [113, 114]. Finally, some approaches suggest training a separate model for each incoming data distribution, which is inefficient as knowledge accumulated from past tasks would not be reused in future ones. To overcome the aforementioned limitations, in this Chapter we propose ContiNILM [115], a continual learning scheme for NILM. Our main contributions can be summarized as follows:

- We propose a novel NILM retraining mechanism that improves model performance on new datasets without notably affecting performance on previous ones, thus mitigating the risk of catastrophic forgetting.
- We introduce the concept of experience replay memory as an alternative dataset structure for NILM to improve NILM model retraining in a resource-efficient manner. The experience replay memory stores a fixed number of samples from the superset of training data and new data, mitigating the risk of requiring an infeasible amount of resources to retrain NILM models.
- We present a model performance evaluation framework according to temporal/seasonal criteria and implement contextual importance sampling, a technique to efficiently decide which samples from the incoming data should be stored in the experience replay memory and which should be discarded. This context-aware importance sampling mechanism improves resource efficiency and model adaptation to new conditions.

## 5.2 Methodology

In a NILM framework [13], the total power consumption $x$ at a given time $t$ is $x(t) = \sum_{a=1}^{M} y_a(t) + s_{noise}(t)$, where $s_{noise}$ describes a noise term, and the model is trained to estimate the appliance consumption pattern of a selected domestic appliance $y_a$, given the aggregate power signal $x$. To extend the definition of NILM in a continual learning setting [116], we consider that the model has been trained and tested on dataset $D_0 := \{D_{0,train}, D_{0,val}, D_{0,test}\}$ and, after deployment, will be exposed to additional sequentially arriving datasets $\{D_1, \ldots, D_T\}$, where $D_i := \{\mathbf{x}, \mathbf{y}\}\downarrow i \to (1, \ldots T)$. Therefore, $D_i$ arrives before $D_{i+1}$, without having any assurances on the data distribution correlation with previous or future datasets. To avoid catastrophic forgetting, the goal of a continual learning framework is to reach a performance level $P_i$ on dataset $D_i$ without inducing a negative impact on the model's performance in previous datasets $(D_0, \ldots D_{i\searrow 1})$. Therefore, an equilibrium between model performance on the current dataset and generalization capabilities over previously acquired knowledge is required. However, we argue that, in contrast to other deep learning domains, continual learning for NILM should tackle catastrophic forgetting in a more data-efficient manner. The energy data produced from household electric consumption is vast and may showcase variations depending on many parameters. Therefore, in our approach, we adapt the general definition of continual learning to suit NILM problem characteristics. An overview of the proposed continual learning framework for NILM is illustrated in Figure 5.1. ContiNILM comprises of three parts:1) experience memory replay 2) contextual importance sampling and 3) model retraining mechanism. In the following Sections, we describe in detail the functionality of each component.

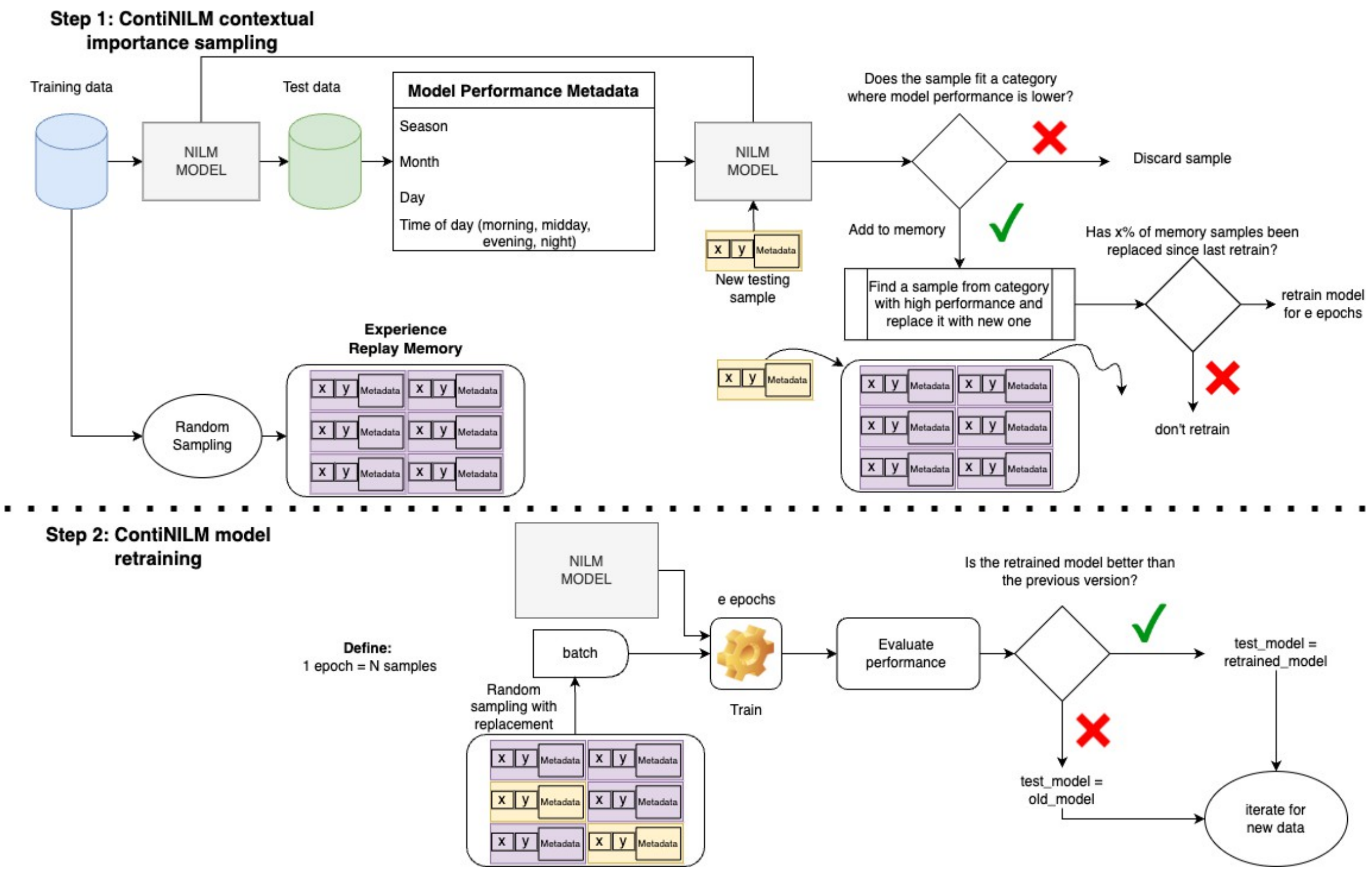

Figure 5.1 The proposed continual learning scheme for NILM. ContiNILM comprises of 3 steps: 1) initialization of experience replay memory and contextual evaluation of model performance, 2) contextual importance sampling and 3) model retraining.

### 5.2.1 Experience Memory Replay

Experience replay is a concept widely used in Reinforcement Learning [117] as an alternative data collection mechanism. The experience replay memory can be considered as a buffer of finite memory size that contains data samples that will be used for model retraining. A sample is defined as an input sequence of the deep learning model. In the beginning, the experience replay memory is filled with random samples from the initial training set $D_{0,train}$. As new samples from datasets $D_1, \ldots D_T$ arrive, each sample is evaluated based on the chosen contextual performance categories $C$ (see Section 5.2.2). If the current sample belongs to a contextual category where the NILM model is underperforming and the sample contains a ground truth or false activation, then the sample will be added to the experience replay memory. A sample already in memory that belongs to a different contextual category must be discarded to add the new sample, under the constraint that all previously encountered datasets $(D_1, \ldots D_{i-1})$ are represented. With this approach, we can create a dataset that exemplifies all data distributions while keeping the dataset size constant. This procedure is presented in Figure 5.1.

### 5.2.2 Contextual importance sampling

NILM model performance is heavily dependent on temporal and seasonal variations. Therefore, to gain more insights on how the model behaves, we define a set $\mathbf{C} = \{C_1, C_2, \ldots, C_c\} \downarrow k \rightarrow (1, \ldots c)$ that denotes the number of contextual performance categories chosen for model evaluation (see Figure 5.1).

Contextual performance categories should reflect different temporal groups, like for example season (winter vs summer), weekday or day part. An unsupervised learning algorithm is utilized to form the contextual performance categories. Then, each dataset $D_i$ can be rewritten as:

$$D_i = \{d_{i,1} \swarrow d_{i,2} \swarrow \dots \swarrow d_{i,c}\} \tag{5.1}$$

where $d_{i,k}$ is defined as $d_{i,k} = \{\dots(x_i^k, y_i^k)\dots\}$ with $(x_i^k, y_i^k) \rightarrow C_k$. Using Equation 5.1, we can now evaluate model performance on each subset and gain insights on whether the model is underperforming in the k-th contextual category out of c available categories, as shown in Equation 5.2

$$E\,(i,k) = \|f(x_i^k; w) \searrow y_i^k\|_2 \tag{5.2}$$

where $E\,(i,\ k)$ denotes the performance error of the NILM using weights $w$ over a subset $d_{i,k}$. To avoid catastrophic forgetting and, at the same time, improve model performance on the current data distribution, the NILM model should be retrained on a dataset that represents the data distribution $D_i$, as well as previously encountered data $(D_0,\ \dots D_{i\searrow 1})$. Thus, we need to estimate an updated weight set $w'$ that better describes both the current and the previously encountered data distributions.

### 5.2.3 Continual Learning Algorithm

When a fixed percentage of memory samples have been replaced with new ones, the model retraining mechanism is triggered. The model with then be fine-tuned with data from the experience replay memory. A key difference in our approach to traditional model retraining methods is that each epoch does not correspond to a full forward and backward pass of the retraining data. Instead, samples from the experience replay memory are sampled randomly with replacement to break the correlations between data samples. After model retraining has been completed, then model performance evaluation is conducted not only on the current dataset $D_i$, but also on previous ones to determine whether catastrophic forgetting has occurred. If model performance is not better than the previous iteration of the model, the process is iterated for new incoming datasets to improve performance.

## 5.3 Experimental setup and results

To validate our methodology, we utilize the UK-Dale dataset [54] and consider the washing machine and dishwasher as disaggregation appliances, since these appliances are commonly subject to contextual changes. House 1 (H1) is treated as dataset $D_0$ and we trained various models using both CNN ([38, 118]) and LSTM-based architectures ([18, 57]), which are contextually evaluated on $D_0$. We then assess model performance at the different parts of the day (night, morning, midday and evening), by splitting each day into four 6-hour periods. Mean Absolute Error (MAE) and F1-Score are used as performance evaluation metrics. Through the contextual performance evaluation on $D_{0,test}$ we discover that model performance is lower on middays (12.00-18.00) or evenings (18.00-24.00). The experience replay memory is then initialized with random samples from $D_{0,train}$ and a memory size of 10,000 samples. We then take data from House 2 (H2) and split it into two parts to create $D_1$, $D_2$. The reason for this choice is to assess how ContiNILM can be utilized to deploy a model in an unseen house and improve disaggregation performance with minimal retraining effort. H2 contains significant variations from H1, as the appliance activation pattern is different. We trigger ContiNILM's retraining mechanism for the best-performing model once 20% of the samples in the experience replay memory has been replaced and fine-tune the model for 20 epochs. $D_2$ is kept outside the scope of ContiNILM and is used as an external validation set. Tables 5.1 and 5.2 present model performance of the static models and ContiNILM. We have implemented two different versions of the ContiNILM model; one with convolutional (Table 5.1) and another with LSTM layers (Table 5.2) to prove the robustness of our proposed method across different architectures. First, we observe that ContiNILM has improved performance on both new datasets ($D_1$, $D_2$). In the CNN case, MAE is ↖ 38% lower than before retraining, while F1-Score showcases an average performance increase of 36.9%, whereas for the LSTM model MAE and F1-score improve by 51.04% and 34.57% respectively. In addition, ContiNILM doesn't suffer from catastrophic forgetting, since performance on H1 remains on a similar level as before retraining. The aforementioned observations are validated by comparing the appliance activations between a static CNN [38] and ContiNILM on all houses, as shown in Figure 5.2. We observe that in previously encountered datasets (H1), ContiNILM avoids catastrophic forgetting, whereas in new datasets the data-efficient retraining mechanism significantly improves disaggregation performance and reduces the probability of false activations.

| Appl. | House | CNN-based Model | MAE | F1 |
|---|---|---|---|---|
| *Washing Machine* | House 1 | Zhang et.al.[38] | 24.65 | 0.857 |
| | | Massidda et.al.[118] | 52.20 | 0.775 |
| | | ContiNILM | 29.30 | 0.853 |
| | House 2.1 | Zhang et.al. [38] | 22.90 | 0.335 |
| | | Massidda et.al. [118] | 17.65 | 0.262 |
| | | ContiNILM | 5.18 | 0.690 |
| | House 2.2 | Zhang et.al. [38] | 24.32 | 0.366 |
| | | Massidda et.al. [118] | 17.97 | 0.277 |
| | | ContiNILM | 6.11 | 0.6761 |
| *Dishwasher* | House 1 | Zhang et.al. [38] | 24.72 | 0.859 |
| | | Massidda et.al. [118] | 47.25 | 0.658 |
| | | ContiNILM | 32.79 | 0.7728 |
| | House 2.1 | Zhang et.al. [38] | 22.37 | 0.631 |
| | | Massidda et.al. [118] | 37.11 | 0.624 |
| | | ContiNILM | 6.60 | 0.8127 |
| | House 2.2 | Zhang et.al. [38] | 20.08 | 0.672 |
| | | Massidda et.al. [118] | 36.41 | 0.603 |
| | | ContiNILM | 6.56 | 0.760 |

Table 5.1 Comparisons between static models with convolutional layers and our proposed continual learning approach

| Appl. | House | LSTM-based Model | MAE | F1 |
|---|---|---|---|---|
| *Washing Machine* | House 1 | Kaselimi et.al.[18] | 77.41 | 0.661 |
| | | Rafiq et.al.[57] | 100.83 | 0.693 |
| | | ContiNILM | 75.85 | 0.711 |
| | House 2.1 | Kaselimi et.al.[18] | 28.45 | 0.188 |
| | | Rafiq et.al.[57] | 12.54 | 0.176 |
| | | ContiNILM | 10.79 | 0.372 |
| | House 2.2 | Kaselimi et.al.[18] | 24.40 | 0.295 |
| | | Rafiq et.al.[57] | 14.53 | 0.228 |
| | | ContiNILM | 12.06 | 0.364 |
| *Dishwasher* | House 1 | Kaselimi et.al.[18] | 52.54 | 0.784 |
| | | Rafiq et.al.[57] | 57.25 | 0.788 |
| | | ContiNILM | 46.28 | 0.771 |
| | House 2.1 | Kaselimi et.al.[18] | 26.99 | 0.604 |
| | | Rafiq et.al.[57] | 27.59 | 0.591 |
| | | ContiNILM | 16.58 | 0.740 |
| | House 2.2 | Kaselimi et.al.[18] | 20.13 | 0.617 |
| | | Rafiq et.al.[57] | 22.57 | 0.608 |
| | | ContiNILM | 9.16 | 0.676 |

Table 5.2 Comparisons between static models with recurrent layers and our proposed continual learning approach

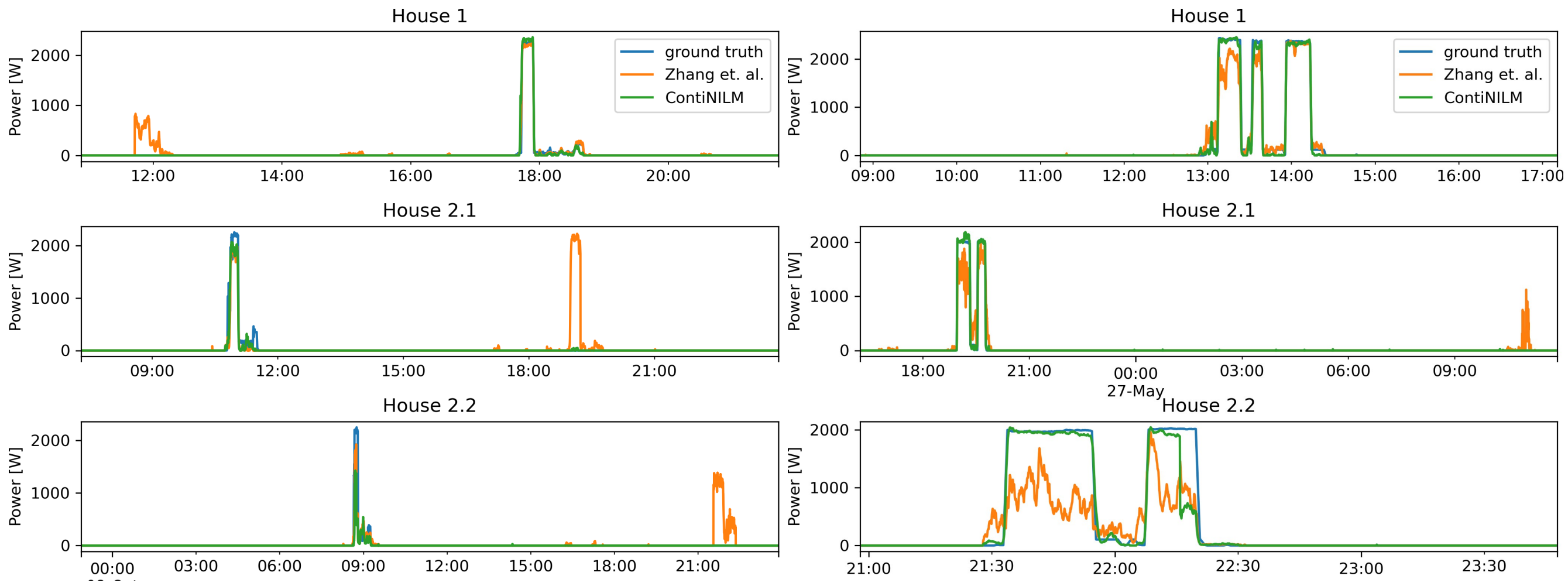

(a) Washing Machine (b) Dishwasher

Figure 5.2 Comparison of appliance activations between a static CNN model and the retrained one using ContiNILM.

## 5.4 Conclusion

In this Chapter, we have presented ContiNILM, a continual learning scheme for energy disaggregation. ContiNILM implements contextual importance sampling and experience memory replay to efficiently choose which samples are useful for model retraining without increasing storage requirements. This addresses computational and data selection issues related to NILM continual learning, further increasing the performance to new contextual conditions compared to traditional static approaches. Experimental results validate that our framework improves performance on new datasets without notably affecting model performance on previous ones.

# Part III

# EV Charging Optimization with Deep Reinforcement Learning

# Chapter 6

# Solar Power driven EV Charging Optimization with Deep Reinforcement Learning

## 6.1 Introduction

The transport sector is the end-user sector with the heaviest dependency on fossil fuels, accounting for 37% of global $CO_2$ emissions according to IEA 2021 Transport report [10]. Due to the COVID-19 pandemic, mobility activity and consequently transport $CO_2$ emissions have temporarily decreased. Despite this downward trend, emissions from transport and especially from road passenger vehicles are expected to continuously rise over the next decades. Even if, there is an increasing number of national energy plans targeting net zero emissions by 2050 and renewable energy share is growing bigger in the global electricity generation mix [119], high short-term gas prices lead to increased coal-fired power generation. Therefore, the challenge of fully decarbonizing vehicle road transport is growing bigger, especially in the short-term where EV technology is under an evolutionary phase and fuel economy regulations are still under development [10].

Local energy flexibility markets can contribute to the aforementioned direction with the exploitation of self-consumed green energy, providing added value not only to the EV owners, but also to the distribution network operators. Controlling residential EV charging during the day with the design of DR schemes can optimally schedule EV charging load so that end-user electricity costs and carbon footprint are being minimized. The introduction of proper incentives can stimulate end-users to shift their electricity consumption from on-peak hours to low-peak periods and, in that way, tackle network-related issues that may rise.

Artificial Intelligence (AI) can unlock the flexibility potential of low voltage power grids, utilizing smart meter readings to identify domestic appliances signatures from aggregated consumption signals [24] or schedule and control residential resources to participate in demand response events [120–122]. Many different methods have been used for scheduling and control of residential energy resources with heuristic algorithms, with particle swarm optimization and genetic algorithm to be among the most common, due to their lower computational requirements and the lack of model training needed [123, 124]. However, Reinforcement Learning (RL) has been recently receiving increased attention in the field of demand response applications, since its dynamic character can better integrate uncertainty aspects, such as end-user preferences consideration, in the problem formulation [125]. In addition, RL can continuously learn from past experiences in a model-free approach, and consequently increase its performance while in operation [126].

The high-dimensional state space in residential resource scheduling and control, especially when considering end-user preferences and PV power self-consumption, has led to an increasing interest in the use of Deep Reinforcement Learning. Solar PV-sourced energy resource scheduling was investigated in [125], where deep Q-learning (DQN) and double deep Q-learning (DDQN) have been compared to PSO on residential resource scheduling. However, EV load scheduling has not been considered in the problem formulation. In [127], various residential appliances have been optimally scheduled with the support of solar PV power. Bidirectional power flows have been considered and end-user preferences have been inferred from measured data, but EV charging load has been neglected in the problem design. PV self-consumption optimization has been included also in [128], where Q-learning, a Deep RL method, has been used to optimally schedule domestic space and water heating. In addition, many works have focused on EV charging load scheduling with Deep RL, when considering end-user driving or charging preferences during the day. However, in the majority of reviewed literature [129–132], EV load charging patterns have been modeled with Gaussian probability density functions (stochastic) or considered known (deterministic) without being inferred from historical data. End-user feedback has been included in a DQN algorithm's rewards [133] to minimize electricity bills and user discomfort. However, clean energy prioritization has not been included in the RL environment that models the energy system. In [134], EV charging optimization based on historical data-inferred end-user preferences has been conducted, but renewable power self-consumption has not been prioritized.

In this Chapter, we propose a novel framework to optimally charge an EV, while prioritizing PV power consumption and cost minimization, with the use of DQN Reinforcement Learning [135]. Our proposed approach:

- Prioritizes solar PV self-consumption for residential EV load scheduling. A solar utilization index is employed to calculate the amount of clean energy that has been self-consumed for EV charging.
- Suggests a flexibility potential index that is introduced in the RL rewards to take into consideration end-user preferences. This index is inferred from analyzing historical consumption data to calculate the average probability for a user to charge the EV at a specific time interval.
- Considers EV battery's technical specifications and daily driving habits through the design of two rewards that are integrated in the RL environment.

The Chapter is structured as follows. In Section 6.2, the problem formulation and modelling, both from the energy and the RL perspectives are introduced. In Section 6.3, the experimental setup and the evaluation results are presented, whereas in Section 6.4 the main conclusions of the proposed methodology are drawn.

## 6.2 Methodology

The EV load scheduling optimization problem is tackled by examining residential EV owners that have solar PV panels installed on their premises. Therefore, the energy system can be modeled as an individual household connected to the main grid that includes a Battery Electric Vehicle (BEV), an EV home charger, PV panels for solar power generation as well as a residual load comprising of the cumulative power consumption of the remaining domestic house appliances. The real-time EV load scheduling can be formulated as a discrete timestep optimization problem with a 15-minute temporal resolution, where the proposed model aims to optimally distribute the charging load throughout the day. A 24-hour modeling horizon ($T = 96$) is considered and, at each discrete timestep $t$, the model should assess whether charging the EV at the given timeslot would be beneficial for the end-user, as well as estimate the required power amount based on techno-economical criteria without jeopardizing user convenience. In this work, user convenience is formulated through a flexibility potential index which considers the within-the-day EV charging potential. The index is inferred through historical data analysis on real household measurements from Austin, Texas, US from the Pecan Street dataset [136]. The flexibility index profile of an indicative household is illustrated in Figure 6.1.

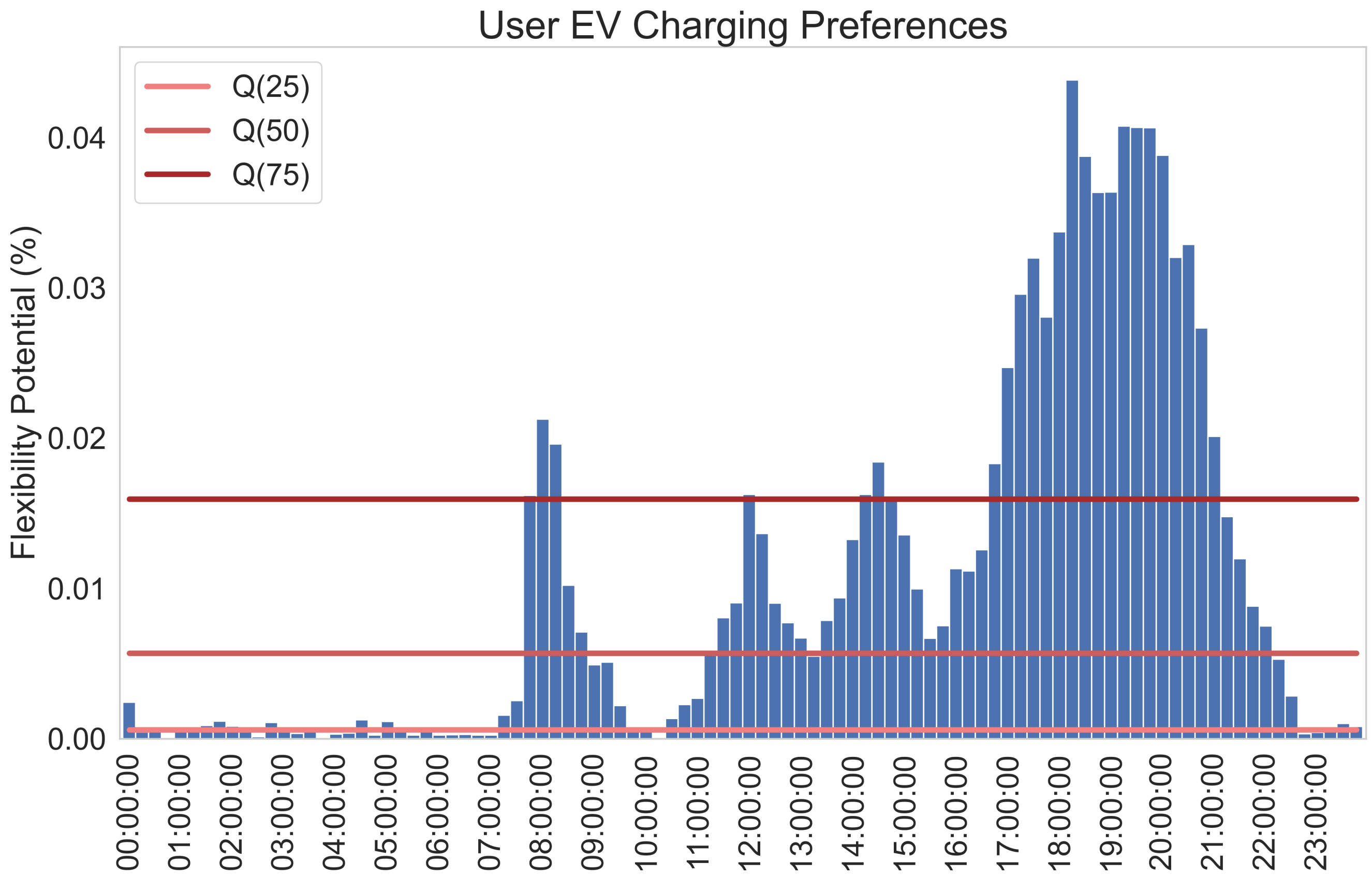

Figure 6.1 Flexibility potential index for household 4373 in Austin, Texas. The index at each 15-minute time step shows the consistency of EV charging throughout the day (%).

In the context of Deep Reinforcement Learning, the optimization task must be defined as an environment-agent duality. The environment produces observations and the Deep Neural Network driven optimizer (agent) evaluates the given observation and chooses an action, for which it is positively or negatively rewarded. Therefore, the EV load scheduling task is formulated as a Markov Decision

Process (MDP) defined by the tuple $(S, A, R, P)$. $S$ denotes the state space, i.e. the observations that will be evaluated by the agent to choose the optimal action. The set $A = \{0, 1\}$ contains the possible actions that the agent can choose from, meaning that the agent can choose to either charge or not charge the EV, for any given timestep $t$. After the agent chooses an action, the action is evaluated by the environment depending on the defined criteria and will assign the respective reward from the set $R$. Then, the agent, which in our approach is modeled as a Deep Q-Network (DQN) [137], receives the next state $s'$ and the same process is iterated through the training phase. Finally, set $P$ contains the probability that action $a$ in state $s$ at timestep $t$ will lead to state $s'$ at timestep $t+1$. This mechanism allows the agent to assess which actions maximize its rewards for a given state $s$, i.e. learn the optimal state-action pairs $(s, a)$. Figure 6.2 visualizes the aforementioned approach. The following subsections present a detailed description of the structure of each sub-component of the DRL model.

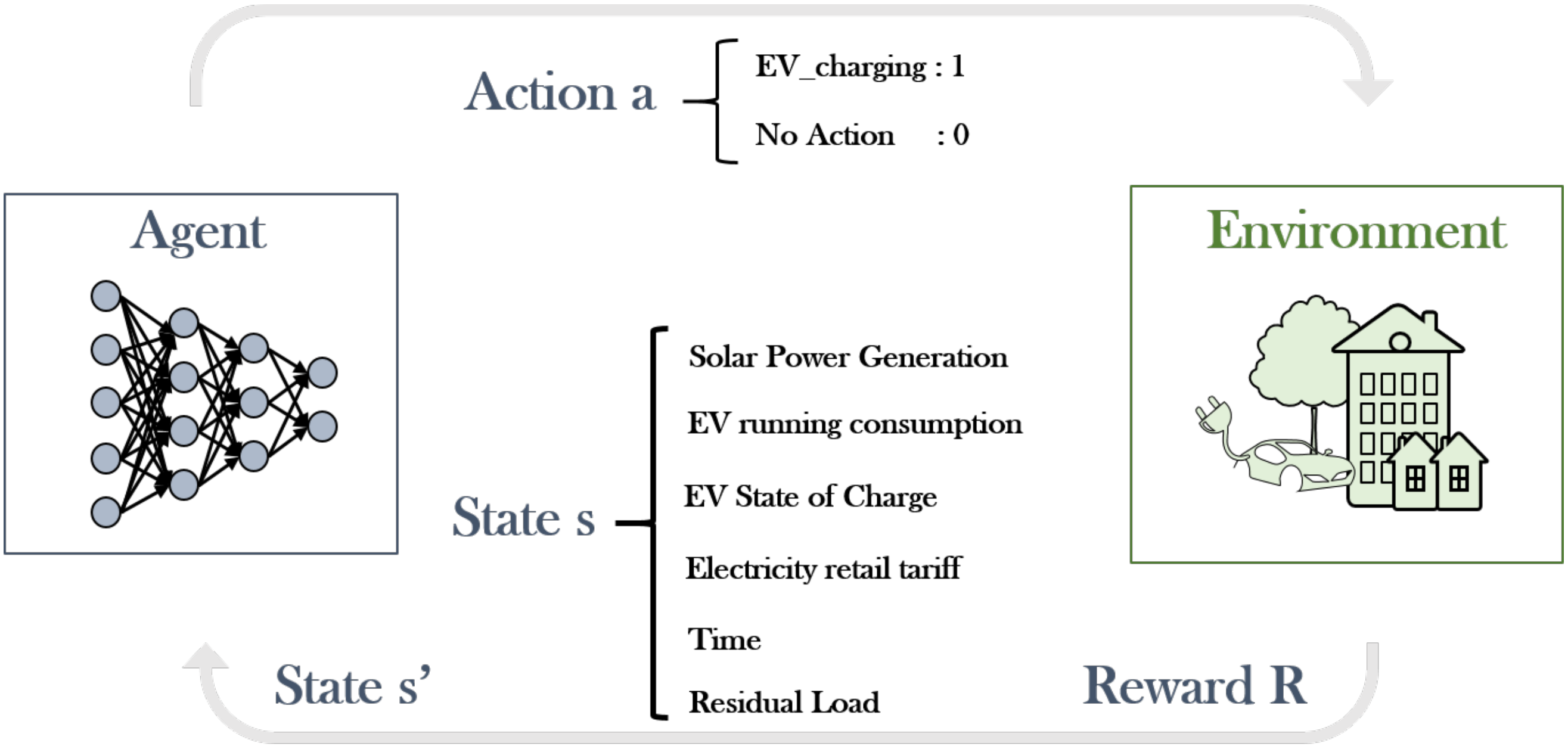

Figure 6.2 High-level overview of the DQN framework for Demand Response. The agent utilizes a neural network to evaluate an observation state of the environment and choose the corresponding action. In return, it receives a reward based on the optimality of its action.

### 6.2.1 State

At each timestep $t$, the optimization agent will receive a vector $s_t = (p_t, P_t^{PV}, P_t^{non\setminus EV}, P_{t,run}^{EV}, SoC_t, t)$ from the environment. Therefore, each state $s$ contains the following information: (1) $p_t$ signifies the Time-of-Use (ToU) electricity tariff at the given timepoint, depending on the period of the day (On-Peak, Mid-Peak, Off-Peak); (2) $P_t^{PV}$ denotes the solar power generated by the PV panel; (3) $P_t^{non\setminus EV}$ describes the residual (non-EV) consumption load; (4) $P_{t,run}^{EV}$ indicates the cumulative EV power consumed from the start of the episode ($t=1$) until the current time step $t$; (5) $SoC_t$ shows the State of Charge (SoC) of the EV at time step $t$ and (6) $t$ contains information about the current time step.

In our approach, episodes are formulated by splitting the data on a daily basis. At the start of each episode, the environment utilizes historical consumption data to calculate the amount of power that was required to charge the end-user's EV ($P_{day}^{EV}$). It is assumed that the consumed power corresponds to a full charging cycle, since the number of charging cycles the BEV has undergone during a single day is uncertain. Therefore, the starting SoC, $SoC_{start}$, is defined as:

$$SoC_{start} = 1 - \eta \frac{P_{day}^{EV}}{4E_{batt}} \tag{6.1}$$

where $\eta$ is the battery pack charging efficiency and $E_{batt}$ is the rated battery capacity, measured in kWh. Through Equation 6.1, it is ensured that the optimized EV daily power consumption, i.e. after the load shifting procedure, will remain close (±5%) to the original consumption $P_{day}^{EV}$.

$$P_{day}^{EV} \in [0.95 \sum_{t=1}^{T} P_t^{EV}, 1.05 \sum_{t=1}^{T} P_t^{EV}] \tag{6.2}$$

### 6.2.2 Action

As previously mentioned, the agent receives a state $s_t$ and selects to either charge or not charge the EV. The selected action is therefore:

$$\alpha_t^{EV} = \{1, 0\}, \forall \alpha \in A, \forall t \in T \tag{6.3}$$

where $\alpha_t^{EV} = 1$ corresponds to charging the BEV and $\alpha_t^{EV} = 0$ means that, according to the DQN agent, it is better to not charge and remain idle.

In addition, the agent actions must be constrained by the physical and technical properties of the battery pack:

$$SoC_{min} \leq SoC_t \leq SoC_{max}, \forall t \in T \tag{6.4}$$

$$SoC_{t+1} = SoC_t + \frac{\eta P_t^{EV}}{4E_{batt}}, \forall t \in T \tag{6.5}$$

$$P_t^{EV} = \begin{cases} 3.3kW, & SoC_{min} \leq SoC_t \leq 0.9 \\ 1.5kW, & SoC_t > 0.9 \end{cases}, \forall t \in T \tag{6.6}$$

where $SoC_{min}$, $SoC_{max}$ denote the minimum and maximum State of Charge. At a given timestep $t$, $SoC_t$ expresses the State of Charge of the EV according to Equation 6.6, and $P_t^{EV}$ is the charging power consumption.

### 6.2.3 Rewards

Every new state $s_{t+1}$ is dependent on the current state ($s_t$) and the action that the agent will select ($\alpha_t^{EV}$). Each action will be assigned with a reward from the set $R$. Therefore, the rewards are the driving force that allow the agent to evaluate whether each action was correct or not and subsequently learn to optimize its decisions. Since the aim of this work is not only to decide the cost-optimal EV charging strategy, but also to utilize green power without violating the battery's operational constraints and the end-user daily habits, a complex multitude of rewards is required. Therefore, we split the total reward into the following sub-rewards: 24-hour power consumption ($r_1$), user flexibility potential ($r_2$), electricity cost optimization ($r_3$) and BEV SoC control ($r_4$). Each reward is formulated according to:

$$r_1 = \begin{cases} 3, & )_t^{EV} = 1 \ \& \ P_t^{PV} > 0 \ \& \ P_{t,run}^{EV} \Rightarrow 1.05 P_{day}^{EV} \\ 2, & )_t^{EV} = 1 \ \& \ P_t^{PV} = 0 \ \& \ P_{t,run}^{EV} \Rightarrow 1.05 P_{day}^{EV} \\ -10, & )_t^{EV} = 1 \ \& \ P_{t,run}^{EV} \propto 1.05 P_{day}^{EV} \\ -0.25, & )_t^{EV} = 0 \ \& \ P_{t,run}^{EV} \Rightarrow 1.05 P_{day}^{EV} \\ 0, & )_t^{EV} = 0 \ \& \ P_{t,run}^{EV} \propto 1.05 P_{day}^{EV} \end{cases} \tag{6.7}$$

From the aforementioned reward it is evident that the agent is strongly rewarded when charging the EV in time steps with solar power generation to promote green energy utilization and self-consumption that reduces electricity costs.

$$r_2 = \begin{cases} -2, & )_t^{EV} = 1 \ \& \ U_t^{flex} \Rightarrow Q_f(0.25) \\ -1, & )_t^{EV} = 1 \ \& \ U_t^{flex} \Rightarrow Q_f(0.50) \\ 0, & )_t^{EV} = 0 \\ 1, & )_t^{EV} = 1 \ \& \ U_t^{flex} \Rightarrow Q_f(0.75) \\ 2, & )_t^{EV} = 1 \ \& \ U_t^{flex} > Q_f(0.75) \end{cases} \tag{6.8}$$

where $U_t^{flex}$ expresses the user flexibility potential, i.e. the probability that the user would charge their EV at this timeslot. $Q_f(X)$ represents the flexibility potential probability quantile, as illustrated in Figure 6.1. $Q_f(X)$ is the quantile of BEV charging throughout the day, as obtained from historical data analysis on household 4373 of the Pecan Street data set [136]. From Equation 6.8 it is evident that the agent will receive a higher reward if the probability distribution function of the end-user charging habits is followed.

$$r_3 = \begin{cases} 2, & )_t^{EV} = 1 \ \& \ C_t \Rightarrow Q_c(0.25) \\ 1, & )_t^{EV} = 1 \ \& \ C_t \Rightarrow Q_c(0.50) \\ 0, & )_t^{EV} = 0 \\ -1, & )_t^{EV} = 1 \ \& \ C_t \Rightarrow Q_c(0.75) \\ -2, & )_t^{EV} = 1 \ \& \ C_t > Q_c(0.75) \end{cases} \tag{6.9}$$

where the electricity cost $C_t$ is formulated according to:

$$C_t = r_t \cdot ()_t^{EV} \cdot P_t^{EV} + P_t^{non-EV} - P_t^{PV}) \tag{6.10}$$

Figure 6.3 depicts the electricity cost quantiles $Q_c(X)$, as well as the daily average electricity cost for house 4373 of the Pecan Street dataset. For the proper calculation of $Q_c(X)$, negative cost periods have been excluded to avoid outlier values in the reward thresholds. Therefore, Equation 6.9 clearly shows that a higher cost will lead to a lower reward assignment.

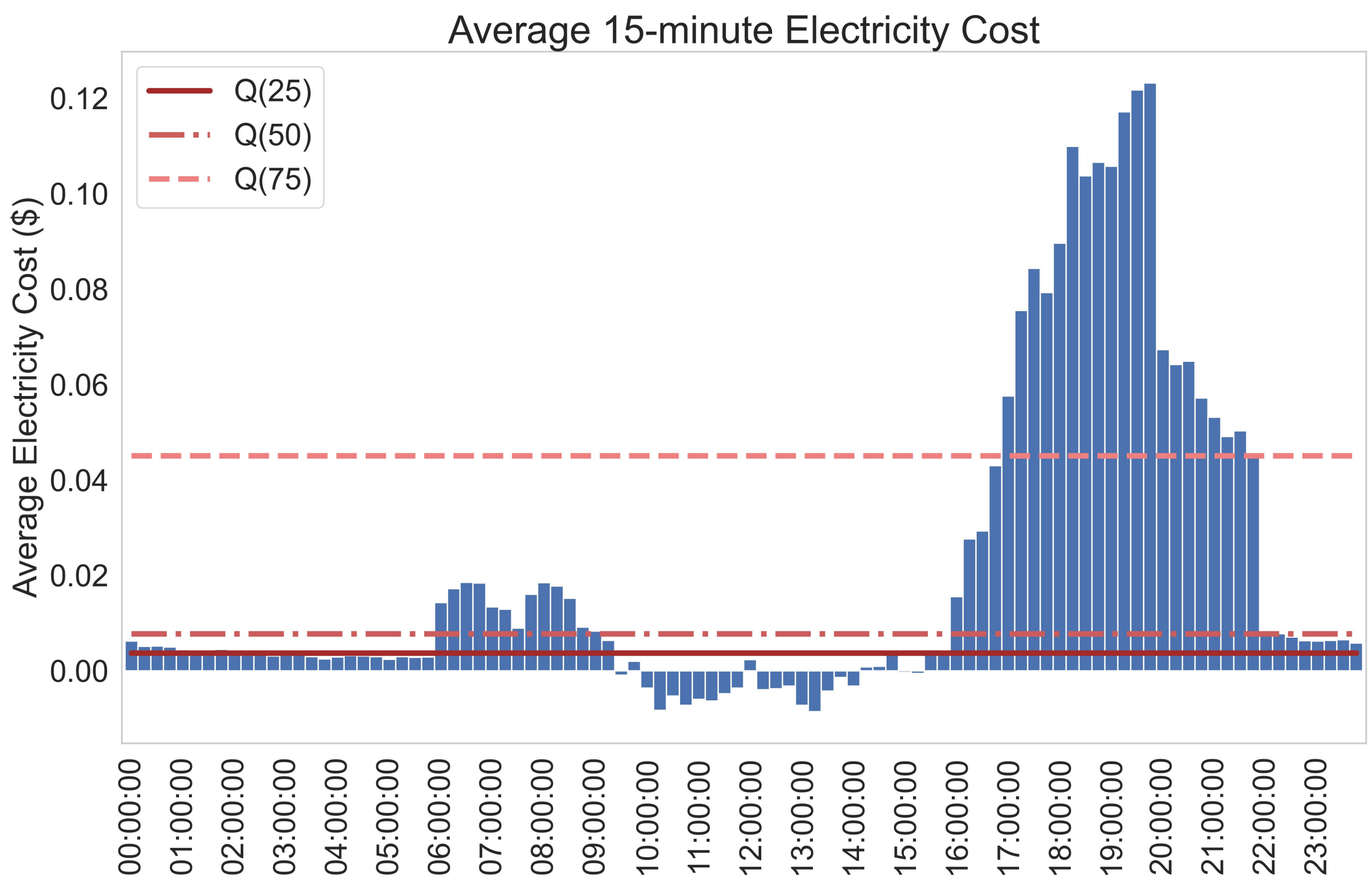

Figure 6.3 Average Electricity Cost for household #4373 in Austin, Texas. The cost quantiles are calculated excluding negative 15-minute cost windows.

Finally, sub-reward $r_4$ monitors the EV Battery SoC and ensures that the agent is strongly penalized for any action that leads to exceeding the battery capacity ($SoC > 100\%$), potentially damaging its physical components.

$$r_4 = \begin{cases} \searrow 10, & )_t^{EV} = 1 \ \& \ SoC_t \propto 1 \\ 0, & otherwise \end{cases} \tag{6.11}$$

The total reward $r$ that the agent receives is computed as the weighted sum of the previously defined sub-rewards:

$$R = \ni_1 \cdot r_1 + \ni_2 \cdot r_2 + \ni_3 \cdot r_3 + \ni_4 \cdot r_4 \tag{6.12}$$

All weights are considered equal in this work, and therefore the weight factors $\ni_k$ are set to 1.

## 6.3 Experimental Setup and Results

### 6.3.1 Experimental Setup

Real measurements for houses in Austin, Texas, provided by the Pecan Street dataset [136], have been thoroughly analyzed. As described in Section 6.2, the end user consumption habits have been translated into a flexibility potential reward, and the electricity cost has been calculated using actual residential ToU rates [138], which are shown in Table 6.1. The ToU rates are divided into Off-Peak hours (night and early morning) that correspond to low electricity prices, On-Peak hours (afternnon and evening), where the electricity cost is higher and Mid-Peak hours covering the rest of the day. We assume that the end-user EV is a Nissan Leaf, which has a rated battery capacity of 24 kWh and a Level 2 (AC) slow charger with a charging efficiency of 90.5% [139]. The evaluation of our approach is conducted on data from house 4373, as provided by the Pecan dataset [136].

Table 6.1 Weekdays time-of-use electricity tariffs of Austin, Texas households, 2018 summertime

| ToU Period | Hours | Electricity tariff ($/kWh) |
|---|---|---|
| Off-Peak | 00:00 - 06:00<br>22:00 - 24:00 | 0.01188 |
| Mid-Preak | 06:00 - 14:00 | 0.06218 |
| On-Peak | 20:00 - 22:00<br>14:00 - 20:00 | 0.01188 |

## 6.3.2 Results comparison for optimal EV load scheduling

The proposed DQN agent is trained for 1000 epochs to learn the optimal cost-reducing EV charging policy by utilizing solar power generation. The model is then evaluated using days that have not been included in the training set. Table 6.2 presents the cost savings and the total power (kW) sourced by solar PV for EV charging (solar power utilization index) for each day in the test set. On average, our proposed approach achieves average cost savings of 5.4% with a solar power utilization index of 88.4%. It can also be observed that the cost savings can reach up to 11.5%, while in some days the proposed scheme charges the EV completely with PV-generated power.

Table 6.2 Electricity cost savings on various test set days

| Day | Daily EV Demand [kW] | Cost Savings [%] | Solar utilization [%] |
|---|---|---|---|
| 22/04/2018 | 21.9 | 8.0 | 100 |
| 27/06/2018 | 58.5 | 4.2 | 91.8 |
| 08/07/2018 | 78.0 | 11.5 | 76.9 |
| 12/08/2018 | 10.8 | 5.8 | 69.4 |
| 13/08/2018 | 41.7 | 5.3 | 96.4 |
| 19/08/2018 | 22.2 | -1.6 | 100 |
| 08/09/2018 | 20.7 | 4.6 | 84.1 |
| **Average** | 36.3 | 5.4 | 88.4 |

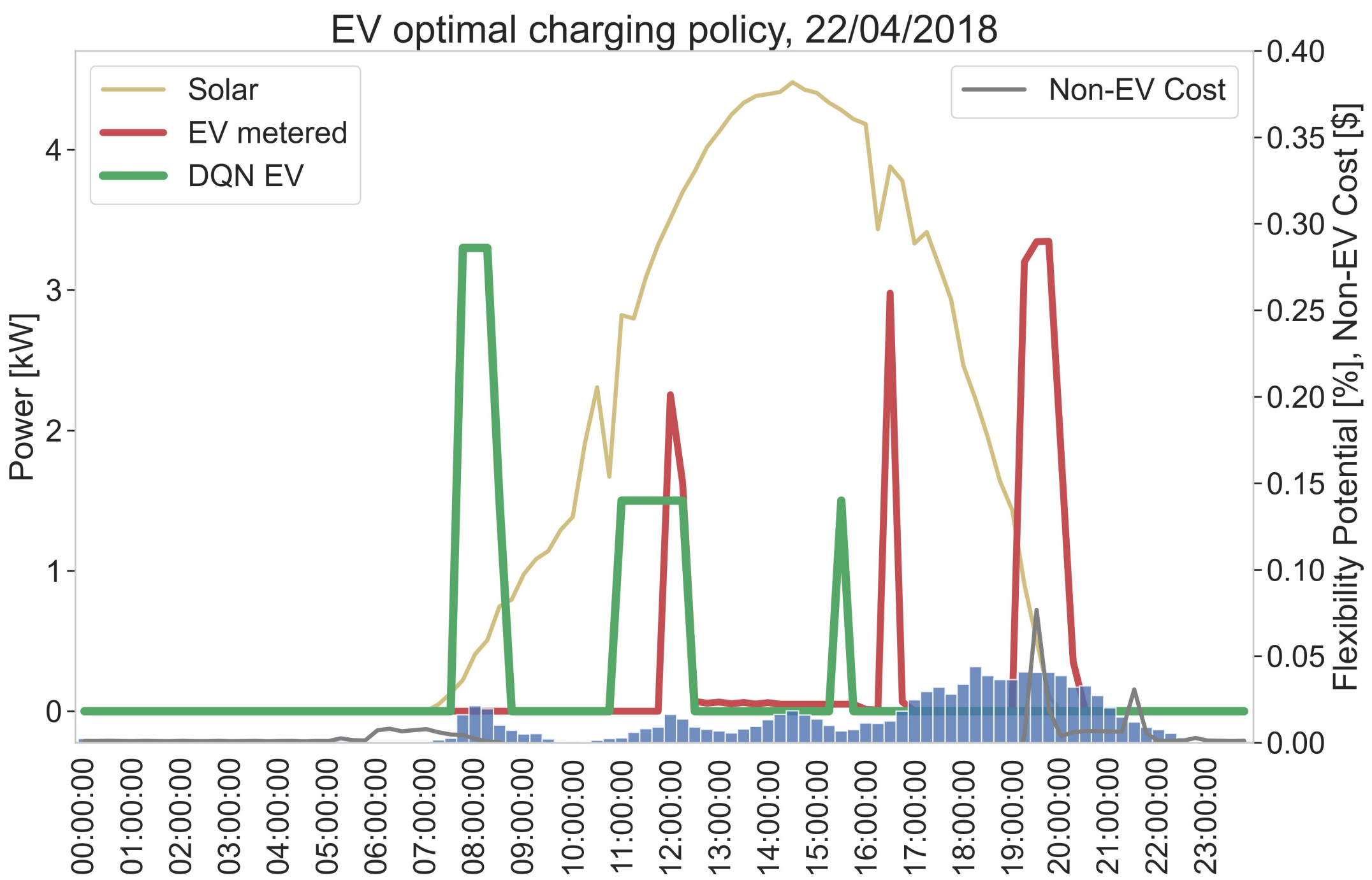

Figure 6.4 Comparison of the metered and the proposed EV load optimization. The suggested policy matches EV charging load with solar power generation hours without jeopardizing the end-user habits.

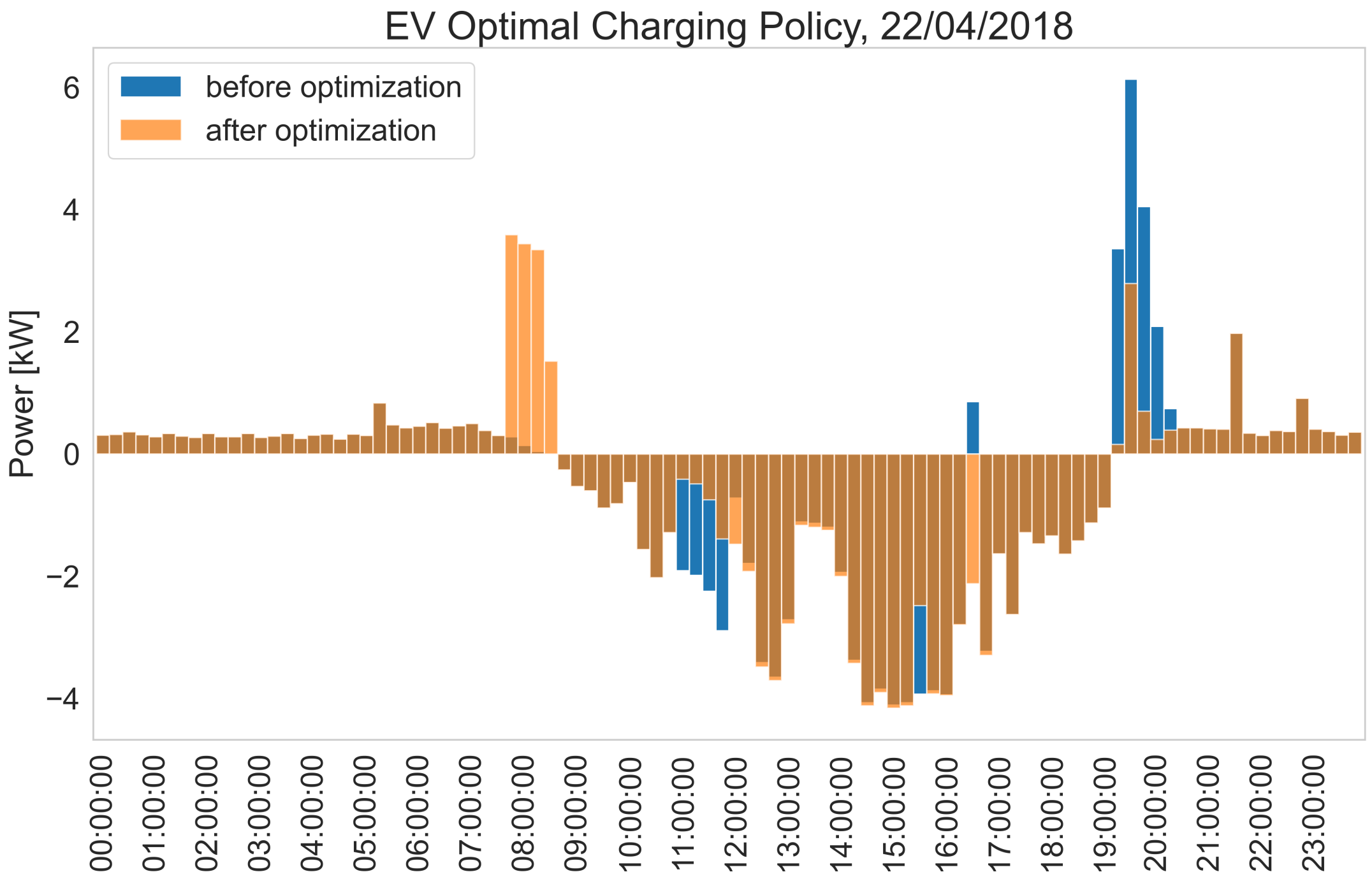

Figure 6.5 Total grid power consumption comparison between the uncontrolled and controlled solutions. Power consumption during On-Peak hours is reduced by completely utilizing solar power generation.

The 22/04/2018 test case displays the ability of the DQN EV proposed algorithm to shift completely EV charging in hours with solar PV generation. The solar utilization index reaches 100% and the electricity cost savings are 8%, as presented in Table 6.2. More specifically, electricity power consumption is being shifted from evening hours to solar power generation periods with low electricity ToU tariffs and high flexibility potential index, as shown in Figure 6.4. In addition, the proposed algorithm reduced peak power consumption from On-Peak hours, shifting demand to hours with self-consumption, as shown in Figure 6.5.

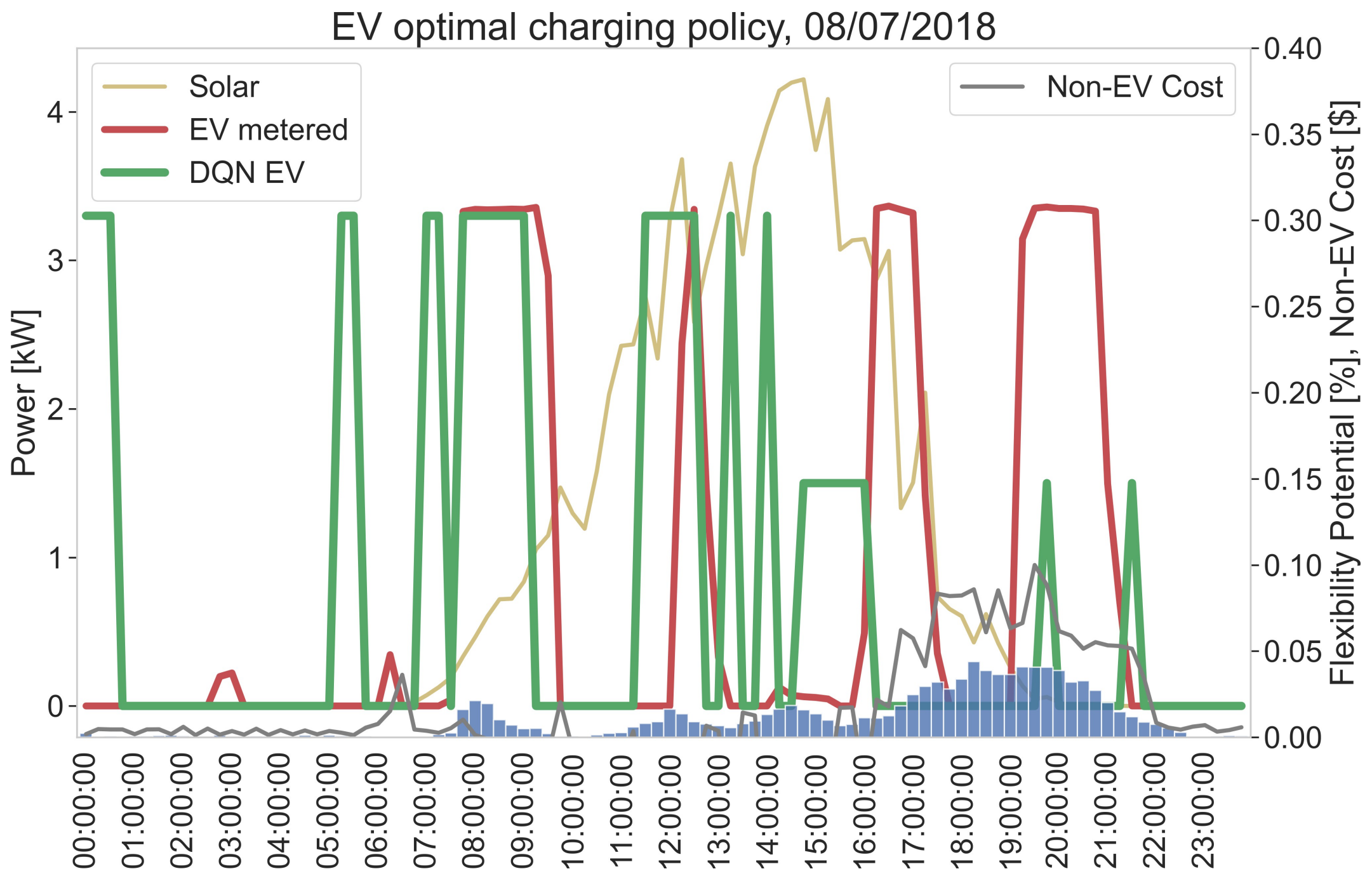

Figure 6.6 Comparison between the EV metered and the proposed DQN EV load optimization solutions. The suggested charging policy shifts EV load to low-cost periods and achieves an electricity bill reduction of 11.5%.

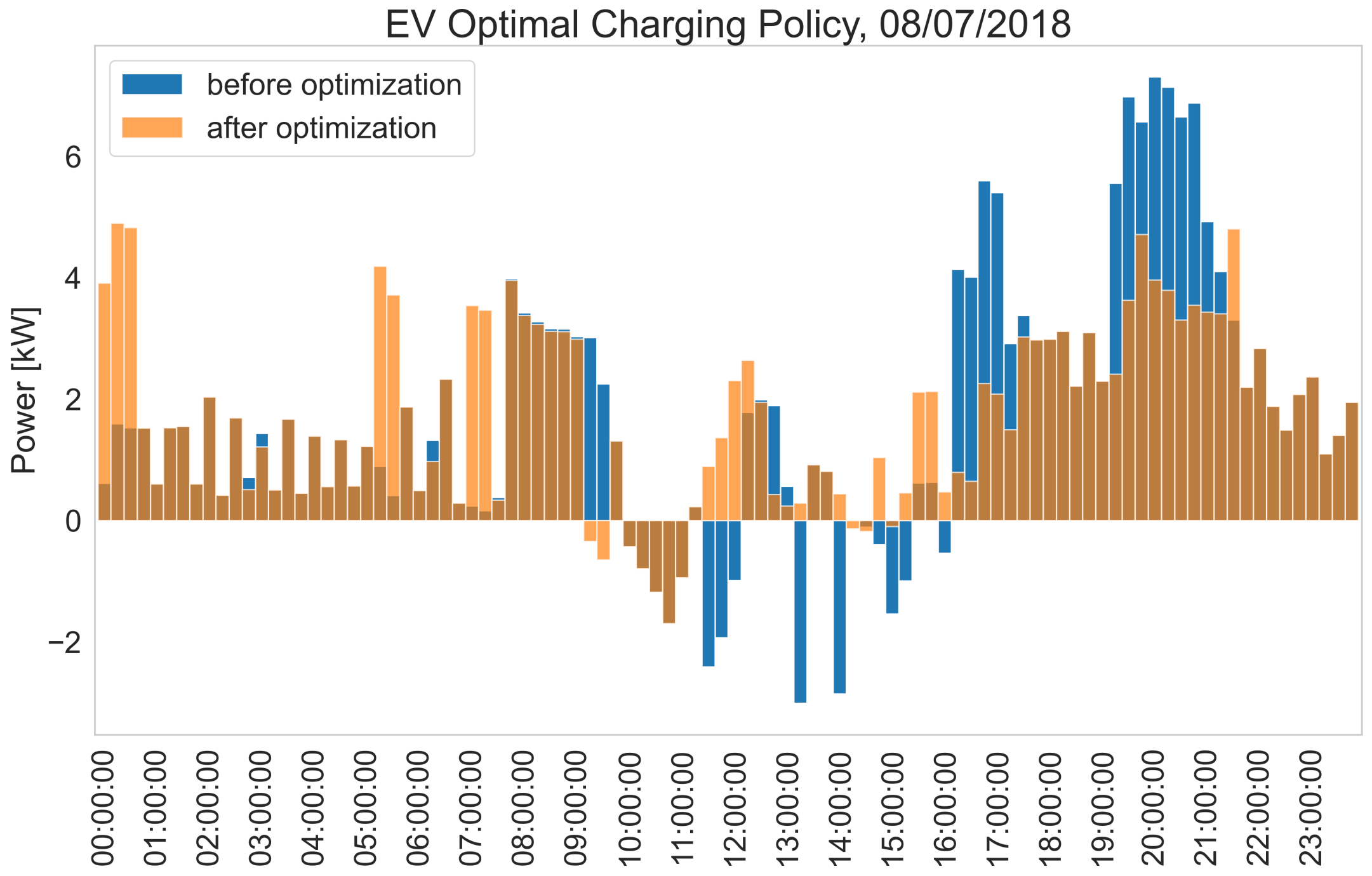

Figure 6.7 Total grid power consumption comparison between the uncontrolled and controlled solutions. Power consumption during On-Peak hours is being reduced in the suggested DQN EV optimal charging policy.

In the test case of 08/07/2018, higher electricity cost savings can be noticed, reaching up to 11.5 %. Even if solar utilization on this day remains considerably high (76.9 %), the EV proposed charging schedule includes hours without any solar PV generation to avoid jeopardizing end user preferences. More specifically, EV charging is being shifted either at night, where electricity tariffs are low, or during hours with high PV power generation, as shown in Figure 6.6. Similarly to the 22/04/2018 test case, peak power consumption is being reduced as seen in Figure 6.7.

## 6.4 Conclusions

In this Chapter, a solar PV-driven EV daily load scheduling model has been proposed and modeled with the use of Deep Q-Networks. A set of comprehensive rewards have been introduced to prioritize solar PV self-consumption, end-user EV charging habits and technical/operational constraints in the optimization process. Real household measurements and corresponding ToU tariffs from Austin, Texas, USA have been used to evaluate the efficiency of the proposed DQN model. Experimental results indicate that the suggested EV charging optimization policy can reduce end users' electricity bill by up to 11.5%. Depending on the daily EV charging demand and the amount of solar PV generation, clean energy utilization for EV charging can reach up to 100%, paving the way towards road passenger vehicle decarbonisation.

# Chapter 7

# A self-sustained EV charging framework with N-Step Deep Reinforcement Learning

## 7.1 Introduction

Transport is the end-use sector with the highest annual growth rate of global $CO_2$ emissions. From 1990 up to 2021 the annual average growth rate has been about 1.7%, followed by an 8% jump in the period 2021-2022, seen as a re-bound effect due to the ease of Covid-19 pandemic restrictions [10]. This unexpected growth showcased even more the necessity for efficient policies implementation and infrastructure development that can accommodate low-emission vehicles and reduce transport sector $CO_2$ emissions. Therefore, electrification of the transport sector constitutes a key strategic decision towards mitigating global emissions in the short-term as well as achieving net-zero emission goals in the long-term. Ongoing policies target public transport electrification and infrastructure development but higher support is needed to accommodate the rapid integration of battery road passenger vehicles to be in line with Net Zero Scenarios by 2050. However, the reduction of transport sector $CO_2$ emissions is based on the assumption that power networks will be decarbonized through the transition from fossil-driven power generation to fossil-free energy sources. This assumption puts the transport sector decarbonization goals at risk, since the share of renewable generation in the energy mix, despite its increase, is still far from fully replacing fossil-fuels [11]. Therefore, it can be assumed that like-for-like replacement of internal combustion engine (ICE) vehicles by EV will provide few environmental benefits until 2030 [12].

On a power grid level, high charging electricity needs due to the increasing penetration of plug-in electric vehicles are expected to stress existing low voltage networks, increasing network demand in peak load periods even further [140]. Avoiding network problems would entail either coordinating optimally network operation and aggregated EV charging activity [141] or applying peak shaving and load shifting techniques. On a household level, behind-the-meter solar PV generation can lead to reverse power flows during midday, high solar yield hours. Even though battery energy storage has been proven an efficient way to tackle this challenge [142], demand-side management for load shifting can be considered a promising alternative for network stress reduction [143]. Therefore, optimally scheduling the operation of flexible loads on a household level, such as EV, can jointly reduce household peak load and potential reverse power flows in the network, leading to a smoothened household consumption profile.

The aforementioned situation stresses the need to find alternative solutions for EV fleet decarbonization. A lot of researchers have investigated the optimization of EV charging load and its distribution to alternative time periods. However, cost reduction is usually the driving motivation behind the optimization, whereas a reduction of peak demand is sometimes seen as a byproduct of their approach. Cost minimization is not necessarily connected to the decarbonization of the EV fleet. We argue that EV charging optimization must be seen as a complicated, multi-objective optimization problem, where reduction of charging cost is only one of the factors that should be taken into consideration. Utilizing distributed energy resources (DER), such as domestic solar photovoltaic (PV) production, for self-consumption purposes can be seen as a key enabler towards a faster transition to a carbon-neutral passenger vehicles fleet.

The main objective of this Chapter is to propose an EV charging framework that promotes solar PV power consumption at a residential level for EV charging, with the use of N-Step Deep Reinforcement Learning [144]. Charging the EV with energy from clean DER not only contributes to the decarbonization of passenger vehicle transportation, but also reduces network stress, since the level of peak power consumption is mitigated. To our knowledge, this is the first attempt to tackle the problem of EV charging optimization from a self-consumption of distributed renewable resources perspective.

## Related Work

The problem of EV charging optimization has been receiving increasing research interest over the recent years, both on a charging station [145, 146], as well as on a residential level. Earlier works utilized model-based approaches to propose optimal EV charging schemes [146–149]. The EV charging problem has been modelled as a cost minimization scenario that was solved with techniques such as linear programming [147–149], whereas uncertainty related quantities have been modelled with random variables (RV) or forecasting models [147, 148] in the experimental setup. However, apart from commanding significant engineering effort during modelling, model-based approaches usually require knowledge over the whole optimization time period to solve the optimization criterion. Therefore, such approaches can work using the predictions of forecasting models, which entails the risk of possible error accumulation [150]. In both cases, the significant variability of household loads is a potential barrier in their application.

To overcome these limitations, more recent approaches have implemented Reinforcement Learning (RL) techniques [132, 151, 152] for EV charging optimization. In Reinforcement Learning, an agent evaluates a set of input data and aims to choose the action that maximizes a reward function. RL algorithms showcase advantages compared to model-based approaches. Due to their model-free nature, they can learn an optimal strategy from experience data and run online, mitigating the need of forecasting models. Popular algorithms, such as Q-Learning, create tables that approximate the action-value function responsible for assessing the quality of the charging schedule. Nevertheless, the optimization goal becomes exponentially harder as the state-action space grows. As a result, discretization of the input states is required, which is not suitable for real-world applications.

Artificial Intelligence (AI) methods have recently started playing a crucial role in various smart grid solutions [120, 121, 123, 124, 153, 154]. Following this trend, EV charging optimization research focus has shifted from model-based and RL approaches to Deep Reinforcement Learning (DRL) [125, 129–131, 134, 155]. Deep Reinforcement Learning is based on the same concept as Reinforcement Learning, but utilizes neural networks instead for approximation of the value function, thus alleviating the need to keep a large state-action table. At the same time, its ability to take advantage of high-dimensional sensory data from a smart home makes it a more attractive approach of EV charging optimization. However, most approaches still consider cost minimization as the sole objective of EV charging optimization, neglecting other factors with potential importance to the end user, such as charging preferences.

A deeper examination of the aforementioned works showcase some further common limitations. First, regardless of modelling approach, EV charging is usually optimized to minimize charging cost [125, 129–132, 134, 147–149, 155], which does not necessarily lead to decarbonization of passenger vehicle transportation. At the same time, solar PV power self-consumption is rarely witnessed, while in many cases EV is not even distinguished from the total household load or is used mainly as a battery storage service [125, 131, 151]. Furthermore, quantities related to uncertainty, such as charging tendencies, are modelled through random variables, which increases the volatility of the approach [129–132, 134, 148, 152]. In addition, most approaches rely on simulation data to showcase the efficacy of the proposed approach [129–132, 146–149, 151, 152, 155]. This limits the utilization of statistical analysis of real-world historical data, which can lead to a more accurate representation of the household consumption profile. Finally, the reduction of network stress through smoothening daily household electricity demand is rarely mentioned and is seen as a side-effect. A summary of the aforementioned approaches can be found in Table 7.1. Overall, even though multiple works have proposed solutions for EV charging optimization, most follow a similar pattern with regards to optimization goals for residential customers, where the advantages of the utilization of DER are usually not taken into consideration. Several challenges still remain to unlock the full potential of smart EV charging.

Table 7.1 Overview of related work for EV charging optimization

| | **Work** | **Optimization Criterion** | **Uncertainty Modelling** | **Energy Data** | **Limitations** |
|---|---|---|---|---|---|
| *Model-based* | Di Giorgio et. al. [147] | Cost<br>Self-consumption | Deterministic | Simulated | Load and micro-generation foresight information requirement. |
| | Lujano-Rojas et. al. [148] | Cost | RV<br>Forecasting | Simulated | Assumes idealized load profiles.<br>Focuses on charging station area. |
| | Liberati et. al. [146] | Grid power exchange | Deterministic<br>RV | Simulated | Focuses on charging station area.<br>PV Power self-consumption is not considered. |
| | Rastegar et. al. [149] | Cost | Deterministic | Simulated | Requires foresight of the household loads over a 24-hour horizon. |
| *RL* | Chis et. al [132] | Cost | RV<br>Forecasting | Simulated | Household loads not considered in the optimization.<br>Requires discretization of input variables. |
| | O' Neill et. al. [151] | Cost | Deterministic | Simulated | Optimization does not focus on EV.<br>State space requires discretization. |
| | Wen et. al. [152] | Cost<br>User Satisfaction | RV | Simulated | Only the price is considered as input variable.<br>EV-specific details not covered. |
| *Deep RL* | Shuvo et. al. [155] | Cost | Historically inferred | Simulated | PV power self-consumption not prioritized.<br>Consumption data simulated based on user activity. |
| | Li et. al. [131] | Cost | RV | Simulated | PV Power self-consumption is not considered. |
| | Li et. al. [129] | Cost | RV | Simulated | Only prices/SoC considered in the input state. |
| | Wan et. al. [130] | Cost | RV | Simulated | Only prices/SoC considered in the input state. |
| | Liu et. al. [125] | Cost | - | Pecan | EV is not distinguished from the total household load. |
| | Ren et. al. [134] | Cost | Forecasting | Pecan | Historical consumption data not used to derive EV charging patterns. |

### Our contribution

In this Chapter, we propose a novel, smart EV charging framework that utilizes N-Step Deep Reinforcement Learning to achieve self-sustainability by sourcing locally-produced solar PV energy for EV charging purposes. Our approach:

- **Promotes solar PV power self-consumption** for EV charging by introducing a solar power utilization reward in the optimization function. This reward significantly increases the amount of solar power utilized for EV charging, therefore contributing to the decarbonization of passenger vehicle transportation.
- **Accounts for temporal fluctuations in house consumption** and EV charging user habits by introducing variability on a weekday basis in the optimization model, allowing the model to propose more accurate charging strategies depending on the household's consumption profile.
- **Reduces network stress and end-user carbon footprint** by smoothening daily electricity demand via local self-consumption and/or via shifting load from on-peak to off-peak hours.

## 7.2 Methodology

This Section describes the problem setup (Section 7.2.1), as well as a) the modeling approach employed to formulate the EV charging optimization problem in a way that enables training Deep Reinforcement Learning models and b) the details of the proposed RL model (Section 7.2.2).

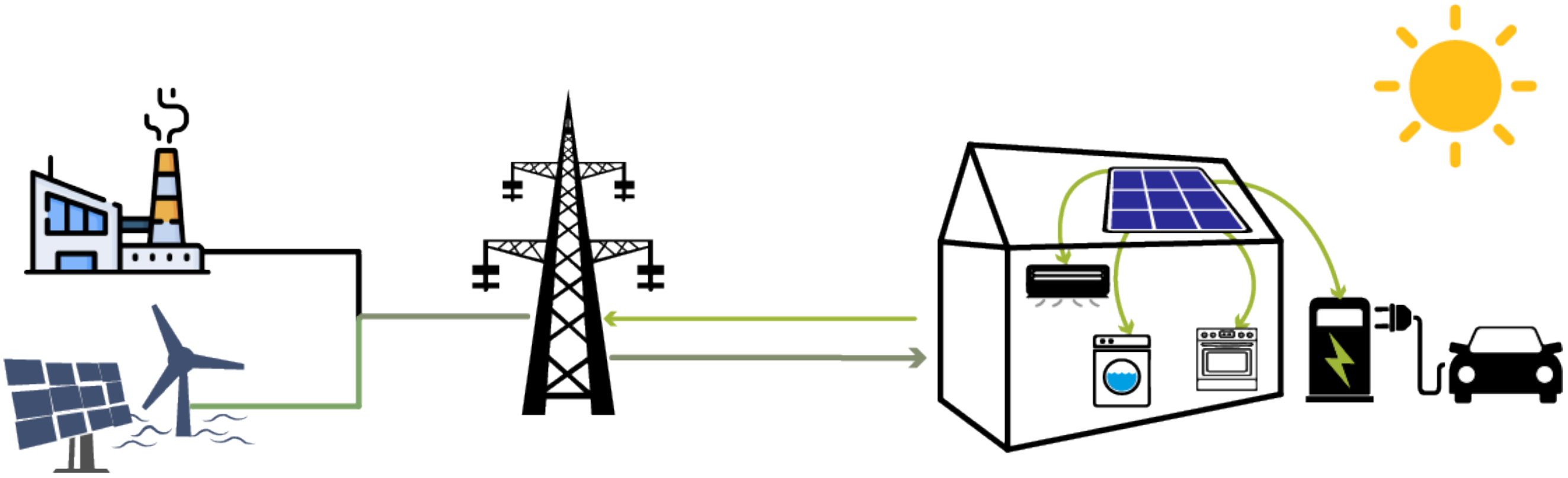

Figure 7.1 Overview of the self-sustained EV charging optimization problem setup.

### 7.2.1 Problem Setup

The EV charging optimization problem is tackled by examining residential EV owners with solar PV panels installed on their premises, as shown in Figure 7.1. Therefore, the energy system can be modeled as an individual household connected to the main grid, with various house appliances and an EV with a slow charger. It is assumed that the house is equipped with a HEMS that can analyze consumption information and shift loads to different time windows through demand-response schemes. This work focuses on the optimization of EV charging, and therefore all other house appliances are considered uncontrolled and constitute the base load of the house. The optimization module can be installed on the household's HEMS to allow for the analysis of the different components. To enable decarbonization of the household's vehicle transportation and, at the same time, reduce stress on the grid when EV adoption increases, utilizing solar power to charge the EV is the priority. However, other major factors that influence end users' behaviour, such as cost or historical charging tendencies should not be neglected. Therefore, self-sustained EV charging can be seen as a constrained optimization problem.

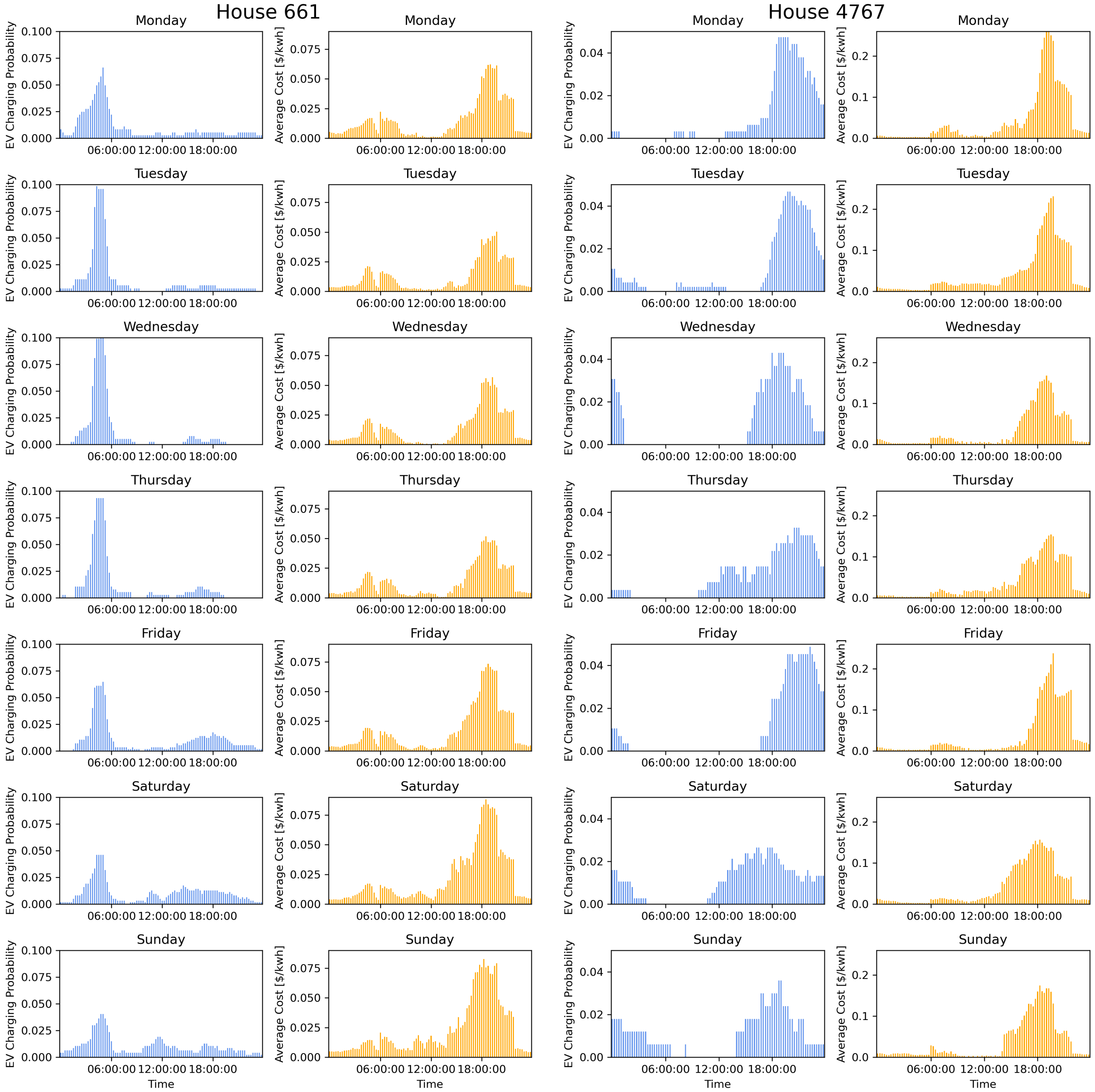

Figure 7.2 EV charging probability and average cost temporal analysis for two households of the Pecan Street dataset.

#### 7.2.1.1 Consumption Variability

End-user EV charging habits and electricity cost play a pivotal role in most EV charging schemes. Even though EV charging is related to the EV charging availability of each user, this information is rarely available. Several works opt to model user habits using stochastic or probabilistic models [129–132], but these methods do not take into consideration the temporal fluctuations of user behavior. The consumption load of the whole house, as well as the charging pattern of the EV can significantly vary depending on the day of the week. An example can be seen in Figure 7.2, where historical data from the Pecan Street dataset [136] have been analyzed to draw two very important results. First, it is evident that end-user charging tendencies may significantly vary between weekdays and that they capture information about the end-user's routine. Second, even though the EV is usually the major consumption load of a household, its charging pattern does not necessarily coincide with time windows that contribute the most to the total electricity bill. This temporal variability will be utilized by the EV charging optimization RL Agent to propose alternative charging policies (see Section 7.2.2.1).

### 7.2.2 Self-sustained EV charging framework with N-Step Deep Reinforcement Learning

Reinforcement Learning (RL) refers to a learning framework where the goal is to learn the optimal decision-making behavior under varying circumstances. The learning process consists of an Environment, which should simulate the optimization goal's mechanics, and an Agent which is trained through continuous interactions with the Environment to learn which actions lead to the maximal reward.

The optimization framework can be formulated as a Markov Decision Process (MDP) defined by the sets ($S$, $A$, $R$), where each set corresponds to a different core mechanic of the procedure [156]. $S$ denotes the State space, which is the set of all possible observable environment conditions that the Agent may encounter. In each step of the process, the Agent may choose an action from the set of allowed actions $A$. The action is passed to the Environment, which transitions to a new state and rewards the Agent with a value from the reward set $R$. An overview of the aforementioned mechanism is illustrated in Figure 7.3. In the following Sections, we describe how the Environment for the self-sustained EV charging framework can be formulated, as well as the N-Step DQN Agent.

#### 7.2.2.1 EV Charging RL-Environment

An RL Environment needs to implement three main functionalities: the state space, the action space and the reward mechanism. In our approach, the EV charging optimization problem is expressed as a discrete timestep optimization problem with a 15-minute temporal resolution, where the Agent's goal is to optimally distribute the EV charging load over a 24-hour ($T = 96$) modelling horizon, while prioritizing drawing energy from the PV to charge the EV. We treat each day as a separate entity to limit the optimization complexity.

**Physical constraints** The optimization process must account for certain physical constraints related to the EV battery and charger. First, since the initial State-of-Charge ($SoC$) at the start of the day is not included in the Pecan Street dataset [136], we assume that the total EV charging load observed throughout a day ($P_{ev}$) corresponds to a full charging cycle. The initial $SoC$ can then be calculated as:

$$SoC_{start} = 1 - \eta \frac{P_{ev}}{4E_{ev,max}} \tag{7.1}$$

where $\eta$ corresponds to the EV battery charging efficiency and $E_{ev,max}$ is the rated battery capacity in kWh [131]. Since a 15-minute temporal reosolution is used and $P_{ev}$ is given in kW, $P_{ev}$ must be divided by 4 to ensure measuring unit consistency. Then, we suppose that the houses are equipped with a slow 2-level EV charger, since it matches the charging pattern in the historical data. Therefore, the EV is charged with different power levels depending on the state of charge (SoC) of the battery.

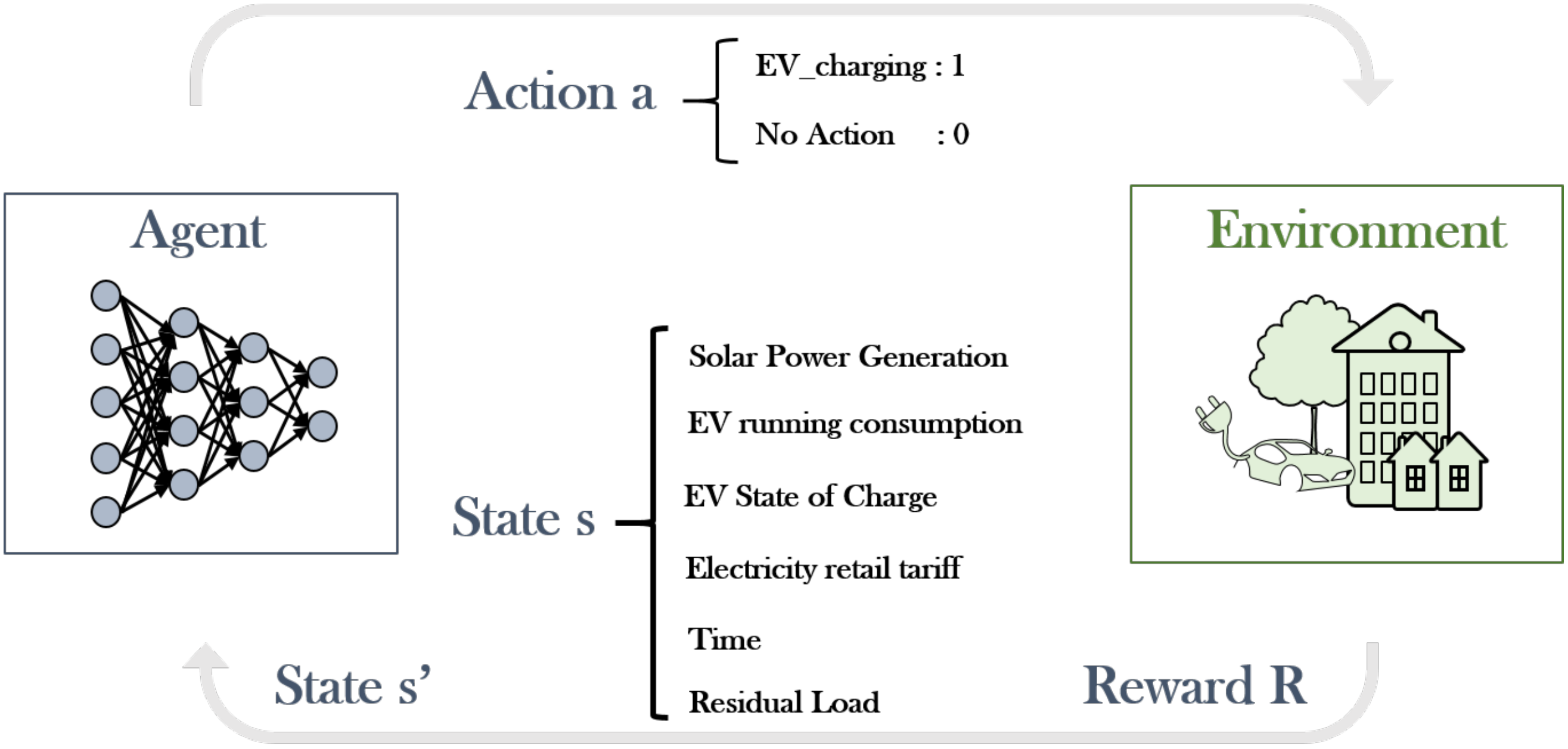

Figure 7.3 Overview of the proposed N-Step Deep Reinforcement Learning framework workflow for self-sustained EV charging.

$$P_{ev,i} = \begin{cases} 3.3\text{kW}, & SoC_{start} \leq SoC_i \leq 0.9 \\ 1.5\text{kW}, & SoC_i > 0.9 \end{cases} \tag{7.2}$$

where indicator $i$ denotes the time step. If the EV is charged, the $SoC$ is updated according to:

$$SoC_{i+1} = SoC_i + \eta \frac{P_{ev,i}}{4E_{ev,max}} \tag{7.3}$$

The smart-metered EV charging data provided by the Pecan Street dataset [136] follow the charging pattern of a 2-level EV charger, they are not discretized and contain minor fluctuations. Therefore, fully matching the dataset load through Equation 7.2 is not always possible. To mitigate this slight limitation, we introduce a relaxation constraint on the simulated EV load and consider a charging cycle successful if the cumulative EV charging load proposed by the model ($P_{ev,sum}$) over a day stays within a 5% tolerance threshold from the historically observed load. This constraint is important for shaping the rewards of the Deep RL model (see Equation 7.8).

$$0.95 \Rightarrow \frac{P_{ev,sum}}{P_{ev}} \Rightarrow 1.05 \tag{7.4}$$

**State Space** In order for the deep RL-Agent to learn the self-sustained EV charging optimization policy, the state $s_i$ should contain all the information required to take an action. In our approach, the environment state at the *i-th* timestep of the day is described with the following information:

- $P_{res,i}$ : Residual (non-EV) power load of the house at timestep $i$
- $P_{pv,i}$ : Solar PV power generation of the house at timestep $i$
- $SoC_i$ : EV state of charge at timestep $i$
- $P_{EV_{sum},i}$ : Cumulative daily load used for EV charging until timestep $i$
- $c_i$ : Time-of-Use (ToU) electricity price at timestep $i$
- $t_i$ : Temporal information about timestep $i$, expressed as $(sin \frac{i}{T}, cos \frac{i}{T})$

where $T = 96$ is the total time steps of the scheduling horizon. Each state vector $s_i$ can consequently be represented as:

$$s_i = (P_{res,i}, P_{PV,i}, SoC_i, P_{EV_{sum},i}, c_i, sin(\frac{i}{T}), cos(\frac{i}{T}))\ \downarrow s_i \rightarrow S \tag{7.5}$$

**Action space** At each discrete timestep $i$ the Agent receives a state observation $s_i$ and evaluates whether the EV should be charged or not. Hence, our approach utilizes a binary action space, where 0 represents remaining idle and 1 means charging the EV, and is mathematically formulated as:

$$a_{ch,i} \rightarrow A := \{0, 1\},\ \downarrow i \rightarrow \{1, \ldots, T\} \tag{7.6}$$

**Reward Shaping** The environment rewards play a pivotal role in Reinforcement Learning, since the Agent's task is to maximize their cumulative rewards over time. Instead of defining one, complicated reward function, it is split into sub-components, where each sub-reward serves a specific purpose. This methodology allows for more precise modelling of the objective function to elicit the desired EV charging behaviour and is given as:

$$r_i = r_{con,i} + r_{hist,i} + r_{c,i} + r_{soc,i} + r_{solar,i} \tag{7.7}$$

The first sub-reward $r_{con,i}$ aims to push the Agent to match the total historically observed load consumption $P_{ev}$ and stay within the acceptable tolerance threshold (see Equation 7.4). $r_{con,i}$ positively rewards the Agent if it chooses to charge the car in timesteps where the total EV load goal has not been met, marginally penalizes it if such timesteps are missed and heavily punishes the Agent if the EV load tolerance threshold is violated.

$$r_{con,i} = \begin{cases} 1, & a_{ch,i} = 1 \ \& \ \frac{P_{ev,sum,i}}{P_{ev}} \leq 1.05 \\ -10, & a_{ch,i} = 1 \ \& \ \frac{P_{ev,sum,i}}{P_{ev}} > 1.05 \\ -0.25, & a_{ch,i} = 0 \ \& \ \frac{P_{ev,sum,i}}{P_{ev}} \leq 1.05 \\ 0, & a_{ch,i} = 0 \ \& \ \frac{P_{ev,sum,i}}{P_{ev}} > 1.05 \end{cases} \tag{7.8}$$

Next, following the analysis conducted in Section 7.2.1.1, the Agent must respect the EV charging practices of the household. However, positively rewarding the Agent for charging at timesteps with high historical charging frequency contradicts with the concept of demand response, since such timesteps only indicate the end user's habits and not their availability. In this work, we have opted for a different approach, where the Agent is penalized if it decides to charge at timesteps with very low historical charging probability, where it is most probable that the user availability in the future is expected to be low. The EV charging patterns of each household are analyzed and a charging probability distribution function $F_{ev}$ is created for each week of the day (see Section 7.2.1.1). $r_{hist,i}$ then penalizes the Agent for charging at timesteps with very low charging frequency by comparing the time step's frequency with the 25th quantile of the charging probability distribution ($Q_{F_{ev},25}$).

$$r_{hist,i} = \begin{cases} -2, & a_{ch,i} = 1 \ \& \ F_{ev,i} < Q_{F_{ev},25} \\ 0, & \text{otherwise} \end{cases} \tag{7.9}$$

The electricity cost is also a major factor in most DR schemes. In our approach, even though the goal is to charge the EV with solar PV-sourced energy, the total cost also plays a significant role to maintain a balance in the resulting policy. The cost $C_i$ at each time step is calculated by multiplying the time-of-use tariff price $c_i$ with the energy that is drawn or given back to the grid, since the houses are equipped with PV panels that can lead to oversupply in high generation hours.

$$C_i = c_i \cdot (a_{ch,i} P_{ev,i} + P_{res,i} - P_{pv,i}) \tag{7.10}$$

To formulate the cost-related sub-reward $r_c$, we create a distribution function $F_c$ that represents the average cost per 15-minute time interval, similar to the user preferences reward (see Figure 7.2, orange-coloured chart). However, in the cost-related sub-reward we positively reward the Agent when charging in low-cost time windows and negatively in higher cost ones by comparing the current price with the respective quantiles, as shown in Equation 7.11.

$$r_c = \begin{cases} 2, & a_{ch,i} = 1 \ \& \ C_i \leq Q_{F_c,25} \\ 1, & a_{ch,i} = 1 \ \& \ C_i \leq Q_{F_c,50} \\ 0, & a_{ch,i} = 0 \\ -1, & a_{ch,i} = 1 \ \& \ C_i \leq Q_{F_c,75} \\ -2, & a_{ch,i} = 1 \ \& \ C_i > Q_{F_c,75} \end{cases} \tag{7.11}$$

In addition, the Agent benefits from a positive reward if it chooses to charge at a timestep with PV power generation, as shown in Equation 7.12. $r_{solar}$ is meant to promote self-consumption for EV charging, but only at timesteps when it is meaningful to do so. To that end, this sub-reward has been structured in a way to not outweigh the other rewards, but complement them and steer the focus of the Agent towards self-consumption of clean energy.

$$r_{solar} = \begin{cases} 2, & a_{ch,i} = 1 \ \& \ P_{pv,i} > 0 \\ 0, & \text{otherwise} \end{cases} \tag{7.12}$$

Finally, $r_{soc}$ acts as a constraint and heavily penalizes the Agent if the $SoC$ exceeds 100%, to ensure that the EV battery is not damaged.

$$rsoc = \begin{cases} \searrow 10, & a_{ch,i} = 1 \ \& \ SoC_i \propto 1 \\ 0, & \text{otherwise} \end{cases} \tag{7.13}$$

#### 7.2.2.2 Deep Q-Learning

Deep Q-Learning (DQN) [137] is a popular algorithm in Reinforcement learning, since the combination of neural networks and Q-learning can perform well in cases where the state space becomes too large [157]. DQN is an off-line algorithm, which extracts the optimal policy u by learning which state-action pairs $(s, a)$ lead to the maximal reward. u is defined as the mapping of environment states to Agent actions, as shown in Equation 7.14, and describes the state trajectory that the Agent follows during optimization.

$$\text{u}(s) := s \rightarrow S\,' \Leftarrow a \rightarrow A \tag{7.14}$$

The policy u is constructed by evaluating the Q-value of each state-action pair $(s, a)$, which is defined as the sum of the immediate reward obtained for action $a$ and the maximum Q-value that can be acquired in the next state $s'$ given Agent model weights θ, discounted by a factor #. This relationship is formulated in Equation 7.15 (Bellman equation). Therefore, for each state, the Agent selects the action that yields the maximum Q-value, as shown in Equation 7.16. To estimate the Q-values of the next state $s'$, a copy of the neural network is utilized (target network), in which the model weights ($\theta^{\leftrightarrow}$) are frozen. The model is periodically updated by replacing $\theta^{\leftrightarrow}$ with θ.

$$Q_{\text{u}}(s, a) = r + \# \max_{a \rightarrow A} Q(s^{\leftrightarrow}, a; \theta^{\leftrightarrow}) \tag{7.15}$$

$$a_{opt} = \underset{a \rightarrow A}{\operatorname{argmax}} Q(s, a; \theta) \tag{7.16}$$

#### 7.2.2.3 N-Step Deep Q-Learning

Equations 7.15 and 7.16, only consider the next state $s^{\leftrightarrow}$ to estimate the optimal Q-value. The estimation performance can be improved and can lead to faster learning, if forward-view multi-step targets are utilized [158]. Therefore, N-Step DQN takes into consideration the next $n$ state rewards to estimate the local ones. Considering this temporal horizon, the Q-value at timestep $i$ is calculated through recursive evaluation of all states up to timestep *i+N-1*, as shown in Equation 7.17. Utilizing "future" Q-values in the assessment of the current state-action pairs leads to a more accurate representation of the value of each action and, therefore, to a better EV charging policy.

$$Q_{u}(s_i, a_i) = r + \sum_{k=1}^{N-1} \gamma^k Q_{u}(s_{i+k}, a_{i+k}) \tag{7.17}$$

## 7.3 Experimental Setup and Results

### 7.3.1 Experimental setup

To validate our approach, we used data from the Pecan Street dataset [136] in Austin, Texas. The houses selected fulfilled the requirements described in Section 7.2.1. In addition, we split the data on a daily level, and removed days with missing values in the household aggregate consumption, car load or solar power generation. Only days with positive solar power production, as well as car daily load between 4 kW and 96 kW were kept to align the dataset with the physical constraints defined in Section 7.2.2.1. Households with less than 50 available days were excluded from the dataset. Finally, for house 4767, only days until June were considered eligible, since the end-user purchased a fast charger in July that hampers data consistency [159]. Table 7.2 presents the final number of eligible days per household.

Table 7.2 Number of days that fulfill the selected criteria per house.

| House | Total number of days | Eligible days | Training Set | Test set |
|---|---|---|---|---|
| 661 | 365 | 242 | 189 | 53 |
| 1642 | 362 | 263 | 206 | 57 |
| 4373 | 361 | 307 | 242 | 65 |
| 4767 | 365 | 85 | 65 | 20 |
| 6139 | 365 | 170 | 134 | 36 |
| 8156 | 365 | 300 | 238 | 62 |

Table 7.3 2018 time-of-use electricity tariffs of Austin, Texas households.

| ToU Period | Hours | Electricity tariff ($/kWh) |
|---|---|---|
| Off-Peak | 00:00 - 06:00 | 0.01188 |
| Mid-Peak | 22:00 - 24:00<br>06:00 - 14:00<br>20:00 - 22:00 | 0.06218 |
| On-Peak | 14:00 - 20:00 | 0.11003 |

To calculate the electricity cost, real-world ToU electricity rates [138] from a provider in the area of Austin, Texas were utilized, as shown in Table 7.3. Prices are divided into low, medium and high cost time windows. Similar to previous works, we assume a net-metering arrangement and set the electricity selling price and buying price to be equal [130, 160, 161]. The EV end-user is assumed to possess a Nissan Leaf with a rated battery capacity of 24 kWh and a Level 2 (AC) slow charger with a charging efficiency of 90.5% [139]. We compare the proposed N-Step DQN approach with historical data, as well as previous work [135]. All experiments were conducted on a workstation with AMD Threadripper Pro 5975x CPU and NVIDIA RTX 4080 GPU.

### 7.3.2 Evaluation Metrics

We use three metrics to evaluate the EV charging profiles. First, to quantify self-consumption for EV charging, we introduce the Self-Consumption Index (*SCI*), which measures the amount of solar power used for charging the EV divided by the total EV charging load, as shown in Equation 7.18. It is assumed that the energy produced by the PV is first directed to the EV and then to the residual household load. SCI ranges between 0 (all energy required to charge the EV comes from the grid) and 1 (all energy required to charge the EV comes from the household's PV panel), with a higher value being desired to ensure solar power self-consumption.

$$SCI = \frac{\sum_{i=1}^{T} min(P_{ev,i}, P_{pv,i})}{\sum_{i=1}^{T} P_{ev,i}} \tag{7.18}$$

In addition, we calculate the total electricity cost ($) (*TEC*) by multiplying the electricity price $c_i$ for each timestep with the aggregated household grid connection $P_{H,i}$.

$$TEC = \sum_{i=1}^{T} c_i \frac{P_{H,i}}{4} = \sum_{i=1}^{T} c_i \frac{P_{res,i} + P_{ev,i} \searrow P_{pv,i}}{4} \tag{7.19}$$

Finally, to measure network stress, we calculate the Peak-to-Average Ratio (*PAR*) of the total grid connection load.

$$PAR = \frac{\max_T |P_H|}{\frac{1}{T}\sum_T |P_H|} = \frac{\hat{P_H}}{\bar{P_H}} \tag{7.20}$$

where $\hat{P}_H$ signifies the maximum absolute power that is either drawn or given back to the grid, and $\bar{P}_H$ represents the average household grid consumption.

### 7.3.3 Experimental Results

During data analysis, it has been observed that the eligible days showcase different characteristics in terms on EV charging load, solar power generation and residual load. To better evaluate model performance, we differentiate days in good and bad days for EV demand response, by assessing whether the EV load is greater than 20kW and, concurrently, solar power generation exceeds the observed EV load. The experimental results are shown in Table 7.4.

It can been noticed that the proposed N-DQN method showcases higher SCI in 5 out of 6 houses on good DR days and 4 out of 6 houses on bad DR days. Averaging across all houses, our approach demonstrates an 19.66% (good days) and 9.52% (bad days) increase in solar energy utilized for EV charging compared to the historical data and an 4.2% (good days) and 8.82% (bad days) SCI increase in comparison to previous work [135]. At the same time, even though our method attains less savings and PAR reduction than [135], the values are similar and there are still improvements compared to the historical data (10.9 % and 7% respectively). This finding is not surprising, since in most of the cases a higher SCI entails the end-user to shift load from lower cost periods (night time) to midday periods where ToU tariffs are higher. Under the implemented pricing scheme, it is beneficial for the end-user to sell the PV-generated power back to the grid in mid-day time periods and charge the EV during night hours, since the cost of electricity during the day is higher than at night (see Table 7.3). However, the scope of this work is to promote solar panel self-consumption, as the total electricity cost is subject to the applied tariffs scheme or the underlying policy with regards to the way domestic excess solar energy is sold back to the grid. Both can change significantly depending on the provider policies.

Table 7.4 Experimental results for each household.

| House ID | Model | *Good days for DR* SCI [%] | TEC [$] | PAR | *Bad days for DR* SCI [%] | TEC [$] | PAR |
|---|---|---|---|---|---|---|---|
| 661 | Historical | 9.36 | 39.69 | 2.94 | 3.79 | 8.57 | 3.36 |
| | DQN [135] | 27.70 | 38.70 | 2.82 | 18.27 | 8.17 | 3.20 |
| | N-DQN | 19.19 | 36.35 | 2.95 | 9.41 | 7.86 | 3.19 |
| 1642 | Historical | 38.04 | 29.08 | 3.62 | 21.84 | 24.56 | 4.63 |
| | DQN [135] | 44.55 | 26.17 | 3.06 | 11.36 | 18.51 | 3.91 |
| | N-DQN | 44.19 | 27.28 | 3.15 | 27.73 | 20.59 | 4.09 |
| 4373 | Historical | 33.40 | 58.96 | 3.18 | 35.86 | 22.45 | 4.52 |
| | DQN [135] | 59.77 | 46.85 | 3.00 | 45.72 | 18.94 | 4.00 |
| | N-DQN | 67.78 | 47.68 | 3.20 | 52.37 | 19.88 | 4.26 |
| 4767 | Historical | 24.21 | 24.35 | 4.18 | 13.01 | 11.21 | 4.66 |
| | DQN [135] | 67.38 | 21.00 | 3.50 | 36.26 | 8.63 | 3.71 |
| | N-DQN | 70.67 | 22.73 | 3.55 | 40.17 | 11.66 | 3.75 |
| 6139 | Historical | 7.65 | 41.00 | 3.79 | 4.46 | 31.61 | 3.93 |
| | DQN [135] | 6.75 | 39.76 | 3.25 | 8.02 | 27.39 | 3.32 |
| | N-DQN | 8.32 | 40.24 | 3.18 | 7.93 | 27.22 | 3.32 |
| 8156 | Historical | 27.61 | 33.73 | 3.33 | 13.83 | 57.05 | 4.19 |
| | DQN [135] | 25.51 | 26.40 | 3.08 | 8.14 | 49.61 | 3.69 |
| | N-DQN | 47.02 | 28.99 | 3.08 | 24.88 | 50.78 | 3.75 |
| Average | Historical | 24.30 | 226.82 | 3.39 | 16.52 | 155.44 | 4.21 |
| | DQN [135] | 39.76 | **198.89** | **3.06** | 17.23 | **131.26** | **3.67** |
| | N-DQN | **43.96** | 203.28 | 3.15 | **26.05** | 137.98 | 3.78 |

In addition, it is important to examine the charging behaviour of the different models under various conditions. To that end, we have identified three meaningful use cases from different households to show the model's generalization potential. First, we examine the charging behaviours on a good DR day with high load complexity, where high air-conditioning consumption activity is in place. Figure 7.4 illustrates the differences among different charging policies as well as the way the proposed EV charging load affects the household's total consumption profile.

Household 4373 on 21/06/2018 exhibits a high EV load that needs to be distributed during the day and, at the same time, the residual load has significant fluctuations due to air-conditioning. Evaluating the proposed charging policy, it is evident that the model proposes time windows that are better for the end-user, since a large amount of the energy used for EV charging stems from PV generation (31.7kW out of 48.6kW). At the same time, our N-Step DQN approach showcases a more effective behaviour compared to the DQN model, since in two cases it avoids a spike in the residual load that the DQN does not, resulting in better savings (18% versus 16%). In the lower plot of Figure 7.4, we see that shifting the load inside the solar power generation hours significantly reduces network stress during high peak hours (18:00-22:00), resulting in a PAR improvement of 20%.

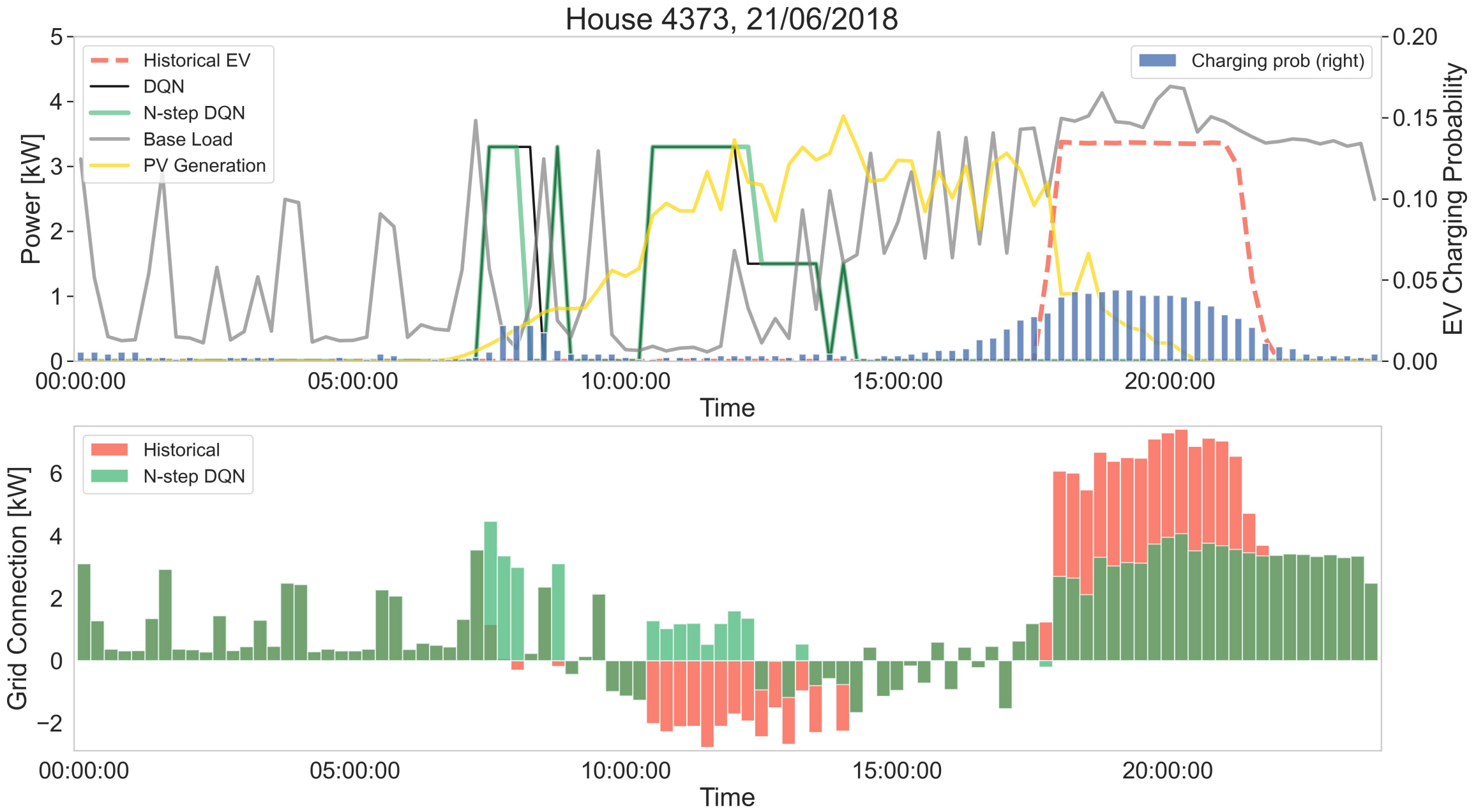

Figure 7.4 Overview of the proposed self-sustained charging policy (upper) and the total household consumption profile (lower) on a good DR day with high load complexity.

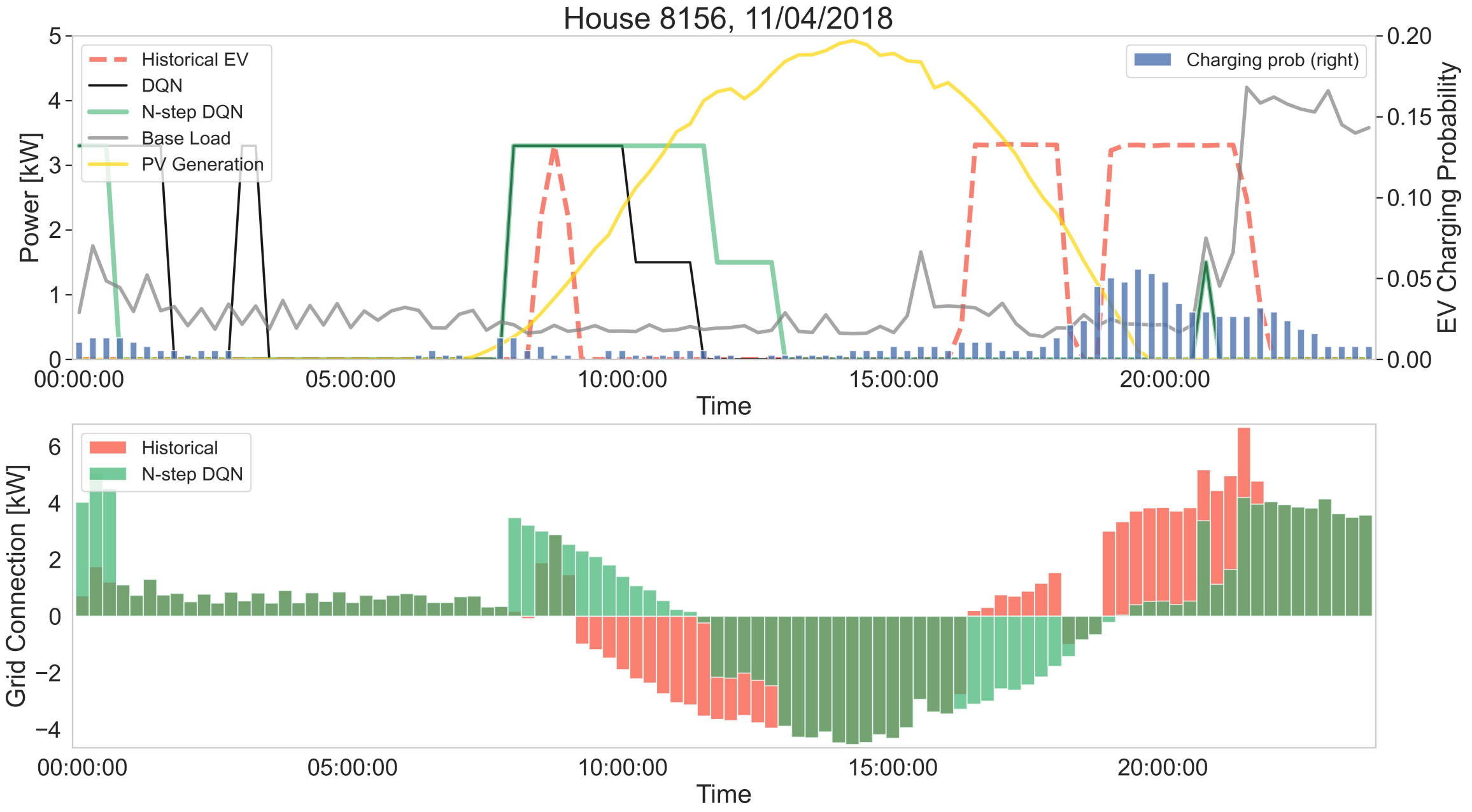

Figure 7.5 Overview of the proposed self-sustained charging policy (upper) and the total household consumption profile (lower) on a good DR day with low load complexity.

Then, we would like to assess the charging behavior for a different household on a good day for DR with significantly lower load complexity, as illustrated in Figure 7.5. Similar to the high complexity case (Figure 7.4), the proposed N-Step DQN method achieves a very high self consumption index (37.2 out of 68.4 kW are drawn from the solar panel), and manages to charge the EV at a time window with low residual load. In contrast, DQN decides to charge the EV more during the night, which leads to higher cost savings (-1.4 $ versus -1.2$) and PAR reduction (28% versus 16%), but violates the user preferences in certain time slots, since it proposes to charge in low frequency periods.

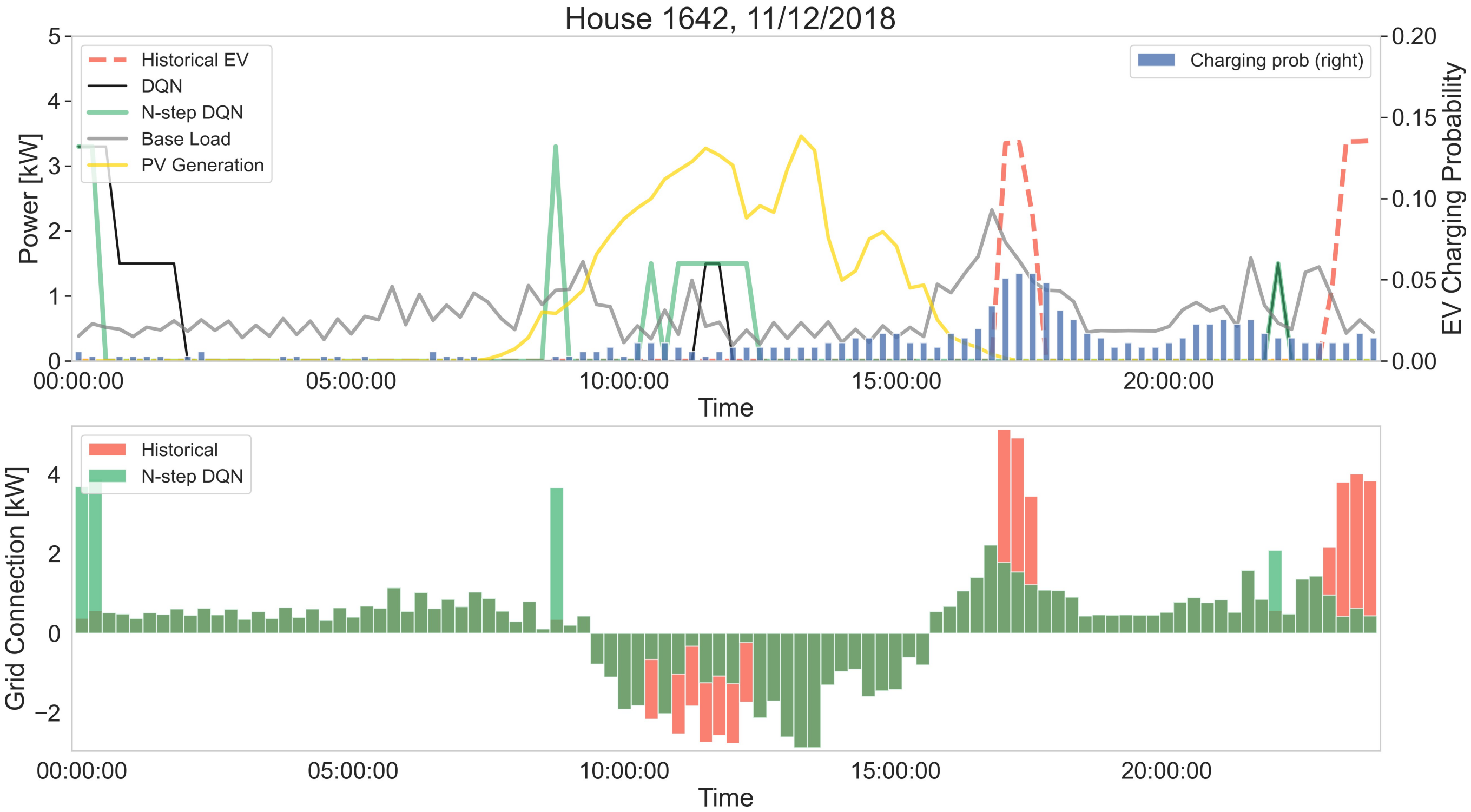

Figure 7.6 Overview of the proposed self-sustained charging policy (upper) and the total household consumption profile (lower) on a bad DR day.

Finally, we evaluate our approach on a bad day for DR, where the EV load is significantly lower ($< 20$ kW), as shown in Figure 7.6. In this case, our N-Step DQN approach behaves better than the DQN, which prefers to charge the EV during night hours, achieving an almost three-fold increase of solar PV self consumption, dedicated to EV charging (11.2 out of 21.9 kW versus 3 out of 21.9 kW). Charging at night results in higher cost savings and PAR reduction, but solar power generation is not utilized.

## 7.4 Conclusions

In this chapter, a self-sustained EV charging optimization framework with N-Step Deep Reinforcement Learning has been proposed. Our approach relies on smart-metered data to decide the optimal timeslots of the day to charge the EV. Real-world smart metered data from the Pecan Street dataset [136] have been utilized. In-depth analysis has shown that end-user charging tendencies may significantly vary between weekdays, a finding that was taken into consideration during the optimization design. The optimization process considered multiple factors that influence end-users' charging habits, such as historical charging probability, electricity cost and solar power generation. Our approach was evaluated on data from six households in Austin, Texas, and resulted in a charging policy that increases solar power self consumption by an average of 19.66% accross all houses compared to historical data, outperforming previous work in the field. At the same time, the proposed charging scheme showcases an average 10.9% electricity cost reduction and limits the household Peak-to-Average ratio by an average of 7%, thus reducing network stress and paving the way towards passenger vehicle transportation decarbonization.

# Chapter 8

# Community-driven Smart EV charging With Multi-Agent Deep Reinforcement Learning

## 8.1 Introduction

Recently, there has been observed an increased mobilization in transforming the global energy consumption profile towards achieving net-zero emissions by 2050 [119]. One domain which has received increasing attention is transportation, which is the human activity sector with the highest yearly $CO_2$ emissions growth rate [10]. Even though ongoing policies are increasing support towards electrification of the transport sector, the corresponding $CO_2$ emission reduction relies on the assumption that power generation will transition to fossil-free energy resources, resulting in transport vehicles utilizing renewable energy. However, the proportion of renewable power in the energy generation mix still lacks compared to fossil fuels, which threatens the achievement of the transport sector decarbonization goals [11].

At the same time, higher electricity needs due to the increasing penetration of electric vehicles (EV) are expected to put further stress on existing low voltage network infrastructure by escalating grid demand in high load periods [140]. A surrogate direction towards EV fleet decarbonization would be the investment into decentralized energy resources, such as residential solar photovoltaic (PV) production. Residential PV can provide an alternative, locally produced energy source which can be used to power EV with clean, renewable energy, thus avoiding energy distribution costs and network stress. Therefore, self-consumption of locally produced renewable energy through smart EV charging optimization can be seen as a key facilitator towards a quicker transition to a carbon-neutral domestic vehicles fleet. Multiple works have explored the optimization of EV charging for domestic households [125, 129–131, 134, 155, 162]. However, that solution to be widely adopted, each EV owner will have to invest in I) a solar PV panel, and II) a Home Energy Management System (HEMS), which presents a significant cost barrier that hinders the broader implementation of such approach.

Owning both a solar PV panel and a HEMS may be not affordable for everyone within a community due to advanced costs, building infrastructure and potential neighborhood obligations [163, 164]. Following a different direction, these costs could be attenuated if the required investments affected multiple households instead of one. Specifically, a community of houses can jointly invest in a solar PV and central energy management system. In this way, the Community Energy Management System (CEMS) will be responsible for optimizing the EV charging profiles for each household, while treating renewable energy generation as a communal, shared resource.

Inspired by the above challenges, in this Chapter we present a community-driven smart EV charging optimization scheme with Multi-Agent Deep Reinforcement Learning (DRL)[165]. EV charging profiles are optimized by shifting power loads to alternative cost-friendly time slots throughout the day, utilizing the communal PV and adapting to the individual EV charging requirements. This is achieved with the use of DRL agents, tailored to the household needs, and parallel operation. The advantages of the proposed approach are the following:

- We introduce a community-driven smart EV charging with Multi-Agent DRL operating and interacting in the same environment while optimizing the individual household needs.
- Our proposed community-driven EV charging optimization scheme is able to optimize multiple EV charging profiles concurrently, allowing for better utilization of community shared resources.
- The community-based approach reduces the cost barrier for residential EV owners that desire smart EV charging, since each household is no longer required to install a HEMS individually.
- The proposed multi-agent optimization of EV charging leads to reduced average costs, higher utilization of the PV generated power and lower Peak-to-Average Ratio on the combined community grid connection.

## Related Work

Numerous EV charging scheduling approaches have been proposed in the recent past, focusing on the areas of both charging stations and residential settings [145, 146]. Many initial works on EV charging optimization approached the problem by employing mathematical model-based optimization methods. In [147–149], the EV charging scheduling was designed as a cost minimization objective, which was solved with linear programming. Other approaches handled the EV charging problem through dynamic programming [166, 167], or a genetic algorithm strategy [168]. However, these model-based methods demand considerable engineering effort to address the smart grid's complexities, as well as hindsight over the full optimization horizon, which complicate their practical deployment in real-world applications [169].

To overcome these limitations, recent approaches have adopted data-driven and machine learning methods that show powerful data processing and interpretation capabilities. Artificial Intelligence methodologies have become increasingly pivotal in a wide range of smart grid solutions [120, 121, 170, 153, 64]. Following this tendency, EV charging scheduling research has transitioned from mathematical model-based methods to those centered on Deep Learning, with a particular focus on Deep Reinforcement Learning (DRL) [125, 129–131, 134, 155, 171]. However, these approaches suffer from several limitations. First, these works don't always tackle residential households, but focus on EV charging stations [171]. Additionally, most works depend on simulated data instead of real-world measurements, which can lead to inaccurate representation of household consumption patterns, given that human activity can often cause outlier values not captured in the simulation [129–131, 155]. Finally, the optimization process focuses only on a single household, which limits its applicability in broader optimization schemes, for instance a community.

Multi-Agent Deep Reinforcement Learning (MADRL) refers to a sub-domain of DRL where multiple agents simultaneously learn and adapt by interacting with an environment [172]. This method is particularly useful in complex systems where multiple entities operate either independently or jointly in an optimization problem. Consequently, MADRL has also been applied to the EV charging optimization problem [169, 173–177]. However, similar to the single-agent approaches, most works focus on optimizing EV charging in charging stations instead of residential premises [169, 173–175]. In addition, the works centered on domestic households either operate in an agnostic way to the households' consumption pattern, focusing only on the EV state-of-charge [177], or utilize the EV mainly as an energy storage entity to lower electricity cost and aggregated household consumption, which may lead to early EV battery degradation [176].

To summarize, even though EV charging has been studied in the literature, most works focus on charging station or individual household optimization. In addition, many works utilize simulated data in their approach, neglecting the impact of real-world measurements on the optimization outcome. Therefore, we can conclude that the concept of parallel optimization of EV charging on a community level remains underexplored, highlighting a significant area for further research and development.

## 8.2 Methodology

In the following sections, we proceed with the formulation of the community-driven EV-charging optimization concept. We begin with the concept definition and proceed to the formulation of mathematical equations that enable concurrent optimization of multiple EV using shared resources.

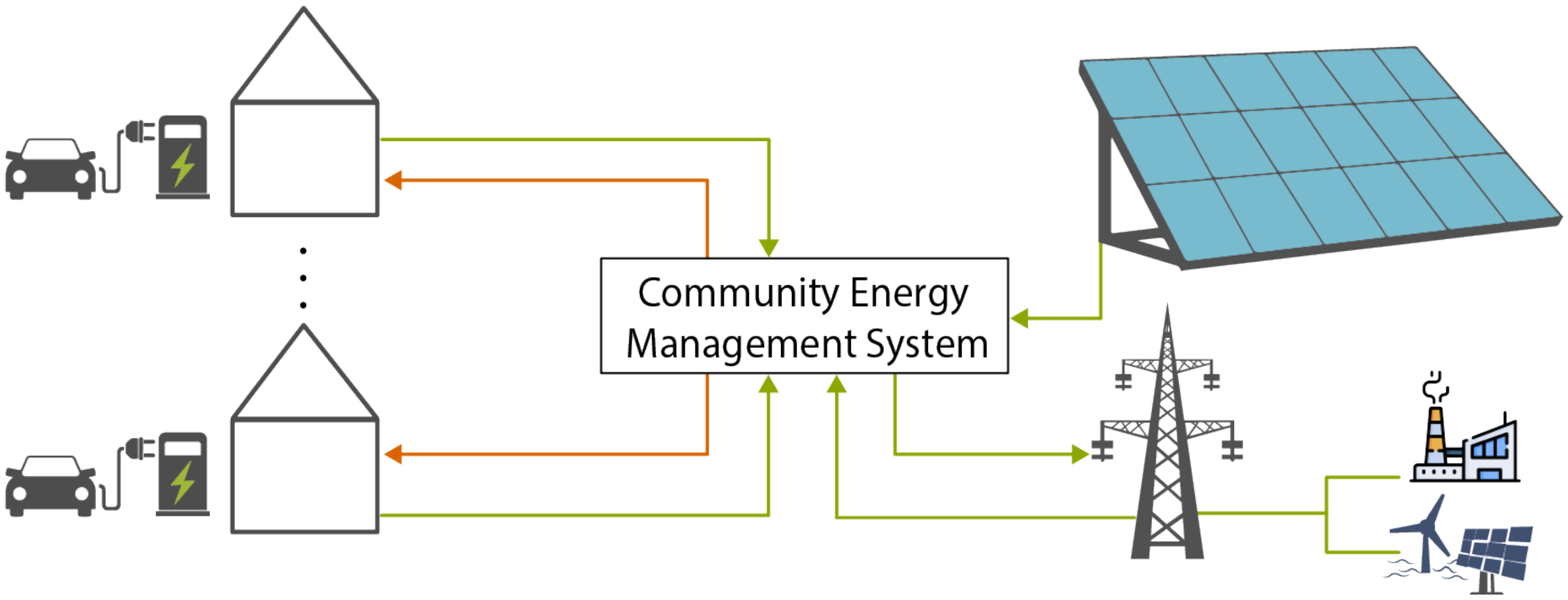

Figure 8.1 Overview of the proposed community-driven EV charging optimization scheme. A community of smart houses have invested in a shared solar PV panel as well as in a community energy management system, responsible for optimizing EV charging for all vehicles of the community according to the proposed DRL approach.

### 8.2.1 Community-driven EV charging Concept

The proposed community-driven EV charging optimization scheme is presented in Figure 8.1. A community of *N* households, each equipped each with an EV and EV charger, have jointly invested in a solar PV power generation panel and a community energy management system (CEMS). The CEMS purpose is to analyze power consumption information and shift power loads to alternative time periods within the day, thereby reducing the cost-barrier for smart EV charging in contrast to traditional individual household power optimization methods. Moreover, the CEMS is responsible for optimizing the charging behavior of each individual EV, utilizing energy from the communal solar PV panel, while respecting the energy consumption and EV charging tendencies of each individual household. In the proposed community-driven EV charging optimization scheme, the EV optimization is realized concurrently, in an intelligent manner using DRL.

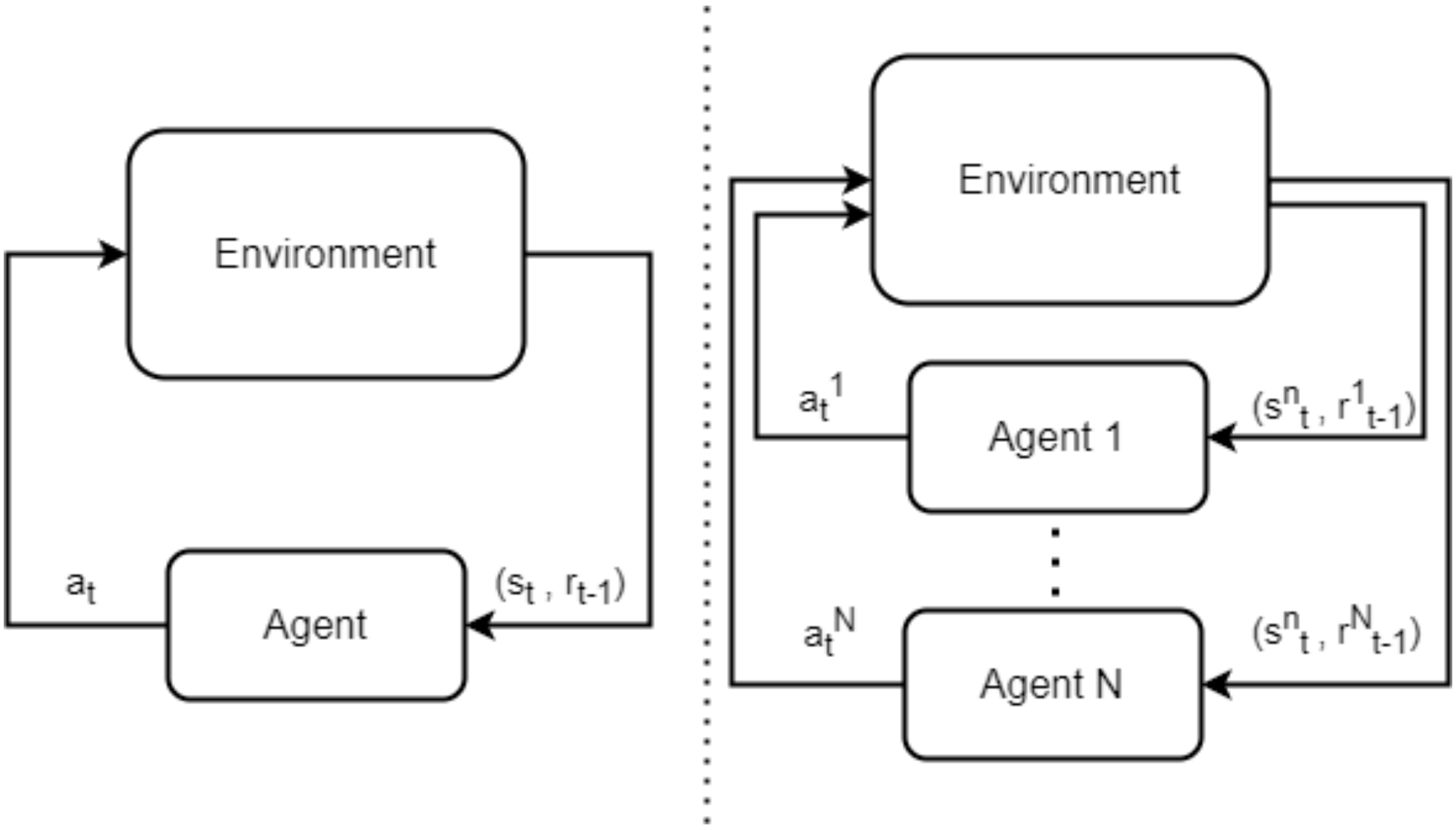

Figure 8.2 Left: Conventional single-agent DRL training. Right: Proposed multi-agent DRL training for community-driven EV charging

### 8.2.2 Community-driven Multi-Agent EV charging DRL-environment Formulation

To proceed with mathematically formulating the Community-driven Multi-Agent EV charging optimization scheme, the single-agent RL framework needs to be formulated. Let the single-agent RL optimization be defined as a Markov Decision Process described by the sets $(S , A , R)$, with random variables $(S_t, A_t, R_t)$ signifying the environment states, actions and rewards respectively. The RL agent considers a policy $u(s) = P_r[S_{t+1} = s_{t+1} / S_t = s_t, A_t = a_t]$, with $a_t \rightarrow A$ , $s_t \rightarrow S$ , to select optimal actions maximizing the value function $Q_u\ (s_t, a_t) = R_t + \#\ \max_{a \rightarrow A} Q(S_{t+1}, A_t)$, where # is the discount factor. The single-agent's behaviour is captured on the left side of Figure 8.2. Now, to advance the operation principle of RL to a deep one, a neural network is used as a function to learn the optimal value function $Q_{\mathsf{u}}(s_t, a_t)$.

In our proposed Community-driven Multi-Agent EV charging optimization scheme, we extend the operation of the single-agent DRL to a multi-agent one, as shown on the right part of Figure 8.2. Considering $n = 1, \ldots, N$ households, the policy for each household is individualized, resulting in their parallel update, defined as a vector $\mathsf{u}(s_t) \rightarrow \mathsf{R}^N$ and given as

$$\mathsf{u}(s) = \begin{bmatrix} \mathsf{u}_1(s_t^1) \\ \vdots \\ \mathsf{u}_N(s_t^N) \end{bmatrix} = \begin{bmatrix} P_r[S_{t+1}^1 = s_{t+1}^1 \mid S_t^1 = s_t^1, A_t^1 = a_t^1] \\ \vdots \\ P_r[S_{t+1}^N = s_{t+1}^N \mid S_t^N = s_t^N, A_t^N = a_t^N] \end{bmatrix}. \tag{8.1}$$

Consequently, using (8.1), a vector $Q_{\mathsf{u}}(s_t, a_t)$ containing all optimal value functions is given as

$$Q_{\mathsf{u}}(s_t, a_t) = \begin{bmatrix} Q_{\mathsf{u}}^1(s_t^1, a_t^1) = R_t^1 + \#^1 \max_{a \rightarrow A} Q(S_{t+1}^1, A_t^1) \\ \vdots \\ Q_{\mathsf{u}}^N(s_t^N, a_t^N) = R_t^N + \#^N \max_{a \rightarrow A} Q(S_{t+1}^N, A_t^N) \end{bmatrix} \tag{8.2}$$

Each agent then selects the action that maximizes (8.2). We implement the $N$ agents as Deep Q-Learning (DQN) agents [137], since it is an algorithm that has gained increasing attraction over the recent years, also able to maintain high optimization performance in scenarios with a large state-action space [157]. The parametrization of the agents, as well as the RL environment are described in the following subsection.

### 8.2.3 Multi-Agent Environment Description

Here we proceed with the description of the multi-agent EV charging optimization environment. The timestep used has a 15-minute temporal resolution, where each Agent's objective is the optimal EV charging load distribution over a 24-hour, $T = 96$, modelling horizon, while utilizing the community-shared solar PV panel. Each 24-hour period is treated as a separate entity to reduce the optimization complexity. We follow an approach similar to [144] and modify it to adapt to the multi-agent optimization case. Moreover, we consider the state of charge (SoC) and assume that the total EV charging load observed over a 24-hour period, $\hat{P}_{ev}$, represents a complete charging cycle. The initial value of SoC for each vehicle is

$$\mathrm{SoC}_{t=0}^n = 1 \searrow \eta \frac{\hat{P}_{ev}}{4C_{ev}} \tag{8.3}$$

where $\eta$ denotes the charging efficiency of the EV battery and $C_{ev}$ refers to the maximum battery capacity, measured in kWh [178]. Furthermore, considering the 2-level slow EV charger [179], each EV is charged according to the SoC value as

$$P^n_{ev,t} = \begin{cases} 3.3\text{kW}, & \text{SoC}^n_{t=0} \leq \text{SoC}^n_t \leq 0.9 \\ 1.5\text{kW}, & \text{SoC}^n_t > 0.9 \end{cases} \tag{8.4}$$

Therefore, considering (8.4) the SoC for all EV is updated as

$$\text{SoC}^n_{t+1} = \text{SoC}^n_t + \eta \frac{P^n_{ev,t}}{4C_{batt}} \tag{8.5}$$

Through (8.3), (8.4) and (8.5), one can reasonably argue that the state change in the DRL depends on a plethora of input parameters, which are in fact, the inputs to the DQN.

Formally, in each of the $N$ DQNs in total, the transition from state $S^n_t$ to $S^n_{t+1}$ is determined by a vector of parameters $s_t \in \mathbb{R}^N$ given as

$$s^n_t = \left[ P^n_{h,t}, P_{pv,t}, \text{SoC}^n_t, \tilde{P}^n_{ev,t}, c_t, \sin\left(\frac{t}{T}\right), \cos\left(\frac{t}{T}\right) \right] \tag{8.6}$$

where each of the parameters is

- $P^n_{h,t}$ : Residual (non-EV) power consumption of house $n$
- $P_{pv,t}$ : Shared solar PV power generation of the community
- $SoC^n_t$ : EV state of charge for vehicle $n$
- $\tilde{P}^n_{ev,t}$ : Cumulative daily power load utilized for EV charging until timestep $t$
- $c_t$ : electricity tariff price at timestep $t$

Similarly, considering the above and the definitions in Section 8.2.2, the resulting action space is such that

$$\mathbf{a}_t = [a^1_t, \ldots, a^N_t]^T \in \{0, 1\}^N. \tag{8.7}$$

Next, we define the reward function, which is a major cornerstone of any RL algorithm, since the agents aim to maximize the rewards returned for their respective actions over time. We follow the concept of reward shaping [180] in which we divide the reward function into sub-modules to allow for more precise modelling behavior and evoke the desired optimization behavior. The overall reward value is calculated as

$$r^n_t = \ast_1 r^n_{d,t} + \ast_2 r^n_{tnd,t} + \ast_3 r^n_{c,t} + \ast_4 r^n_{\text{soc},t} + \ast_5 r^n_{pv,t}. \tag{8.8}$$

In (8.8), $r^n_{d,t}$ is utilized to avoid scenarios where the optimized load does not match the historically observed quantity. The agent is positively rewarded if the EV is charged in time steps where the load goal has not yet been achieved, and is heavily penalized if charging would lead to exceeding the maximum SoC, described as

$$r^n_{d,t} = \begin{cases} 1, & a^n_t = 1 \;\&\; \frac{\tilde{P}^n_{ev,t}}{\hat{P}^n_{ev}} \leq 1.05 \\ -10, & a^n_t = 1 \;\&\; \frac{\tilde{P}^n_{ev,t}}{\hat{P}^n_{ev}} > 1.05 \\ -0.25, & a^n_t = 0 \;\&\; \frac{\tilde{P}^n_{ev,t}}{\hat{P}^n_{ev}} \leq 1.05 \\ 0, & a^n_t = 0 \;\&\; \frac{\tilde{P}^n_{ev,t}}{\hat{P}^n_{ev}} > 1.05 \end{cases} \tag{8.9}$$

Since we use discrete charging values and a 100% match of the historically observed load is computationally infeasible, we ease the constraint and allow for up to a 5% overcoming of the historical EV load.

Next, $r^n_{tnd,t}$ aims to punish the RL agents if they decide to charge in timeslots where EV charging in the historical data is infrequent. To model this behaviour, we utilize an EV charging probability distribution function $F^n_{ev,t}$ for each EV, which is derived from the charging data in the dataset. The agent is then penalized for charging in timeslots where the value of $F^n_{ev,t}$ is lower than the 25-th percentile of the distribution.

$$r^n_{tnd,t} = \begin{cases} -2, & a^n_t = 1 \;\&\; F^n_{ev,t} < Q_{F^n_{ev},25} \\ 0, & \text{otherwise} \end{cases} \tag{8.10}$$

To quantify the impact of electricity cost in the optimization, we follow a similar approach as above. Specifically, the cost is calculated by multiplying the time-of-use tariff price $c_t$ from Table 8.1 with the aggregated household grid connection, as shown in (8.11). Since the grid connection is expressed in kW and the price in \$/kWh, a division by a factor of 4 is required. We then compare the actual electricity cost of each house with the with the 25, 50 and 75-th percentiles of the cost distribution and reward the agents accordingly, as shown in (8.12).

$$C^n_t = \frac{c_t}{4}\left(a^n_t P^n_{ev,t} + P^n_{h,t} - P^n_{pv,t}\right) \tag{8.11}$$

$$r^n_{c,t} = \begin{cases} 2, & a^n_t = 1 \;\&\; C_t \leq Q_{F_c,25} \\ 1, & a^n_t = 1 \;\&\; C_t \leq Q_{F_c,50} \\ 0, & a^n_t = 0 \\ -1, & a^n_t = 1 \;\&\; C_t \leq Q_{F_c,75} \\ -2, & a^n_t = 1 \;\&\; C_t > Q_{F_c,75} \end{cases} \tag{8.12}$$

The final two rewards relate to the utilization of solar energy for EV charging, and the battery overflow. The agent is rewarded positively if a charging action is taken in timesteps with positive solar power generation, and is heavily penalized if the charging action would surpass the capacity of the EV maximum battery capacity. This is described with the following

$$r^n_{pv,t} = \begin{cases} 2, & a^n_t = 1 \;\&\; P_{pv,t} > 0 \\ 0, & \text{otherwise} \end{cases} \tag{8.13}$$

and

$$r^n_{soc,t} = \begin{cases} -10, & a^n_t = 1 \;\&\; SoC^n_t \geq 1 \\ 0, & \text{otherwise} \end{cases} \tag{8.14}$$

Table 8.1 Electricity prices under a time-of-use tarif in Austin,Texas (2018).

| ToU Period | Hours | Electricity tariff ($/kWh) |
|---|---|---|
| Off-Peak | 00:00 - 06:00<br>22:00 - 24:00 | 0.01188 |
| Mid-Preak | 06:00 - 14:00 | 0.06218 |
| On-Peak | 20:00 - 22:00<br>14:00 - 20:00 | 0.01188 |

# 8.3 Experimental Setup and Results

## 8.3.1 Experimental Setup

To confirm the efficacy of our approach, we utilized the Pecan Street dataset [136], which contains power consumption measurements from residential households in Austin, Texas. We filtered the data to keep only households with EV charging data and solar power generation, resulting in a dataset consisting of 6 households. In addition, the data were split on a daily level, and only days that met the requirements described in [144] were considered. The pricing scheme is a real-world Time of Use (ToU) tariff from an electricity provider in the area, which divides the day into low, medium and high cost intervals. To align with previous work in the area, we assume a net-metering arrangement for electricity buying and selling price, in the case that solar power generation exceeds consumption demand [130, 160, 161]. Each household is assumed to possess only 1 EV, and the EV model used in the evaluation is a Nissan Leaf, which has a maximum battery capacity $C_{ev}$ of 24 kWh, charges from a Level 2 charger and showcases 90.5% charging efficiency [139]. To model the communal solar PV panel, we aggregate the solar power production of the individual household panels. All experiments were conducted with an AMD Threadripper Pro 5975x CPU and NVIDIA RTX 4080 GPU, in Python 3.9.

## 8.3.2 Evaluation Metrics

Since our approach relies on multi-objective optimization, we utilize three metrics to evaluate model performance. First, we calculate the total electricity cost for the six households, which can be obtained by multiplying the electricity tarif price $c_t$ with the aggregated community grid (in kWh).

$$\mathrm{TEC} = \sum_{n=1}^{N} \sum_{t=1}^{T} c_t \frac{P^n_{h,t} + P^n_{ev,t} - P^n_{pv,t}}{4}. \tag{8.15}$$

In addition, we modify the Self-Consumption Index introduced in [144] to apply in the proposed communal optimization scheme, presented in

$$\mathrm{SCI_{comm}} = \frac{\sum_{t=1}^{T} \min\left( \sum_{n=1}^{N} P^n_{ev,t}, P_{pv,t} \right)}{\sum_{n=1}^{N} \sum_{t=1}^{T} P^n_{ev,t}}. \tag{8.16}$$

Considering that comparative methods conduct single-house optimization, a slight modification to the original formula is required. To make the index comparable, SCI for individual optimization is calculated by dividing the total EV energy stemming from solar panels from all households and the total EV charging energy, as

$$\mathrm{SCI_{ind}} = \frac{\sum_{n=1}^{N}\sum_{t=1}^{T} \min\left(P_{ev,t}^{n}, P_{pv,t}^{n}\right)}{\sum_{n=1}^{N}\sum_{t=1}^{T} P_{ev,t}^{n}}. \tag{8.17}$$

Finally, we compute the community Peak-to-Average Ratio (PAR) for the aggregated community power profile, as the ratio between the maximum (positive or negative) and the average community grid flow.

$$\mathrm{PAR} = \frac{\max_{T} \sum_{n=1}^{N} |P_{h,t}^{n}|}{\frac{1}{T}\sum_{t=1}^{T}\sum_{n=1}^{N} |P_{h,t}^{n}|}. \tag{8.18}$$

Table 8.2 Experimental results

| Day | Total Electricity Cost (TEC) [$] | | | Self-Consumption Index (SCI) [%] | | | Peak-to-Average Ratio (PAR) | | |
|---|---|---|---|---|---|---|---|---|---|
| | UC | IOC | COC | UC | IOC | COC | UC | IOC | COC |
| 14/01/2018 | 0.07 | -0.35 | 0.05 | 59.4 | 44.9 | 67.4 | 2.47 | 2.50 | 2.48 |
| 18/01/2018 | 7.08 | 6.81 | 6.96 | 6.60 | 31.8 | 35.6 | 3.16 | 2.52 | 3.01 |
| 26/01/2018 | 8.07 | 7.26 | 7.08 | 1.10 | 20.7 | 28.3 | 3.24 | 2.02 | 2.71 |
| 30/01/2018 | 0.60 | 0.37 | 0.20 | 1.40 | 49.6 | 47.3 | 2.30 | 2.68 | 2.58 |
| 01/02/2018 | 3.02 | 1.85 | 1.88 | 17.4 | 40.4 | 50.5 | 2.70 | 2.42 | 2.33 |
| 05/02/2018 | 4.94 | 4.21 | 3.65 | 5.2 | 23.7 | 28.6 | 2.89 | 2.73 | 2.35 |
| 06/02/2018 | 9.78 | 9.18 | 9.41 | 0.20 | 6.1 | 19.0 | 3.40 | 2.36 | 2.29 |
| 07/02/2018 | 5.71 | 4.57 | 4.29 | 5.5 | 30.7 | 28.6 | 3.05 | 2.13 | 2.35 |
| 13/02/2018 | 10.32 | 9.75 | 9.39 | 4.5 | 14.6 | 34.7 | 2.56 | 2.08 | 2.08 |
| 16/02/2018 | 6.57 | 6.19 | 6.23 | 3.7 | 23.0 | 41.8 | 3.27 | 2.49 | 2.57 |
| 20/02/2018 | 9.12 | 8.49 | 7.90 | 0.5 | 4.9 | 20.4 | 2.83 | 2.01 | 2.57 |
| 26/02/2018 | -0.40 | -1.21 | -1.49 | 12.6 | 56.0 | 48.5 | 3.17 | 3.45 | 3.42 |
| 28/02/2018 | 3.07 | 2.79 | 2.96 | 1.7 | 35.0 | 63.1 | 2.79 | 2.48 | 2.48 |
| 01/03/2018 | -0.38 | -2.07 | -2.39 | 16.7 | 50.2 | 42.5 | 2.54 | 2.84 | 2.59 |
| 03/04/2018 | 6.50 | 5.45 | 5.74 | 16.6 | 16.1 | 45.3 | 3.29 | 2.66 | 2.42 |
| 04/04/2018 | -4.13 | -6.10 | -5.97 | 37.3 | 47.4 | 48.7 | 2.75 | 2.73 | 2.69 |
| 13/04/2018 | 7.11 | 6.16 | 5.99 | 10.2 | 27.4 | 29.8 | 2.96 | 2.21 | 2.14 |
| 28/04/2018 | 4.03 | 2.80 | 2.62 | 56.9 | 53.1 | 45.5 | 2.68 | 2.51 | 2.20 |
| 02/05/2018 | 11.02 | 9.53 | 9.18 | 17.1 | 29.8 | 30.9 | 2.72 | 2.12 | 2.03 |
| 06/05/2018 | 4.27 | 3.91 | 3.69 | 62.0 | 66.9 | 50.8 | 2.75 | 2.14 | 1.81 |
| 15/05/2018 | 12.32 | 10.99 | 10.87 | 25.5 | 48.2 | 45.9 | 3.20 | 2.60 | 2.41 |
| 17/05/2018 | 14.45 | 13.67 | 13.53 | 40.2 | 63.2 | 50.6 | 2.45 | 2.22 | 2.00 |
| 12/06/2018 | 11.72 | 10.69 | 10.34 | 24.5 | 55.7 | 44.1 | 2.74 | 2.10 | 1.92 |
| 14/06/2018 | 9.25 | 10.05 | 10.33 | 9.1 | 50.5 | 53.1 | 2.46 | 2.37 | 2.16 |
| Average | 6.00 | 5.21 | **5.10** | 17.5 | 37.5 | **40.8** | 2.85 | 2.43 | **2.40** |

Table 8.3 Condensed Experimental results and relative improvement

| Approach | TEC [$/day] | SEC [%] | PAR |
|---|---|---|---|
| UC | 6.00 | 17.5 | 2.85 |
| IOC | 5.21 | 37.5 | 2.43 |
| COC | 5.10 | 40.8 | 2.40 |
| | Improvement(%) | | |
| COC VS UC | 17.65 | 133.14 | 18.75 |
| COC VS IOC | 2.16 | 8.80 | 1.25 |

### 8.3.3 Results

To validate the effectiveness of our approach, we compare the aggregated community consumption profiles i) in historical data, where no EV charging optimization occured and charging was uncontrolled and ii) after optimizing each household individually, as presented in [144]. We refer to the examined charging patterns as uncontrolled charging (UC), individually optimized charging (IOC) [144], and the proposed community-driven optimized charging (COC). The experimental results are shown in Tables 8.2 and 8.3.

From Tables 8.2 and 8.3 we observe that the community-driven multi-agent optimization approach (COC) showcases better performance, on average, across all metrics. In particular, our community driven optimization approach can reduce the total energy cost of the community by 17.65%, compared to uncontrolled EV charging, and 2.16% compared to individual household optimization. The latter number is further strengthened by the fact that only a single energy management system and solar PV panel investment is required, which significantly reduces the cost barrier towards smart energy management. In addition, we observe a notable increase in self-consumption of solar energy on a community level.

Compared to the net-metered data from the Pecan street dataset households, self-consumption of solar PV power generation rises a significant by 133.14%. It also ranks higher compared to individual household optimization, which already prioritized solar power utilization, outperforming it by 8.8%. Finally, our multi-agent optimization approach presents a 18.75% reduction of PAR compared to historical data, and 1.25% compared to individual household optimization, further reducing stress on the grid.

To further showcase the effectiveness of our optimization scheme we also present an example of the differences between historical data, individual household optimization and community-driven optimization in Figure 8.3. It can be observed that the proposed approach is able to better handle EV charging loads, focusing on charging during early morning hours, when solar power generation has started, and avoid more costly and demand-heavy periods of the day, potentially taxing the community's power network.

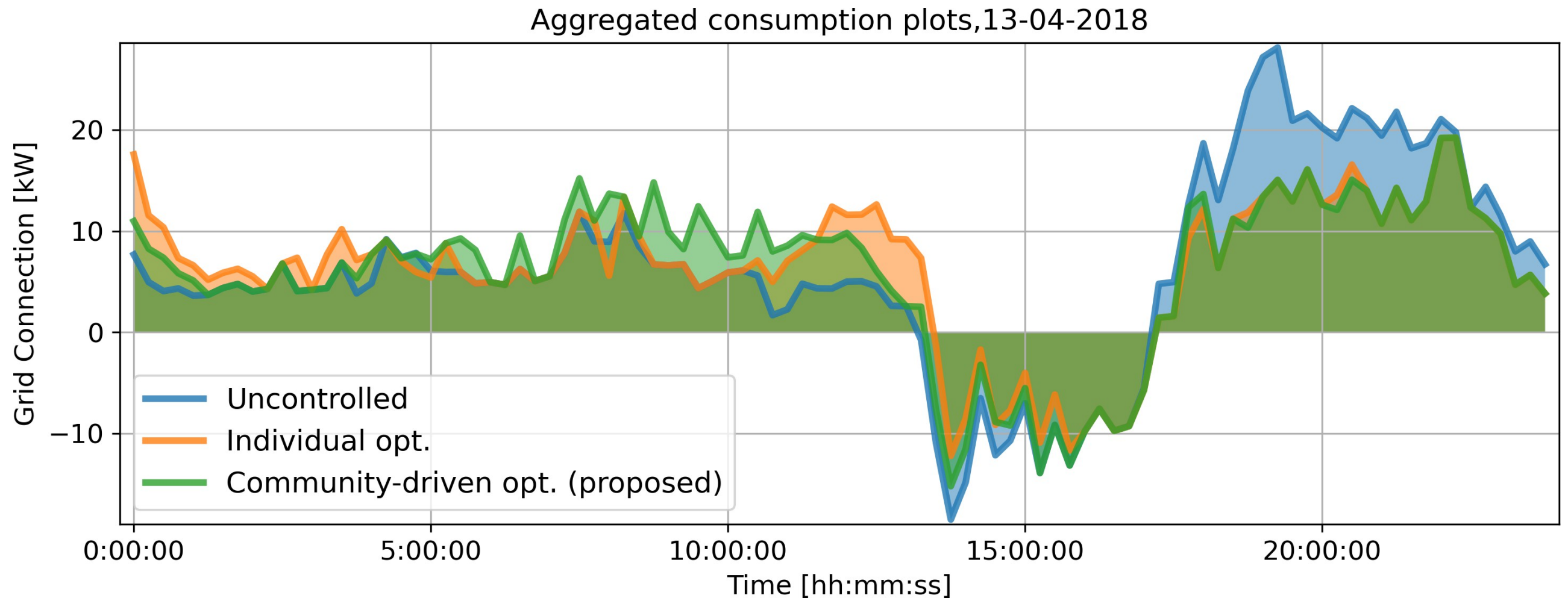

Figure 8.3 Example of an aggregated community power load. The total community consumption profile is presented before any optimization (blue), after being optimized individually (orange) and in a community-driven way (green) using the proposed Multi-Agent EV charging DRL optimization scheme.

## 8.4 Conclusions

In this Chapter, we proposed a community-driven EV charging optimization scheme with Multi-Agent Reinforcement Learning. In our approach, a community of residential households with EV have invested in a community owned solar PV panel and an energy management system, which is responsible for optimizing the EV charging behavior of the individual EV concurently. We implement the optimization scheme with Multi-Agent Reinforcement learning, where each household has its individual needs and the solar power energy is treated as a communal resource. Experimental results show that, compared to individual household optimization, our community-driven approach excels across all metrics, achieves a further cost reduction, better solar power utilization and reduction of the community Peak-to-Average ratio. These findings pave the way for further research into community-oriented solutions for energy efficiency in the ongoing energy transition.

# Chapter 9

# Conclusions

## 9.1 Summary

This dissertation investigated the development of novel Deep Learning techniques to create tools which solve limitations in two key energy domains; the reduction of residential energy consumption through end-user empowerment via Non-Intrusive Load Monitoring and the optimization of EV charging through Deep Reinforcement Learning. The dissertation was divided into three parts:

- Part I (Chapters 1, 2) presented the introduction and motivation of problem at hand. The motivation for the importance of the tackled topic was translated into research questions which would be answered in the subsequent chapters. Moreover, a mathematical formulation of the two examined energy efficiency domains was presented, serving as a foundational block for the better and deeper understanding of the conducted work.

- Part II encompassed the developed techniques contributing to the reduction of residential energy consumption via Non-Intrusive Load Monitoring. In particular, Chapter 3 presented a novel deep learning architecture for NILM. The model structure was based on Transformer layers, which enabled the utilization of attention mechanisms to learn long-range temporal dependencies in the data without heavily relying on data pre-processing and balancing techniques. Furthermore, Chapter 4 investigated the deployment of NILM models on edge devices, a process which maximizes end-user data privacy. Specifically, since the relation between model compression and model performance is rarely investigated, the presented work established a methodology to find the optimal equilibrium between the two. This was achieved by integrating evaluation of the model in the compression process in an iterative way, thus paving the way towards wider adoption of edge deployment solutions for NILM applications. Finally, Chapter 5 investigated the intricacies of post-deployment model monitoring and maintenance. It introduced a contextual performance tracking mechanism, as well as a model fine-tuning process which requires minimal additional samples to mitigate catastrophic forgetting.

- Part III deals with the optimization of EV charging with Deep Reinforcement Learning. In Chapter 6, a Deep Q-Learning Network is utilized to prioritize the utilization of locally produced solar energy from PV panels for EV charging, wihtout jeopardizing other optimization factors. In addition, Chapter 7 promotes self-consumption of distributed energy sources, such as PV panels, for EV charging. An N-step DQN is developed which achieves to maximize the utilization of solar power consumption, thus reducing stress put on the grid through energy backflows. Finally, Chapter 8 extends the aforementioned concept to a community-oriented optimization. More specifically, a (bigger) solar panel is assumed to be a communal resource, shared by multiple EV owners. Then , a multi-agent DQN approach was designed, which enables the concurrent optimization for multiple EV, thus maximizing the utilization of the shared renewable resource, and achieving both lower cost and less grid stress.

## 9.2 Innovation and Originality

The work presented in the previous Chapters was conducted with the ultimate goal of developing Deep Learning-based applications to advance applications which promote energy efficiency. Significant innovations were presented, with equal focus laid on both application domains (residential energy consumption and EV charging optimization). The main contributions of the dissertation are highlighted below.

1. **Development of a novel adaptable deep learning architecture for Non-Intrusive Load Monitoring, based on Transformer layers, which overcomes the limitations of existing sequential models** (see Chapter 3). The developed architecture can operate with minimal data pre-processing, and utilizes attention mechanisms to accurately estimate the power consumption of the individual appliance.

2. **Development of a performance-aware edge deployment framework for NILM models** (see Chapter 4). The proposed scheme alleviate the main limitation of existing NILM model compression approaches which utilize compression techniques arbitrarily without considering the trade-off between model performance and computational cost. The proposed edge optimization engine optimizes a NILM model for edge deployment depending on the edge devicefls limitations and includes a novel performance-aware algorithm to reduce the modelfls computational complexity, thus minimizing performance loss and limiting the required computational resources.

3. **Design and development of a performance-tracking and continual learning paradigm for NILM** (see Chapter 5). NILM models usually require frequent post-deployment maintenance to deal with non-stationary appliance data distributions, thereby elevating continual adaptability to a high priority of practical NILM applications. The methodology proposed in this Chapter introduces performance tracking according to environmental/seasonal alterations, which, combined with a contextual importance sampling mechanism, minimizes the amount of samples required for model fine-tuning. This novel retraining strategy alleviates the risk of catastrophic forgetting and can be regarded as an efficient post-deployment workflow for NILM models.

4. **Design and implementation of an EV charging optimization method, based on Deep Reinforcement Learning, which prioritizes the utilization of solar energy for EV charging** (see Chapter 6). The proposed approach utilizes real-world measurements to shape the end-user's flexibility potential, which, combined with Time-of-Use tariffs,creates a price-based Demand Response (DR) mechanism that can incentivize end-users to optimally shift EV charging load in hours of high solar PV generation. A DQN Agent is trained to learn a data-driven EV charging policy, leading to lower electricity cost and higher solar power utilization rate.

5. **Development of an EV charging optimization scheme with Deep Reinforcement Learning, which promotes self-consumption of distributed energy sources to minimize both the charging cost and the network stress by proposing alternative load scheduling** (see Chapter 7). The presented approach employs N-Step Deep Reinforcement Learning to charge the EV with clean energy from a PV, without neglecting other major factors that influence end usersfl behavior, such as electricity cost or EV charging tendencies. The importance of the proposed framework is highlighted by considering the fact that, even though the penetration of EV in the passenger vehicle fleet is increasing, the energy mix is not yet dominated by renewable energy sources, leading to the utilization of fossil-based power generation for EV charging. Optimizing EV charging with regard to the presented optimization parameters contributes significantly to the decarbonization of the transport sector.

6. **Design and development of a community-driven smart EV charging optimization scheme with Multi-Agent Deep Reinforcement Learning** (see Chapter 8). A concurrent EV charging optimization scheme for multiple vehicles is formulated, which considers a community-owned, shared resource. Then, through the use of Multi-Agent Deep Reinforcement Learning, multiple EV charging schedules can be optimized in parallel, which leads to better optimization performance and, at the same time, removes financial barriers for wider smart EV charging adoption.

## 9.3 Future Prospects

Even though this dissertation made significant contributions, both to Non-Intrusive Load Monitoring techniques and to EV charging optimization, it also sparked ideas for additional work in each of the topics covered. Hence, in this final Section, some ideas for future research are discussed.

In the case of NILM, most approaches in the literature focus on univariate time-series, utilizing active power as the sole model input to perform the disaggregation. The main motivation behind this choice is that measuring active power is easier. Devices, like power quality meters (PQ meters), which capture additional consumption information, such as reactive power or harmonics distortion, are significantly more expensive, which makes their adoption in residential households improbable. However, these variables capture characteristics of power signals which could greatly improve disaggregation performance. For example, appliances with capacitive loads (e.g. washing machine, air-conditions etc.) can be detected much easier when reactive power readings are present. An interesting application for NILM would be to utilize knowledge distillation with missing modality techniques [181] to improve model performance. Knowledge distillation refers to the process of distilling information from a teacher to a student model usually to a) reduce model size (if the student model is smaller than the teacher) or b) to improve performance. In the case of missing modality knowledge distillation, the teacher model is trained having access to a multivariate input signal, whereas the student receives only a subset of the input modalities. Through specific techniques, the student model can achieve significant performance increased compared to normal model training. In the case of NILM, the teacher model would have access to both active and reactive power, whereas the student model would receive only active power as input. It would be intersting to see if this approach can translate to the field of NILM and harvest reactive power knowledge, even if it is not present in the captured data.

In this dissertation, multiple EV charging optimization scenarios have been considered. Objectives such as electricity charging cost minimization, peak-to-average ratio minimization for grid congestion alleviation, as well as utilization of renewable energy sources have been successfully investigated, taking into consideration multiple constraints, such as user preferences, electricity tariff cost etc. In summary, EV charging optimization can be defined as the sequential decision making, within a predefined time window, on whether to charge or not an EV. Having this main principle in mind, the conducted work can be extended to account for additional optimization criteria, such as the real-time composition of the energy mix, to avoid time intervals where it is dominated by fossil-fueled electricity generation. In addition, the impact of charging on the EV battery, in terms of duration and amplitude, may be investigated, by utilizing a detailed physical battery model in the optimization process. The optimization could be constrained by the long-term impact on the EV battery to prolong its health.

# Publications

The following publications resulted through this research:

## Journals

- Stavros Sykiotis, Maria Kaselimi, Anastasios Doulamis, and Nikolaos Doulamis. ELECTRIcity: An Efficient Transformer for Non-Intrusive Load Monitoring. Sensors, 22(8), 2022.
- Stavros Sykiotis, Sotirios Athanasoulias, Maria Kaselimi, Anastasios Doulamis, Nikolaos Doulamis, Lina Stankovic, and Vladimir Stankovic. Performance-Aware NILM Model Optimization for Edge Deployment. IEEE Transactions on Green Communications and Networking, 7(3):1434`1446, 2023.
- Stavros Sykiotis, Christoforos Menos-Aikateriniadis, Anastasios Doulamis, Nikolaos Doulamis, and Pavlos S. Georgilakis. A self-sustained EV charging framework with N-step deep reinforcement learning. Sustainable Energy, Grids and Networks, 35:101124, 2023.
- Christoforos Menos-Aikateriniadis, Stavros Sykiotis, and Pavlos S. Georgilakis. Unlocking the potential of smart EV charging: A user-oriented control system based on deep reinforcement learning. Electric Power Systems Research, 230:110255, 2024.
- Sotirios Athanasoulias, Stavros Sykiotis, Maria Kaselimi, Anastasios Doulamis, Nikolaos Doulamis, and Nikolaos Ipiotis. OPT-NILM: An Iterative Prior-to-Full-Training Pruning Approach for Cost-Effective User Side Energy Disaggregation. IEEE Transactions on Consumer Electronics, 2023.
- Nicholas E. Protonotarios, Iason Katsamenis, Stavros Sykiotis, Nikolaos Dikaios, George A. Kastis, Sofia N. Chatziioannou, Marinos Metaxas, Nikolaos Doulamis and Anastasios Doulamis, A few-shot u-net deep learning model for lung cancer lesion segmentation via PET/CT imaging, Biomed. Phys. Eng. Express 8 (2) (2022), 025019

## Conferences

- Stavros Sykiotis, Maria Kaselimi, Anastasios Doulamis, and Nikolaos Doulamis. Continilm: A Continual Learning Scheme for Non-Intrusive Load Monitoring. In ICASSP 2023 - 2023 IEEE International Conference on Acoustics, Speech and Signal Processing (ICASSP), pages 1-5, 2023.